\documentclass{article}

 \usepackage[main, final]{neurips_2026}

\usepackage[utf8]{inputenc} %
\usepackage[T1]{fontenc}    %
\usepackage{hyperref}       %
\usepackage{booktabs}       %
\usepackage{amsfonts}       %
\usepackage{nicefrac}       %
\usepackage{microtype}      %
\usepackage{xcolor}         %

\definecolor{mydarkblue}{rgb}{0,0.08,0.45}
\hypersetup{ %
   colorlinks=true,
   linkcolor=mydarkblue,
   citecolor=mydarkblue,
   filecolor=mydarkblue,
   urlcolor=mydarkblue,
}
\setcitestyle{round}

\usepackage{xurl}
\usepackage{graphicx}
\usepackage{subcaption}
\usepackage{xspace}

\usepackage{amsmath}
\usepackage{amssymb}
\usepackage{mathtools}
\usepackage{amsthm}

\usepackage[capitalize,noabbrev]{cleveref}

\usepackage{tikz}           %
\usetikzlibrary{backgrounds, calc, positioning, arrows.meta, pgfplots.groupplots} %
\usepackage{pgfplots}       %
\pgfplotsset{compat=1.18}   %

\usepackage{pifont} %
\usepackage[shortlabels]{enumitem} %
\newcommand{\paren}[1]{\left(#1\right)} %
\newcommand{\brackets}[1]{\left[#1\right]} %
\newcommand{\abs}[1]{\left\lvert #1\right\rvert} %
\newcommand{\set}[1]{\left\{#1\right\}} %

\newcommand{\RR}{\mathbb{R}}
\newcommand{\NN}{\mathbb{N}}
\DeclareMathOperator*{\argmax}{arg\,max}

\newcommand{\indicator}{\mathbb{I}} %

\newcommand{\ourMethod}{\textsc{Rules}\xspace} %
\newcommand{\ourMethodSideChannel}{\textsc{Runes}\xspace} %
\newcommand{\ourMethodTitle}{\textsc{Rules}\xspace} %
\newcommand{\ourMethodSideChannelTitle}{\textsc{Runes}\xspace} %
\newcommand{\ruleName}{Temporal Conjunctive Prediction Rule\xspace}
\newcommand{\seqElementText}{event}
\newcommand{\SeqElementText}{Event}

\newcommand{\seq}{S} %
\newcommand{\seqLen}{L} %
\newcommand{\seqLenNorm}{\seqLen_{c}} %
\newcommand{\seqInd}{l} %
\newcommand{\predInd}{k} %
\newcommand{\seqElement}{x} %
\newcommand{\antElement}{v} %
\newcommand{\seqTime}{t} %
\newcommand{\numSamples}{N} %
\newcommand{\sampleInd}{n} %
\newcommand{\vocab}{\mathcal{V}} %
\newcommand{\vocabSize}{D} %
\newcommand{\vocabInd}{d} %
\newcommand{\seqVocab}{\mathcal{S}\paren{\vocab}} %

\newcommand{\ant}{A} %
\newcommand{\cons}{B} %
\newcommand{\consLogits}{o} %
\newcommand{\ruleTo}{\Rightarrow} %
\newcommand{\genericRule}{\ant\ruleTo\cons}
\newcommand{\indicatorRule}{\mathbb{I}} %
\newcommand{\minOffset}{\delta^{\mathrm{min}}} %
\newcommand{\maxOffset}{\delta^{\mathrm{max}}} %
\newcommand{\delayInterval}{\Delta} %
\newcommand{\ruleset}{\mathbf{R}}

\newcommand{\defeq}{=} %

\newcommand{\predAct}{p} %
\newcommand{\eventAct}{e} %
\newcommand{\softInterval}{\pi}
\newcommand{\antAct}{a} %
\newcommand{\firstWeights}{W} %
\newcommand{\secondWeights}{U} %
\newcommand{\outputWeights}{O} %
\newcommand{\countInNeurons}{I} %
\newcommand{\countAntNeurons}{J} %
\newcommand{\maxLookback}{\omega} %
\newcommand{\threshold}{\theta} %

\DeclareMathOperator{\softmax}{softmax}
\DeclareMathOperator{\softplus}{softplus}
\DeclareMathOperator{\lse}{lse}
\newcommand{\CE}{\mathcal{L}_{\mathrm{CE}}}

\newcommand{\fracCons}{r_{\mathrm{cons}}} %
\newcommand{\fracTarg}{r_{\mathrm{targ}}} %
\newcommand{\simRules}{s} %
\DeclareMathOperator{\simRulesets}{sim} %

\colorlet{storeblue}{blue}
\colorlet{storeorange}{orange}

\definecolor{blue}{rgb}{0, 0.447, 0.698}
\definecolor{orange}{rgb}{0.96, 0.624, 0}
\definecolor{green}{rgb}{0.55, 0.71, 0.0}
\definecolor{darkgreen}{rgb}{0.18, 0.35, 0.05}
\definecolor{purple}{rgb}{0.68, 0.36, 0.51}
\definecolor{red}{rgb}{0.9, 0.17, 0.31}
\definecolor{yellow}{rgb}{0.85, 0.694, 0.125}
\definecolor{cyan}{rgb}{0.301, 0.745, 0.933}

\title{What Comes Next and Why: Interpretable Next-Event Prediction with Neuro-Symbolic Rules}

\author{%
    Tim N.~Bauerschmidt\thanks{Correspondence to \textcolor{mydarkblue}{\texttt{bauerschmidt@cs.uni-saarland.de}}.} \\
    Department of Computer Science \\
    Saarland University \\
    \texttt{bauerschmidt@cs.uni-saarland.de} \\
    \And
    Isabel Valera \\
    Department of Computer Science \\
    Saarland University \\
    \texttt{ivalera@cs.uni-saarland.de} \\
    \And 
    Jilles Vreeken \\
    CISPA Helmholtz Center for Information Security \\
    \texttt{vreeken@cispa.de}
}

\begin{document}

\maketitle

\begin{abstract}
   In real-world applications, being able to explain \emph{why} a prediction is made is often as important---if not more so---than achieving high accuracy. 
    However, existing methods for sequential event prediction typically prioritize accuracy at the expense of interpretability.
    Neuro-symbolic approaches offer a promising direction to improve interpretability by learning symbolic rules, but extending them to next-event prediction is challenging due to the presence of temporal dependencies.
    We introduce \ourMethod, a neuro-symbolic model that learns sequential rules directly from data, and combine it with a residual network to form \ourMethodSideChannel{} (Rule Network for Event Sequences). 
    \ourMethod{} accurately recovers ground-truth rules from synthetic data and discovers compact, interpretable rule sets on real data.
    \ourMethodSideChannel{} combines these rules additively with a residual network, achieving competitive accuracy while providing transparent access to active rules and how residuals adjust predictions.
     Our code is available at \url{https://github.com/IamginE/runes-rules}.
\end{abstract}

\section{Introduction}
Next-\seqElementText{} prediction is the task of predicting the next categorical element in a temporally ordered sequence.
Beyond next-token prediction in language models---where token positions can be seen as discrete event times---next-\seqElementText{} prediction arises in many domains, including e-commerce \citep{2023_application_product_recommendations}, healthcare \citep{2022_application_healthcare_1,2023_application_healthcare_2}, and process log analysis \citep{2019_application_process_logs_1,2021_application_process_logs_2}. 
Neural sequence models provide accurate predictions across domains and are thus increasingly adopted in practice.
However, their internal representations are opaque, making predictions difficult to interpret or justify. 
This is especially problematic in safety- and regulation-sensitive contexts, where interpretability is key. 

Symbolic approaches \citep{1997_freq_episodes,2012_cmrules,2021_pfv_4,2026_seqret} instead focus on describing event dependencies using sequential rules. 
A sequential rule $\genericRule$ expresses that when the antecedent $\ant$ has been observed, the consequent $\cons$ is likely to occur later in the sequence.
In e-commerce, such a rule might specify that customers who purchased a computer case and a computer processor are likely to subsequently purchase a motherboard or a power supply unit.
However, existing rule-discovery methods focus on enumerating frequent or high-confidence rules without a principled way to integrate them into a unified predictive model.

Neuro-symbolic models offer a promising approach to address the accuracy--interpretability tension.  
For instance, recent neuro-symbolic methods for tabular data use differentiable relaxations to learn interpretable and accurate prediction rules~\citep{2021_accurate_decision_rules_NN,2021_RRL_MLLP_v2,2024_diffnaps,2025_neurules}.
However, their reliance on binarized tabular inputs hinders applicability to sequential data, where temporal structure is central.
Naively flattening sequences leads to sparse, high-dimensional representations, limiting both scalability and expressiveness.

To fill this gap, we define \emph{a class of sequential rules} that incorporate temporal information and can be efficiently integrated into neural network layers for prediction.
We then propose \ourMethod{}, \emph{a neuro-symbolic model} that learns these rules directly from data, and its extension \ourMethodSideChannel{}, which combines \ourMethod{} with a residual network.
Training proceeds in two phases using a self-supervised next-\seqElementText{} prediction objective.
In the \emph{symbolic phase}, we train \ourMethod{} using temperature annealing~\citep{2017_gumbel_softmax} to enable gradient-based rule learning. 
After annealing, the layers of \ourMethod{} represent symbolic conditions over \seqElementText{}s and their relative offsets to the predicted \seqElementText{}, yielding meaningful binary representations for sequences. 
In the \emph{predictive phase}, we obtain \ourMethodSideChannel{} by training a residual network that corrects the predictions of the frozen \ourMethod{} through additive logit residuals.
This phase improves predictive performance while preserving rule-level interpretability.

We evaluate \ourMethod{} and \ourMethodSideChannel{} on synthetic and real-world datasets, showing that \ourMethod{} recovers ground-truth generative rules on synthetic data and learns compact, meaningful rule sets on real data. 
The residual network enables \ourMethodSideChannel{} to achieve competitive accuracy by capturing structure not explained by \ourMethod{}. 
Importantly, \ourMethodSideChannel{} allows tracing the impact of the residual network, which is mostly used to correct errors of \ourMethod{}, maintaining interpretability when learned rules are reliable. 
\section{Neuro-Symbolic Sequential Rule Learning}
\label{sec:methodology}
In this section, we formally introduce \ourMethod, an inherently interpretable network for learning \ruleName{}s, and \ourMethodSideChannel, a neuro-symbolic architecture that combines \ourMethod with a residual network for improved accuracy.

We consider a dataset $\paren{\seq^1, \seq^2,\dots, \seq^\numSamples}$ of $\numSamples$ sequences over a finite \textbf{vocabulary} $\vocab \defeq \set{1, 2,\dots, \vocabSize}$.
An \textbf{event sequence} $\seq\defeq \paren{\paren{\seqElement_1, \seqTime_1},\dots, \paren{\seqElement_\seqLen, \seqTime_\seqLen}}$ consists of $\seqLen$ tuples $(\seqElement_\seqInd, \seqTime_\seqInd)$, where $\seqElement_\seqInd \in \vocab$ is the \seqElementText{} with associated occurrence time $\seqTime_\seqInd\in \RR_{\geq 0}$, and $\seqTime_{\seqInd} < \seqTime_{k}$ for $\seqInd < k$.
We denote by $\seq_{<\seqTime} = \paren{\paren{\seqElement_1, \seqTime_1},\dots, \paren{\seqElement_{\predInd-1}, \seqTime_{\predInd-1}}}$ the subsequence of $\seq$ that contains all \seqElementText{}s before time $\seqTime$, i.e., $\seqTime_{\predInd-1} < \seqTime \leq \seqTime_{\predInd}$; and 
by $\seqVocab$  the set of all valid sequences over $\vocab$.

\subsection{\ruleName{}s}
\label{subsec:tcp_rules}
We are interested in \textbf{sequential rules} $\genericRule$ that predict the next event $\seqElement_\predInd$ from a historical sequence $\seq_{<\seqTime_\predInd}$.
A rule consists of an \textbf{antecedent} $\ant$, which determines when the rule applies to a sequence $\seq$, and a \textbf{consequent} $\cons$, which specifies the predictions when the antecedent conditions are satisfied. 
An example of such a rule in e-commerce (within our proposed rule class) is: ``\textbf{if} \textit{a customer purchased a `computer case' 1--43 days ago} \textbf{and} \textit{a `computer processor'  2--21 days ago}, \textbf{then} \textit{the next \seqElementText{} is likely to be the purchase of a `motherboard'} \textbf{or} \textit{a `power supply unit'.}''

We impose four desiderata on sequential rules: they should (i) be \emph{interpretable}; (ii) capture \emph{temporal dependencies}, requiring consequents to occur after antecedents and specifying the time interval during which a rule remains active; (iii) be \emph{composable}, allowing multiple rules to be combined within a single predictive model; and (iv) support \emph{efficient learning and inference}. 
To meet these requirements, we introduce \textbf{\ruleName{}s}, hereafter referred to as rules.

\textbf{Antecedent.} The antecedent $\ant$ of a rule consists of a set of \seqElementText{}s $\antElement_i \in \vocab$, each paired with a delay interval $[\minOffset_i, \maxOffset_i]$.
A rule activates for a prediction on sequence $\seq$ at time $\seqTime$ if, for every antecedent element, there exists a matching \seqElementText{} in the history $\seq_{<\seqTime}$ whose temporal offset from $\seqTime$ lies within the corresponding interval.
Thus, each antecedent element induces an \textbf{event-time predicate} %
and rule activation is determined by the \textbf{conjunction} of these predicates.
In other words, the antecedent  symbolically captures relevant \seqElementText{}s and their temporal offsets relative to the predicted consequent. 

Formally, we define the antecedent of a rule $\genericRule$ as a set $\ant \defeq \set{\paren{\antElement_i, \brackets{\minOffset_i, \maxOffset_i}}\mid 1\leq i\leq I}$, where $\antElement_i\in \vocab$, $\minOffset_i, \maxOffset_i\in \RR_{>0}$, $\minOffset_i \leq \maxOffset_i$, and $\ant\neq \emptyset$.
Denoting $\delayInterval_i \defeq \brackets{\minOffset_i, \maxOffset_i}$, the activation function $\indicatorRule_\ant:\seqVocab \times \RR_{\ge 0} \to \{0,1\}$ of antecedent $\ant$ is given by
\begin{equation}
      \indicatorRule_\ant\paren{\seq,\seqTime} \defeq  \bigwedge_{(\antElement_i,  \delayInterval_i)\in \ant} \underset{\text{event-time predicate } \predAct^*_i\paren{\seq_{<\seqTime}}}{\underbrace{\bigvee_{\paren{\seqElement_\seqInd, \seqTime_\seqInd} \in \seq_{<\seqTime}}
       (\seqElement_\seqInd = \antElement_i) \land \paren{\seqTime - \seqTime_\seqInd \in \delayInterval_i}}}, \label{eq:ant_activation_logical}
\end{equation}
such that $\indicatorRule_\ant(\seq,\seqTime) = 1$ indicates that $\ant$ is satisfied at time~$\seqTime$.
Antecedents are unordered, but delay intervals induce an implicit partial order over their events.

\textbf{Consequent.} The consequent is a symbolic subset $\cons \subseteq \vocab$ that specifies the likely outcomes when the antecedent is satisfied. 
However, symbolic consequents are problematic for prediction: they do not convey relative likelihoods among \seqElementText{}s, and they hinder composability because the consequents of rules are not directly comparable. 
We therefore relax consequents into logit vectors $\consLogits_\cons \in \RR^\vocabSize$ over $\vocab$, and compute the predictive distribution for a sequence $\seq$ at time $\seqTime$ using a set of rules $\ruleset$ as
\begin{equation}
    \softmax \paren{\sum_{\genericRule\in \ruleset} \indicatorRule_\ant\paren{\seq,\seqTime}\cdot \consLogits_\cons}.
    \label{eq:softmax_ruleset}
\end{equation}

In the following, we show how to learn a set of rules via continuous optimization, avoiding explicit traversal of the super-exponential search space resulting from possible antecedents and rule selection.

\subsection{The \ourMethodSideChannelTitle{} Architecture}
\label{subsec:architecture}
\begin{figure}[t]
    \centering
    \begin{subfigure}{0.36\linewidth}
    \raggedright
    \begin{tikzpicture}[
        scale=0.82, transform shape,
        heading/.style={rectangle, draw=none, text=black, minimum size=7mm, align=center, font=\small},
        layer/.style={rectangle, thick, draw=black, text=black, minimum size=7mm, align=center, font=\small},
        blank/.style={rectangle, draw=none, fill=white, text=black, font=\small}
    ]
        \node[layer, fill=white] (input) at (5,5) {$\seq_{<\seqTime_{\predInd}}$, $\seqTime_{\predInd}$};
        \node[layer, fill=blue!40, minimum height=8mm] (rulenet) at ($(input) - (1.3,1.1)$) {\ourMethod{}};
        \node[layer, fill=green!40] (resnet) at ($(input) - (-1.8,1.1)$) {Residual \\ Network};
        \node[heading] (phase1) at ($(input) + (-1.3,1.0)$) {\textbf{Symbolic Phase:} \\ Rule Learning};
        \node[heading] (phase1) at ($(input) + (1.8,1.0)$) {\textbf{Predictive Phase:} \\ Boosting Accuracy};
    
        \draw[-{Stealth}, draw=black, thick] (input.west) -| (rulenet.north);
        \draw[-{Stealth}, draw=black, thick] (input.east) -| (resnet.north);
    
        \node[layer, fill=purple!40] (xsym) [below=0.5cm of rulenet] {$\hat{\seqElement}_{\predInd}^{(r)}$};
        \node[layer,  fill=green!40] (xres) [below=0.5cm of resnet]  {$\hat{\seqElement}_{\predInd}^{(\delta)}$};
    
        \draw[-{Stealth}, draw=black, thick] (rulenet.south) -- (xsym.north);
        \draw[-{Stealth}, draw=black, thick] (resnet.south)  -- (xres.north);
    
        \node[layer] (symloss) [below=1.0cm of xsym] {$\CE\left(\mathcolor{purple}{\hat{\seqElement}_{\predInd}^{(r)}}, \seqElement_\predInd\right) $};
        \node[layer] (resloss) [below=1.0cm of xres]  {$\CE\left(\hat{\seqElement}_{\predInd}^{(r)}+ \mathcolor{green}{\hat{\seqElement}_{\predInd}^{(\delta)}}, \seqElement_\predInd\right) $};
        \node[circle, thick, draw=black, text=black, minimum size=3mm, inner sep=0pt, align=center, font=\small] (plus) at ($(xres.south)!0.5!(resloss.north)$) {$+$};
        \coordinate (symplus) at ($(xsym.south)!0.5!(symloss.north)$);
        \node[blank] at ($(symplus)!0.5!(plus) + (-0.2,0.25)$) {frozen};
    
        \draw[-{Stealth}, draw=black, thick] (xsym.south) -- (symloss.north);
        \draw[draw=black, thick] (xres.south) -- (plus.north);
        \draw[-{Stealth}, draw=black, thick] (plus.south) -- (resloss.north);
        \draw[-{Stealth}, draw=black, thick] (xsym.south)  |- (plus.west);
    
        \begin{scope}[on background layer]
           \draw[thick, dashed, black!40] ($(symloss.east)!0.5!(resloss.west) - (0, 0.5)$) -- ($(symloss.east)!0.5!(resloss.west) + (0, 5.6)$);
        \end{scope}
    
    \end{tikzpicture}
    \caption{Architecture of \ourMethodSideChannel{}.}
    \label{subfig:model_architecture}
    \end{subfigure}
    \hfill
    \begin{subfigure}{0.63\linewidth}
    \raggedleft
    \begin{tikzpicture}[
        scale=0.82, transform shape,
        heading/.style={rectangle, draw=none, text=black, minimum size=7mm, align=center, font=\small},
        layer/.style={rectangle, thick, draw=black, text=black, minimum size=7mm, align=center, font=\small},
        vdot/.style={rectangle, draw=none, fill=none, text=black, font=\small}
    ]

        \coordinate (input-top)   at (0,10);
        \coordinate (hidden1-top) at ($(input-top)   + (3.1,0)$);
        \coordinate (hidden2-top) at ($(hidden1-top) + (3.2,0)$);
        \coordinate (output-top)  at ($(hidden2-top) + (2.3,0)$);

        \node[heading] [above=0.15cm of input-top] {\textbf{Input Sequence} \\ \phantom{(Antecedent)}};
        \node[heading] at ($(input-top) + (0.0,-0.1)$) {$\seq_{<\seqTime_{\predInd}}$};

        \node[layer, minimum width=2cm] (i1) [below=0.65cm of input-top] {$(\seqElement_1, \seqTime_1)$};
        \node[layer, minimum width=2cm] (i2) [below=0.15cm of i1]        {$(\seqElement_2, \seqTime_2)$};
        \node[layer, minimum width=2cm] (i3) [below=0.15cm of i2]        {$(\seqElement_3, \seqTime_3)$};
        \node[vdot]  (i4) [below=0.0cm of i3]        {$\vdots$};
        \node[layer, minimum width=2cm] (i5) [below=0.10cm of i4]        {$(\seqElement_{\predInd-1}, \seqTime_{\predInd-1})$};

        \node[heading] [above=0.15cm of hidden1-top] {\textbf{Event-Time Predicates} \\ \phantom{(Antecedent)}};
        \node[heading] at ($(hidden1-top) + (0.0,-0.1)$) {$\paren{\seqElement_\seqInd = \argmax_\vocabInd \firstWeights_{i,d}}$\\[0.3em] $\land\ \paren{t_{\predInd} - t_l \in \Delta_i}$};

        \node[layer, fill=blue!40] (h1a) [below=0.65cm of hidden1-top] {$\predAct_1\paren{\seq_{<\seqTime_{\predInd}}; \firstWeights_1, \Delta_1}$};
        \node[layer, fill=blue!40] (h1b) [below=0.15cm of h1a]         {$\predAct_2\paren{\seq_{<\seqTime_{\predInd}}; \firstWeights_2, \Delta_2}$};
        \node[vdot] (h1c) [below=0.0cm of h1b] {$\vdots$};
        \node[layer, fill=blue!40] (h1d) [below=0.10cm of h1c]         {$\predAct_i\paren{\seq_{<\seqTime_{\predInd}}; \firstWeights_i, \Delta_i}$};
        \node[layer] (tin) [below=0.5cm of h1d] {$t_{\predInd}$};
        
        \draw[-{Stealth}, draw=black, thick] (tin.north) -- ($(h1d.south) + (0.0,-0.1)$);
        \node[rectangle, draw=none, text=black, minimum size=7mm, align=center, font=\footnotesize] at ($(i5)!0.5!(h1d) - (0.0,0.3)$) {$\firstWeights_i$};

        \node[heading] [above=0.15cm of hidden2-top] {\textbf{Conjunction} \\ (Antecedent)};
        \node[heading] at ($(hidden2-top) + (0.0,-0.1)$) {$\bigwedge_{i: \secondWeights_{j,i} > 0} p_i$};

        \node[layer, fill=orange!40] (h2a) [below=0.65cm of hidden2-top] {$\antAct_1\paren{\seq_{<\seqTime_{\predInd}};\secondWeights_1}$};
        \node[layer, fill=orange!40] (h2b) [below=0.15cm of h2a]         {$\antAct_2\paren{\seq_{<\seqTime_{\predInd}};\secondWeights_2}$};
        \node[vdot] (h2c) [below=0.0cm of h2b]   {$\vdots$};
        \node[layer, fill=orange!40] (h2d) [below=0.10cm of h2c]         {$\antAct_j\paren{\seq_{<\seqTime_{\predInd}};\secondWeights_j}$};

        \node[rectangle, draw=none, text=black, minimum size=7mm, align=center, font=\footnotesize] at ($(h1d.east)!0.5!(h2d.west) - (0.0,0.5)$) {$\secondWeights_j$};

        \node[heading] [above=0.15cm of output-top] {\textbf{Prediction} \\ (Consequent)};
        \node[circle, thick, draw=black, text=black, minimum size=3mm, inner sep=0pt, align=center, font=\small] (o1) [right=0.3cm of h2b] {$+$};
        \node[layer, fill=purple!40] (o2) [below=1.525cm of output-top] {$\hat{\seqElement}_{\predInd}^{(r)}$};
        \node[rectangle, draw=none, text=black, minimum size=7mm, align=center, font=\footnotesize] at ($(h2d.east)!0.5!(o1.west) + (0.6,-0.1)$) {$\outputWeights_j$\\(logits)};

        \begin{scope}[on background layer]
            \foreach \i in {i1,i2,i3,i5} {
                \foreach \h in {h1a,h1b} {
                    \draw[draw=black!30, thick] (\i.east) -- (\h.west);
                }
            }
            \foreach \i in {i1,i2,i3,i5} { 
                \draw[draw=black, thick] (\i.east) -- (h1d.west); 
            }

            \foreach \hA in {h1a,h1b,h1d} {
                \foreach \hB in {h2a,h2b} {
                    \draw[draw=black!30, thick] (\hA.east) -- (\hB.west);
                }
            }

            \foreach \hA in {h1a,h1b,h1d} {
                \draw[draw=black, thick] (\hA.east) -- (h2d.west);  
            }

            \draw[draw=black!30, thick] (h2a.east) -- (o1.north);
            \draw[draw=black!30, thick] (h2b.east) -- (o1.west);
            \draw[draw=black, thick] (h2d.east) -- (o1.south);
            \draw[draw=black, thick] (o1.east) -- (o2.west);
        \end{scope}
    \end{tikzpicture}
    \caption{Architecture of \ourMethod{}.}
    \label{subfig:rule_network_architecture}
    \end{subfigure}
    \caption{\textit{Architecture overview of \ourMethodSideChannel{}.} 
        \textbf{(a)}~\ourMethodSideChannel{} is trained in two phases: \ourMethod{} first learns sequential rules, after which a residual network is trained with the \ourMethod{} network frozen.
        \textbf{(b)}~\ourMethod{} represents rules in its layers:
        Neurons in the first layer encode event-time predicates.
        Neurons in the second layer select a subset of these predicates and combine them into conjunctions (antecedents).
        The output layer sums the softmax logits of all active rules to produce the final prediction. 
     }
     
    \label{fig:model_architecture}
\end{figure}
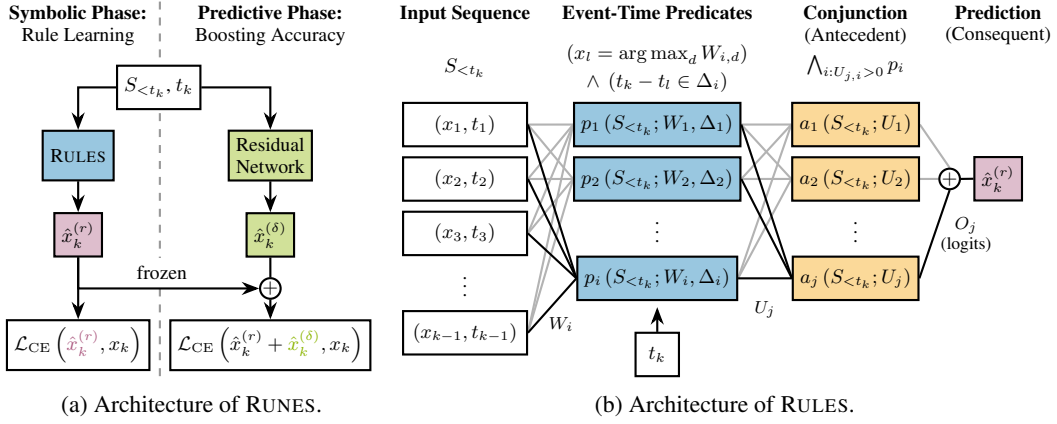

We begin by describing the architecture of \ourMethodSideChannel{}.
As shown in \cref{subfig:model_architecture}, \ourMethodSideChannel{} comprises two components: a rule network, denoted as \ourMethod{}, and a residual network that captures structure not explained by rules.
Both predict the next \seqElementText{} $\seqElement_\predInd$ from a historical sequence $\seq_{<\seqTime_\predInd}$ and its occurrence time $\seqTime_\predInd$.
To ensure stable rule discovery and prevent rules from being subsumed by the residual network, training is decoupled into two phases: (i) a \textit{symbolic phase}, in which \ourMethod learns rules via temperature annealing, followed by (ii) a \textit{predictive phase}, in which \ourMethod is frozen and the residual network is trained to improve accuracy.
We now discuss both networks in detail.

\textbf{\ourMethod.}
\ourMethod (\cref{subfig:rule_network_architecture}) learns a set of sequential rules by encoding antecedents in two hidden layers and consequents in the output layer.
The first hidden layer consists of neurons that approximate \textcolor{blue}{\textbf{event-time predicates} $\predAct_i$}, such as ``\textit{the event `computer case' occurred 1--43 days ago}''.
Each neuron~$i$ is parameterized by learnable weights $\firstWeights_i \in \RR^\vocabSize$, which identify an \seqElementText{}, along with delay bounds $\minOffset_i, \maxOffset_i \in \RR_{>0}$. 
Temperature annealing binarizes the activation of each neuron, after which $\firstWeights_i$ selects an \seqElementText{} via its maximum weight and $\Delta_i = [\minOffset_i, \maxOffset_i]$ specifies the delay interval.

To encode rule antecedents, neurons in the second hidden layer combine subsets of predicates from the previous layer into \textcolor{orange}{\textbf{conjunctions} $\antAct_j$}.
For example, one might represent ``\textit{the event `computer case' occurred 1--43 days ago} \textbf{and} \textit{the event `computer processor' occurred 2--21 days ago}''.
Each antecedent neuron $j$ selects a subset of predicates $\predAct_i$ via the positive entries in its weights $\secondWeights_{j,i}$ and, after annealing, activates only if all selected predicate neurons are active.

To \textcolor{purple}{\textbf{predict the next \seqElementText{}} $\hat{\seqElement}_{\predInd}^{(r)}$}, \ourMethod{} represents consequents with a softmax-linear output layer.
Since the second layer encodes binary conjunctions after training, the softmax layer sums the learned logits of all active rules without any weighting.
This allows us to identify the most likely consequent \seqElementText{}s of the rules learned by the network, e.g., \textit{motherboard} \textbf{or} \textit{power supply unit}. %
Particularly, since rule logits appear unweighted in the symbolic regime ($\antAct_j \in \{0,1\}$), the most influential consequent events of a rule can be identified directly via their largest logits, and, respecting the translation invariance of softmax, these logits are comparable across rules.
We detail an algorithm that uses these insights to extract rules with symbolic consequents in Appendix~\ref{app:subsec_rule_set_extraction} (concrete example in \cref{fig:example_rule_set_exrtraction}).

\textbf{Residual Network.}
Not all data strictly follows symbolic rules.
To improve accuracy without sacrificing interpretability, we augment \ourMethod with a residual network to obtain \ourMethodSideChannel{}, following hybrid design strategies from concept bottleneck models \citep{2022_CBM_side_channel_1,2023_CBM_side_channel_2,2025_CBM_side_channel_3}.
Given a \ourMethod network that outputs softmax logits $\hat{\seqElement}_\predInd^{(r)}\in \RR^\vocabSize$, the residual network outputs corrective logits $\hat{\seqElement}_\predInd^{(\delta)}$, and the final prediction is computed as $\hat{\seqElement}_\predInd^{(r)} + \hat{\seqElement}_\predInd^{(\delta)}$.
During training, we freeze \ourMethod, restricting the residual network to learning logit residuals.
At inference time, interpretability of the learned rules is preserved: \emph{for each input, we can inspect which rules activate and quantify how the residual network adjusts the final prediction}, with the additive logits enabling direct attribution of each rule and residual contributions.

\subsection{\ourMethod{}: Predicate Layer}
We now detail how we can differentiably learn sequential rules, starting with the first layer of \ourMethod{}.
In this layer, each neuron $i$ encodes the disjunctive logical formula of an event-time predicate $\predAct^*_i$  (\cref{eq:ant_activation_logical}), checking whether a specific \seqElementText{} occurred within a target time interval relative to the prediction time.
We approximate this predicate with a differentiable function $\predAct_i$ that depends on the temperature of the first layer $\tau_1$ and satisfies
$\lim_{\tau_1\to 0^+} \predAct_i\paren{\seq_{<\seqTime_{\predInd}}; \tau_1} =  \predAct_i^*\paren{\seq_{<\seqTime_{\predInd}}}$.\footnote{
    Equality may not hold in some exceptional cases, which do not negatively impact learning and are discussed in Appendix~\ref{app:sec_convergence_proofs}.
}
Here we omit the prediction time $\seqTime_{\predInd}$ as an explicit argument.

\textbf{Learning \SeqElementText{}s.}
To learn \seqElementText{}s that form rules, we parameterize $\antElement_i$ in the  \seqElementText{} match ($\seqElement_{\seqInd} = \antElement_i$) with a vector $\firstWeights_i\in \RR^\vocabSize$.
Each \seqElementText{} in the vocabulary $\vocab$ is assigned a weight in $\firstWeights_i$, and at the end of training, the maximum weight determines the selected \seqElementText{} for neuron~$i$.
Mathematically, we compute a soft match $\eventAct_i$, indicating whether $\seqElement_\seqInd$ corresponds to the maximum weight of $\firstWeights_i$
\begin{equation}
     \eventAct_i(\seqElement_\seqInd; \firstWeights_i, \tau_1) = \sigma\paren{\frac{1}{\tau_1^2}\paren{\firstWeights_{i,\seqElement_\seqInd} - (1-\tau_1)\lse(\firstWeights_i; \tau_1)}}, \label{eq:event_act}
\end{equation}
where $\lse(w;\tau) = \tau \log\paren{\sum_{\vocabInd=1}^\vocabSize\exp\paren{\frac{w_\vocabInd}{\tau}}}$ is the temperature-scaled LogSumExp function \citep{2004_convex_optimization} and $\sigma$ denotes the sigmoid function. 
At $\tau_1 = 1$, the activation $\eventAct_i$ behaves like a standard sigmoid layer applied to the one-hot encoding of $\seqElement_{\seqInd}$.
As $\tau_1 \to 0$, the LogSumExp term converges to the max operator, while the sigmoid converges to the Heaviside step function.
With the scaling ($\frac{1}{\tau_1^2}$), this yields an activation that is $1$ for the \seqElementText{} with maximal weight and $0$ for all others.
Formally, $\lim_{\tau_1 \to 0^+} \eventAct_i(\seqElement_\seqInd; \firstWeights_i, \tau_1) = \indicator(\seqElement_\seqInd \defeq \argmax_d \firstWeights_{i,d})$, where $\indicator$ is the indicator function (see Appendix~\ref{app:par_temp_annealing_event_match}).
If $\max_d \firstWeights_{i,d} < 0$, the limit instead yields the zero function $\eventAct_i(\seqElement_{\seqInd}) = 0$ for all inputs, allowing the model to disable neurons by setting all weights to negative values.

\textbf{Learning Delay Intervals.}
The second component that the model needs to learn is the temporal offsets of antecedent \seqElementText{}s relative to the prediction time (consequent).
We adopt the differentiable soft-binning function of \citet{2018_deep_neual_decision_trees}, applied to learning discrete intervals by \citet{2024_syflow}:
\begin{equation}
     \softInterval(z;a,b,\tau) = \frac{e^{\frac{1}{\tau} \paren{2z - a}}}{e^{\frac{1}{\tau} z} + e^{\frac{1}{\tau} \paren{2z - a}} + e^{\frac{1}{\tau} \paren{3z - a - b}}}. \label{eq:delay_act}
\end{equation}
Specifically, for neuron $i$, we introduce parameters $\minOffset_i, \maxOffset_i\in \RR_{>0}$ to form the delay interval $\delayInterval_i = [\minOffset_i,\maxOffset_i]$.
We then use the relationship
$\lim_{\tau_1\to 0^+} \softInterval_i\paren{\seqTime_\predInd - \seqTime_\seqInd;\minOffset_i,\maxOffset_i, \tau_1} = \indicator\paren{\seqTime_\predInd - \seqTime_\seqInd\in \delayInterval_i}$
to find the delay constraint for the learned antecedent \seqElementText{}. 
To ensure valid intervals, we clip $\minOffset_i$ and $\maxOffset_i$ after each gradient step, performing projected gradient descent \citep{2018_pgd}.

\textbf{Learning Event-Time Predicates.}
The soft predicates $\predAct_i$ and $\softInterval_i$ both take values in $[0,1]$, and converge to their logical counterparts as the temperature decreases.
Thus, we can approximate their conjunction $\indicator((\seqElement_\seqInd = \antElement_i)\land (\seqTime_\predInd - \seqTime_\seqInd\in \delayInterval_i))$ by their product
\begin{equation}
    \tilde{p}_i(\seqInd,\predInd) \defeq \eventAct_i(\seqElement_\seqInd; \firstWeights_i, \tau_1)\cdot \softInterval_i\paren{\seqTime_\predInd - \seqTime_\seqInd;\minOffset_i,\maxOffset_i, \tau_1}.
\end{equation}

To complete the predicate, we model the disjunction checking whether any \seqElementText{} in the sequence satisfies the type and delay specified by $\tilde{p}_i$.
We approximate this disjunction using a softmax relaxation of the maximum (Appendix~\ref{app:subsec_temp_annealing_basic}), which in the limit behaves like a logical OR when $\tilde{p}_i(\seqInd,\predInd)\in \{0,1\}$:
\begin{equation}
    \predAct_i\paren{\seq_{<t_{\predInd}};\firstWeights_i, \minOffset_i, \maxOffset_i, \tau_1} = \sum_{\seqInd=1}^{\predInd-1}\tilde{p}_i(\seqInd,\predInd)\cdot\frac{\exp\paren{\frac{\tilde{p}_i(\seqInd,\predInd)}{0.1\tau_1}}}{\sum_{m=1}^{\predInd-1}\exp\paren{\frac{\tilde{p}_i(m,\predInd)}{0.1\tau_1}}}. \label{eq:pred_act}
\end{equation}
We use a reduced temperature $0.1\tau_1$, as the inputs $\tilde{p}_i(\seqInd,\predInd)$ lie in $[0,1]$, where their proximity amplifies approximation errors at high temperatures.
In this step, masking can be done analogously to attention masking in transformers \citep{2017_attention_is_all_you_need} by setting certain $\tilde{p}_i(\seqInd,\predInd) = 0$. 
For example, to enforce a user-specified maximum lookback window $\maxLookback\in\RR_{>0}$, we mask all $\tilde{p}_i(\seqInd,\predInd)$ with $\seqTime_\predInd - \seqTime_\seqInd > \maxLookback$.

\subsection{\ourMethod{}: Antecedent Layer}
After modeling the event-time predicates in \cref{eq:ant_activation_logical}, we construct the outer conjunction via the second layer of \ourMethod{}, which combines predicates into antecedents.
Here, each neuron $j$ selects a set of predicates $\predAct_i$ with its weights $\secondWeights_{j,i}$. %
Following \citet{2020_LNN}, we use $\sum_{i}\secondWeights_{j,i}\cdot (\predAct_i(\seq_{<\seqTime_\predInd}) - 1)$ to implement a soft conjunction.
For $\secondWeights_{j,i} \geq 0$ and $\predAct_i(\seq_{<\seqTime_{\predInd}}) \in [0,1]$, this expression is non-positive and achieves its maximum of $0$ only when every selected predicate ($\secondWeights_{j,i}>0$) satisfies $\predAct_i=1$.
Thus, the neuron activates maximally when all chosen predicates hold.
To map the conjunction score to the binary range, we apply a sigmoid scaled with the temperature $\tau_2$ of the second layer
\begin{equation}
    \antAct_j\paren{\seq_{<\seqTime_{\predInd}};\secondWeights_j,\tau_2}\defeq \sigma\paren{\frac{1}{\tau_2}\paren{\sum_{i=1}^I\secondWeights_{j,i}\paren{\predAct_i(\seq_{<\seqTime_{\predInd}}) + \frac{\tau_2}{2}-1}}-\log(\tau_2)}.
    \label{eq:ant_act}
\end{equation}
The term $\tfrac{\tau_2}{2}$  prevents the summed input from becoming overly negative early in training, avoiding saturation of the sigmoid.
The bias term $-\log(\tau_2)$ shifts the sigmoid leftward affecting its convergence only around $0$, where it ensures that the activation approaches $1$ when the conjunction is satisfied.
We ensure $\secondWeights_{j,i}\geq 0$ through the temperature-based softplus function, which converges to the ReLU while allowing gradient flow for negative inputs.
In the limit $\tau_2\to 0^+$, $\antAct_j$ converges to $\indicator(\bigwedge_{i=1}^I(\secondWeights_{j,i}>0)\rightarrow \predAct_i(\seq_{<\seqTime_{\predInd}}))$, where $\rightarrow$ denotes implication (see Appendix~\ref{app:subsec_temp_annealing_antecedent}).

\subsection{\ourMethod{}: Consequent Layer}
The output layer of \ourMethod is a standard softmax-linear layer that sums over $\antAct_j(\seq_{<\seqTime_{\predInd}})\cdot \outputWeights_j$, where $\outputWeights_j\in \RR^\vocabSize$ are the logits of rule $j$.
After annealing, $\antAct_j$ are effectively binary, so \ourMethod aggregates the logits of all active rules, as in \cref{eq:softmax_ruleset}. 
The only difference is that we include a bias $b\in \RR^\vocabSize$, which suffices to learn independent noise and encourages the model to learn only meaningful rules.

\subsection{\ourMethodSideChannel{}: Training}
\label{subsec:training}
We train both networks of \ourMethodSideChannel{} using the cross-entropy loss $\CE$ for next-\seqElementText{} prediction over the training sequences $\seq^1,\dots,\seq^\numSamples$. 
Formally, the loss is given by
\begin{equation}
    \frac{1}{\seqLenNorm}\sum_{\sampleInd=1}^\numSamples \sum_{\predInd = 2}^{\seqLen^\sampleInd} \CE\paren{f(\seq^\sampleInd_{<\seqTime_\predInd^\sampleInd}, \seqTime_\predInd^\sampleInd), \seqElement^\sampleInd_{\predInd}}, \quad \text{where } \seqLenNorm = \sum_{\sampleInd=1}^\numSamples (\seqLen^\sampleInd - 1) \label{eq:objective}
\end{equation}
normalizes over all valid prediction steps, $f$ denotes the model used in the current training phase, and  $\sampleInd$ indicates the sequence.
This objective resembles next-token prediction \citep{2003_next_word_loss}, except the first \seqElementText{} of each sequence is excluded because no rule can be active without context.
Unlike conventional next-token models, both \ourMethod and \ourMethodSideChannel{} condition predictions on the target time, enabling rule learning from sequences with variable inter-event offsets.
As a consequence, inference requires specifying the prediction time. 
For regularly sampled sequences (constant $\seqTime_{\seqInd+1} - \seqTime_{\seqInd}$), the next prediction time is known, while for irregularly sampled sequences it can be provided as a user query or estimated by a separate model, e.g., a temporal point process \citep{2003_tpp_book}.

\textbf{Phase 1: Training \ourMethod.}
In the symbolic phase, we train \ourMethod{} in a self-supervised manner using the objective in \cref{eq:objective}.
For temperature annealing, we initialize $\tau_1=\tau_2=1$ and first anneal the predicate-layer temperature $\tau_1 \to 0$, followed by the antecedent-layer temperature $\tau_2 \to 0$, which yields more stable optimization than a simultaneous schedule.
The sizes of the predicate and antecedent layers control learning capacity and are treated as hyperparameters. 
We provide details on the layer initialization in Appendix~\ref{app:subsec_layer_init}.
Similar to transformers, \ourMethod{} scales quadratically in the sequence length $\seqLen$ while supporting parallel prediction across time steps (see Appendix~\ref{app:sec_discussion_complexity} for~details).

\textbf{Phase 2: Training the Residual Network.}
After training \ourMethod, we freeze its parameters, replace activations with their logical equivalents, and train a residual network using the same objective (\cref{eq:objective}) by adding its logits to those of \ourMethod.
In that way, the residual network learns to correct the errors made by \ourMethod.
The residual network can be instantiated by \textit{any} sequence model and trained following standard practice; in this work, we use a transformer \citep{2017_attention_is_all_you_need}.

\section{Related Work}
Modeling sequences with logical rules is related to temporal logic, which formalizes reasoning over time: our antecedents can be viewed as a past-time fragment of metric temporal logic (Appendix~\ref{app:sec_connection_temporal_logic}).
Prior work combining temporal logic with neural networks typically uses formulas as prior knowledge, constraints, or supervision \citep{2021_teaching_ltl_nn,2021_ltl2action}, rather than learning them.
In contrast, inductive logic programming (ILP), recently extended to temporal settings with differentiable variants \citep{2023_tilp,2024_tilr,2026_diff_temp_ilp}, learns logical rules but assumes relational data with predefined predicates.
We consider categorical sequences, where such relational structure is absent.
Closest to our setting are interpretable predictive models built directly from sequential data, which have been primarily studied in sequential rule mining and temporal point processes, while neuro-symbolic rule learning has so far focused largely on tabular data.

\textbf{Sequential rule mining} extracts human-interpretable dependencies from sequence data.
Most methods retrieve rules over ordered subsequences \citep{1997_freq_episodes,2002_association_rules_time_lags,2005_prediction_rule_mining} or unordered sets of \seqElementText{}s \citep{2012_cmrules,2014_erminer,2021_pfv_3,2021_pfv_4}.
Many of these approaches rely on user-specified frequency and confidence thresholds, often either returning an overwhelming number of rules or missing informative ones.
Methods based on the minimum description length \citep{2021_MDL_rules_no_gaps,2022_compress_freq_rule,2026_seqret} address rule selection by optimizing compression, but rely on combinatorial search over a super-exponential space.
Crucially, most types of sequential rules are not directly usable for next-\seqElementText{} prediction, as they neither specify the timing between \seqElementText{}s nor provide a principled way to combine multiple rules.

\textbf{Temporal point processes} (TPPs) model event timing by predicting when the next event occurs \citep{2003_tpp_book}.
While black-box neural TPPs achieve strong performance when jointly predicting event time and type \citep{2016_recurrent_mtpp,2023_contiformer}, existing interpretable TPPs focus on the temporal component.
For example, \citet{2020_logical_tpp} define intensity functions using logical formulas over past events, an approach extended to a differentiable neuro-symbolic framework for rule selection by \citet{2024_neuro_symbolic_tpps}.
These models explain \textit{when} events occur, whereas \ourMethodSideChannel{} targets interpretable next-\seqElementText{} type prediction by conditioning on the next event time.

\textbf{Neuro-symbolic models} that learn logical rules via gradient-based optimization have primarily been developed for tabular classification \citep{2020_MLLP,2021_RRL_MLLP_v2,2021_accurate_decision_rules_NN,2024_diffnaps}.
Closest to our approach is that of \citet{2025_neurules}, which combines differentiable binarization of real-valued inputs with a conjunction layer to induce rule lists.
These models assume fixed-length binary feature vectors, making them ill-suited for categorical sequence data, where flattening leads to sparse, high-dimensional representations.
Other approaches for temporal data make incompatible assumptions: for example, \citet{2025_diff_rule_sequences} rely on clustering consecutive segments of real-valued time series and do not support categorical \seqElementText{}s or rules with gaps, while \citet{2021_neural_ltl_learning} learn LTL formulas for binary classification of sequences without metric time and use a formula extraction that scales exponentially in the number of distinct \seqElementText{}s (propositions).
To the best of our knowledge, \ourMethodSideChannel{} \emph{is the first neuro-symbolic approach to perform next-event prediction on categorical sequences.}
 
\section{Experiments}
\label{sec:experiments}
We empirically evaluate \ourMethod{} and \ourMethodSideChannel{} on synthetic and real-world data.
As strong baselines with symbolic components that allow them to be decomposed into rules, we compare against time-agnostic variable-order Markov models based on the \textsc{PPM-C} algorithm \citep{1990_ppm_c} and gradient-boosted tree ensembles, including \textsc{XGBoost} \citep{2016_xgboost} and \textsc{CatBoost} \citep{2018_catboost}.
As a high-capacity black-box benchmark, we include a state-of-the-art transformer \citep{2017_attention_is_all_you_need} with rotary position embeddings (RoPE) \citep{2024_roformer}. 
We generalize RoPE to incorporate temporal differences ($\seqTime_\predInd - \seqTime_\seqInd$) instead of positional ones ($\predInd - \seqInd$).
We provide more details about the models and how we adapt them to the sequence setting in Appendix~\ref{app:sec_baseline_details}.
For synthetic data, we fix the hyperparameters of \ourMethod and the transformer, while for real data, we tune both models and \ourMethodSideChannel{} per dataset.
For \textsc{PPM}, \textsc{XGBoost}, and \textsc{CatBoost}, we always perform a dataset-specific grid search. 
We describe hyperparameter tuning and model selection in Appendix~\ref{app:sec_hyperparameters}.

\subsection{Synthetic Data}
\label{subsec:synthetic_data}
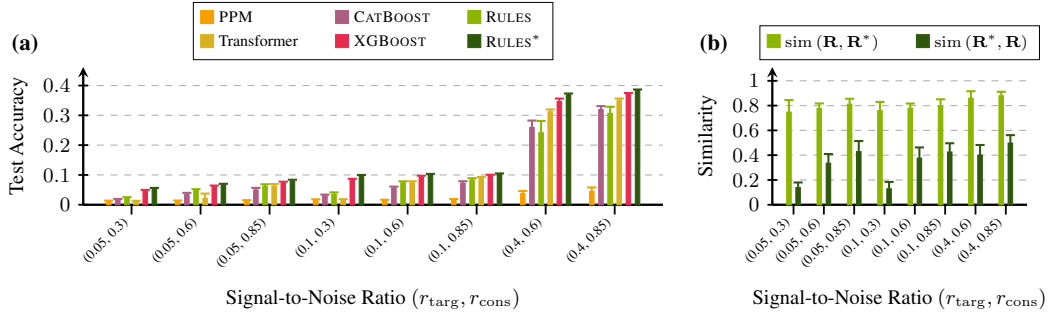
\begin{figure}[t]
    \centering
    \begin{tikzpicture}[scale=0.86, transform shape]
    
        \begin{axis}[
            width=10.5cm,
            height=3.7cm,
            clip=false,
            axis lines=left,
            axis line style={line width=1pt},
            x axis line style={-},
            tick style={line width=0.8pt, black},
            tick label style={font=\small},
            label style={font=\small},
            ymajorgrids,
            grid style={black!30,line width=0.5pt,dashed},
            ybar,
            ymin=0,
            enlarge x limits=0.1,
            enlarge y limits={upper=0.1},
            yticklabel style={/pgf/number format/fixed},
            xlabel={Signal-to-Noise Ratio $(\fracTarg, \fracCons)$},
            ylabel={Test Accuracy},
            xtick=data,
            xticklabels={},
            xticklabels={(0.05, 0.3),(0.05, 0.6),(0.05, 0.85),(0.1, 0.3),(0.1, 0.6),(0.1, 0.85),(0.4, 0.6),(0.4, 0.85)},
            xticklabel style={rotate=45, anchor=east, font=\tiny},
            ymax=0.42,
            ytick={0.0,0.1,0.2,0.3,0.4},
            scaled y ticks=false,
            legend columns=3,
            legend style={
                font=\scriptsize,
                legend cell align=left,
                at={(0.5,1.05)},
                anchor=south,
                /tikz/every even column/.append style={column sep=0.5cm}
            },
            legend image code/.code={\draw [#1] (0cm,-0.075cm) rectangle (0.2cm,0.125cm);}
        ]

            \addplot+[
                fill=orange, draw=orange, bar width=2pt, error bars/.cd, y dir=plus, y explicit, error mark=|,
                error bar style={draw=orange,line width=0.9pt}, error mark options={draw=orange,line width=0.9pt}
            ] coordinates {(0,0.0123) +- (0,0.0015) (1,0.0132) +- (0,0.0012) (2,0.0138) +- (0,0.0017) (3,0.0163) +- (0,0.0019) (4,0.0143) +- (0,0.0026) (5,0.0169) +- (0,0.0023) (6,0.0371) +- (0,0.0089) (7,0.0446) +- (0,0.0135)};

            \addplot+[
                fill=purple, draw=purple, bar width=2pt, error bars/.cd, y dir=plus, y explicit, error mark=|,
                error bar style={draw=purple,line width=0.9pt}, error mark options={draw=purple,line width=0.9pt}
            ] coordinates {(0,0.0166) +- (0,0.0030) (1,0.0339) +- (0,0.0061) (2,0.0496) +- (0,0.0070) (3,0.0303) +- (0,0.0038) (4,0.0559) +- (0,0.0050) (5,0.0723) +- (0,0.0066) (6,0.2595) +- (0,0.0230) (7,0.3193) +- (0,0.0116)};

            \addplot+[
                fill=green, draw=green, bar width=2pt, error bars/.cd, y dir=plus, y explicit, error mark=|,
                error bar style={draw=green,line width=0.9pt}, error mark options={draw=green,line width=0.9pt}
            ] coordinates {(0,0.0233) +- (0,0.0021) (1,0.0477) +- (0,0.0046) (2,0.0625) +- (0,0.0063) (3,0.0354) +- (0,0.0062) (4,0.0737) +- (0,0.0048) (5,0.0828) +- (0,0.0061) (6,0.2417) +- (0,0.0396) (7,0.3067) +- (0,0.0220)};

            \addplot+[
                fill=yellow, draw=yellow, bar width=2pt, error bars/.cd, y dir=plus, y explicit, error mark=|,
                error bar style={draw=yellow,line width=0.9pt}, error mark options={draw=yellow,line width=0.9pt}
            ] coordinates {(0,0.0119) +- (0,0.0014) (1,0.0211) +- (0,0.0164) (2,0.0630) +- (0,0.0066) (3,0.0169) +- (0,0.0018) (4,0.0740) +- (0,0.0048) (5,0.0897) +- (0,0.0036) (6,0.3124) +- (0,0.0074) (7,0.3507) +- (0,0.0056)};

            \addplot+[
                fill=red, draw=red, bar width=2pt, error bars/.cd, y dir=plus, y explicit, error mark=|,
                error bar style={draw=red,line width=0.9pt}, error mark options={draw=red,line width=0.9pt}
            ] coordinates {(0,0.0463) +- (0,0.0036) (1,0.0604) +- (0,0.0040) (2,0.0729) +- (0,0.0045) (3,0.0837) +- (0,0.0034) (4,0.0934) +- (0,0.0041) (5,0.0976) +- (0,0.0031) (6,0.3462) +- (0,0.0100) (7,0.3700) +- (0,0.0050)};

            \addplot+[
                fill=darkgreen, draw=darkgreen, bar width=2pt, error bars/.cd, y dir=plus, y explicit, error mark=|,
                error bar style={draw=darkgreen,line width=0.9pt}, error mark options={draw=darkgreen,line width=0.9pt}
            ] coordinates {(0,0.0555) +- (0,0.0010) (1,0.0661) +- (0,0.0042) (2,0.0783) +- (0,0.0056) (3,0.0985) +- (0,0.0015) (4,0.1020) +- (0,0.0016) (5,0.1038) +- (0,0.0013) (6,0.3686) +- (0,0.0046) (7,0.3854) +- (0,0.0013)};

            \legend{\textsc{PPM},\textsc{CatBoost},\ourMethod{},Transformer,\textsc{XGBoost},$\text{\ourMethod{}}^*$}
            \node at (axis cs:\pgfkeysvalueof{/pgfplots/xmin},\pgfkeysvalueof{/pgfplots/ymax}) [anchor=east, xshift=-16pt, yshift=10pt] {\textbf{(a)}};
        \end{axis}

        \begin{axis}[
            at={(10.6cm,0)},
            width=5.5cm,
            height=3.7cm,
            clip=false,
            axis lines=left,
            axis line style={line width=1pt},
            x axis line style={-},
            tick style={line width=0.8pt, black},
            tick label style={font=\small},
            label style={font=\small},
            ymajorgrids,
            grid style={black!30,line width=0.5pt,dashed},
            ybar,
            ymin=0,
            enlarge x limits=0.1,
            enlarge y limits={upper=0.1},
            yticklabel style={/pgf/number format/fixed},
            xlabel={Signal-to-Noise Ratio $(\fracTarg, \fracCons)$},
            ylabel={Similarity},
            xtick=data,
            xticklabels={},
            xticklabels={(0.05, 0.3),(0.05, 0.6),(0.05, 0.85),(0.1, 0.3),(0.1, 0.6),(0.1, 0.85),(0.4, 0.6),(0.4, 0.85)},
            xticklabel style={rotate=45, anchor=east, font=\tiny},
            ymax=1.01,
            ytick={0.0,0.2,0.4,0.6,0.8,1.0},
            scaled y ticks=false,
            legend columns=2,
            legend style={
                font=\scriptsize,
                legend cell align=left,
                at={(0.5,1.05)},
                anchor=south,
                /tikz/every even column/.append style={column sep=0.5cm}
            },
            legend image code/.code={\draw [#1] (0cm,-0.075cm) rectangle (0.2cm,0.125cm);}
        ]
        
            \addplot+[
                fill=green, draw=green, bar width=2pt, error bars/.cd, y dir=plus, y explicit, error mark=|,
                error bar style={draw=green,line width=0.9pt}, error mark options={draw=green,line width=0.9pt}
            ] coordinates {(0,0.7473) +- (0,0.0972) (1,0.7763) +- (0,0.0412) (2,0.8115) +- (0,0.0435) (3,0.7592) +- (0,0.0704) (4,0.7783) +- (0,0.0384) (5,0.7958) +- (0,0.0550) (6,0.8596) +- (0,0.0561) (7,0.8792) +- (0,0.0322)};

            \addplot+[
                fill=darkgreen, draw=darkgreen, bar width=2pt, error bars/.cd, y dir=plus, y explicit, error mark=|,
                error bar style={draw=darkgreen,line width=0.9pt}, error mark options={draw=darkgreen,line width=0.9pt}
            ] coordinates {(0,0.1386) +- (0,0.0411) (1,0.3354) +- (0,0.0739) (2,0.4289) +- (0,0.0851) (3,0.1290) +- (0,0.0554) (4,0.3763) +- (0,0.0862) (5,0.4260) +- (0,0.0701) (6,0.4009) +- (0,0.0813) (7,0.4981) +- (0,0.0635)};

            \legend{$\simRulesets\paren{\ruleset,\ruleset^*}$,$\simRulesets\paren{\ruleset^*,\ruleset}$}
            \node at (axis cs:\pgfkeysvalueof{/pgfplots/xmin},\pgfkeysvalueof{/pgfplots/ymax}) [anchor=east, xshift=-16pt, yshift=10pt] {\textbf{(b)}};
        \end{axis}

    \end{tikzpicture}
    \caption{
         \textit{\ourMethod{} reliably recovers predictive rules from synthetic data across varying signal-to-noise ratios.} 
         Results are averaged over 10 rule sets and corresponding datasets per setting; error bars indicate one empirical standard deviation.
        \textbf{(a)}~Test accuracy compared to baselines; $\text{\ourMethod{}}^*$ denotes \ourMethod{} with embedded ground-truth antecedents.
        \textbf{(b)}~Similarity between the rule set $\ruleset$ discovered by \ourMethod{} and the ground truth $\ruleset^*$. 
    }
    \label{fig:synthetic_noise}
\end{figure}

We first focus on \ourMethod{}, evaluating its ability to recover ground-truth structure from synthetic data.
We create datasets from ground-truth rule sets $\ruleset^*$ of size $20$. 
Unless stated otherwise, we use a vocabulary $\vocab$ of size $\vocabSize = 128$ and generate $\numSamples = 1024$ training sequences of length $\seqLen\in [96,128]$.
We use regularly sampled sequences ($\seqTime_{\seqInd+1} - \seqTime_\seqInd = 1$), and allow a maximum distance of $64$ between antecedent events and consequents. 
To generate rules, we sample the size of the antecedent from $[1,8]$, then choose its events uniformly from $\vocab$.
For each element, we create its delay interval $[\minOffset_i,\maxOffset_i]$ by uniformly sampling $\minOffset_i$ from $[1,63]$ and its width from $[1,10]$, ensuring that $\maxOffset_i\leq 64$.
We complete rules by adding a random single-event consequent.

After generating a rule set, we create sequences by filling about 30\% of their positions with uniformly sampled rules, then filling gaps with random noise events.
To avoid ambiguous ground truth, we restrict the training objective to consequent and noise events, preventing unintended sub-rules from antecedent placements.
Two signal-to-noise parameters govern task difficulty: the probability $\fracCons\in[0,1]$ that a rule antecedent is paired with its true consequent (rather than a random replacement), and the fraction $\fracTarg\in [0,1]$ of events in the training objective that are consequents rather than noise.

\textbf{Sanity Check.}
We first test whether \ourMethod{} learns spurious rules by applying it to 10 datasets with uniformly random \seqElementText{}s.
Using the same rule extraction as for other synthetic data (Appendices~\ref{app:subsec_rule_set_extraction} and \ref{app:subsec_antecedent_pruning}), the lowest-training-loss model over four random seeds per dataset correctly returns no~rules.

\textbf{Predictive Performance.}
Next, we evaluate prediction accuracy on \seqElementText{}s generated from sequential rules.
For each synthetic setting, we generate datasets from 10 rule sets, train the models, and report accuracy on held-out test data, with results averaged across rule sets.
As a reference, we include our model with ground-truth antecedents embedded in its hidden layers, which we denote by $\text{\ourMethod{}}^*$.

In \cref{fig:synthetic_noise}~(a), we present results for varying signal-to-noise ratios; results for varying dataset and vocabulary sizes are  in Appendix~\ref{app:sec_results_synthetic}.
\textsc{PPM} fails to capture long-range dependencies between antecedents and consequents, leading to poor predictive performance. The best model often uses exclusively a 0-order context, predicting the most frequent training event everywhere.
In contrast, \ourMethod{} achieves strong accuracy, performing close to more complex tree-based ensembles and the black-box transformer, even outperforming \textsc{CatBoost} and the transformer in noisy settings.
Moreover, the performance gap to the other models narrows with more training samples (Appendix~\ref{app:sec_results_synthetic}).

\textbf{Ground Truth Recovery.}
Next, we assess whether \ourMethod{} is faithful to the ground truth by extracting the rules learned in the previous experiment, simplifying theses rules as described in Appendix~\ref{app:subsec_symbolic_rule_refinement}, and comparing them to the ground-truth rule set.
To quantify similarity between rule sets, we follow a greedy matching strategy.
First, we define the similarity of two predicates $\alpha_i \defeq (\antElement_i, \delayInterval_i)$ as the Jaccard similarity of their delay intervals when their events match: $\simRules(\alpha_1, \alpha_2) \defeq \indicator(\antElement_1 = \antElement_2)\cdot \frac{\abs{\delayInterval_1 \cap \delayInterval_2}}{\abs{\delayInterval_1 \cup \delayInterval_2}}$. %
We then define the similarity of two rules $r_i = \ant_i\ruleTo \cons_i$ as
\begin{equation}
    \simRules_{\genericRule}(r_1,r_2) = \frac{\abs{\cons_1\cap\cons_2}}{\abs{\cons_1\cup\cons_2}}\cdot \simRules_\ant(\ant_1,\ant_2),\qquad 
    \simRules_\ant(\ant_1,\ant_2) = \frac{1}{\abs{\ant_1}} \sum_{\alpha_1\in \ant_1} \max_{\alpha_2\in \ant_2} \simRules(\alpha_1, \alpha_2) \label{eq:ant_sim}.
\end{equation}
The asymmetric rule similarity $\simRules_{\genericRule}(r_1,r_2)\in [0,1]$ measures how well rule $r_1$ can be mapped onto rule $r_2$.
To calculate the similarity of two  rule sets $\ruleset_1, \ruleset_2$, we use the same strategy as in \cref{eq:ant_sim} and match all rules in $\ruleset_1$ to the rules in $\ruleset_2$ that maximize $\simRules_{\genericRule}$, then take the mean.

As shown in \cref{fig:synthetic_noise}~(b), rules learned by \ourMethod faithfully map onto the ground truth ($\simRulesets\paren{\ruleset,\ruleset^*}$).
However, recovery of the full ground-truth structure ($\simRulesets\paren{\ruleset^*,\ruleset}$) is not perfect, because (i)~\ourMethod frequently approximates large antecedents with smaller ones, and (ii)~it occasionally misses rules.
Ground truth recovery is mainly driven by the reliability of consequents following antecedents ($\fracCons$), while \ourMethod{} remains highly robust to random noise events ($\fracTarg$).%

\subsection{Real-World Data}
\label{subsec:experiments_real}

Having verified that \ourMethod{} recovers ground-truth rules on synthetic data, we now evaluate \ourMethod{} and \ourMethodSideChannel{} on five diverse real-world datasets, ranging from musical compositions (Chorales) to e-commerce transactions (Amazon); dataset details are in Appendix~\ref{app:sec_real_world_dataset_stats}.
We use 60-20-20 train-validation-test splits, tuning hyperparameters based on the validation cross-entropy (Appendix~\ref{app:subsec_hyperparams_real}) and reporting test performance for the best models.
For the residual network in \ourMethodSideChannel{}, we use a transformer with the same architecture as the black-box baseline.
To encourage sparse rules, we extract the rules learned by \ourMethod{}, refine them (Appendix~\ref{app:subsec_symbolic_rule_refinement}), re-embed the refined antecedents into the hidden layers, and retrain only the output layer.
All reported results use this refined model. %
We analyze the effect of this post-processing loop in Appendix~\ref{app:subsec_effect_rule_processing} and variability of model performance in Appendix~\ref{app:subsec_variability}.
In Appendix~\ref{app:subsec_pattern_mining}, we further study the integration of pre-mined patterns into \ourMethod{}. %

\begin{table}[t]
    \centering
    \caption{
        \textit{Test accuracy (Acc.) in \% and number of rules ($\abs{\ruleset}$) on real-world datasets.}
        Best accuracy results are in \textbf{bold}, second best are \underline{underlined}.
        \ourMethodSideChannel{} performs close to the black-box transformer.
    }
    \label{tab:table_real_results}
    {\footnotesize
    \centering
    \begin{tabular}{l rr r rr rr rr r}
        \toprule
                            & \multicolumn{2}{c}{\ourMethod{}} & \multicolumn{1}{c}{\ourMethodSideChannel{}} & \multicolumn{2}{c}{\textsc{PPM}} & \multicolumn{2}{c}{\textsc{CatBoost}} & \multicolumn{2}{c}{\textsc{XGBoost}} & Transformer \\ 
        \cmidrule(lr){2-3} \cmidrule(lr){4-4} \cmidrule(lr){5-6}  \cmidrule(lr){7-8} \cmidrule(lr){9-10} \cmidrule(lr){11-11}
        \textbf{Datasets}    & $\abs{\ruleset}$   & Acc.                & Acc.                & $\abs{\ruleset}$   & Acc.                & $\abs{\ruleset}$   & Acc.                & $\abs{\ruleset}$   & Acc.                & Acc.                 \\  
        \midrule
        Holidays             &                26  &              97.57  &     \textbf{100.00} &                14  &     \textbf{100.00} &              1129  &     \textbf{100.00} &              1455  &     \textbf{100.00} &      \textbf{100.00} \\
        Chorales             &                33  &              21.00  &              37.00  &               180  &              31.40  &             11596  &              37.10  &             34462  &   \underline{37.72} &      \textbf{39.35}  \\
        Commits              &                51  &   \underline{19.60} &              18.31  &                80  &              18.11  &               380  &              15.55  &             83965  &              17.03  &      \textbf{20.10}  \\
        Rolling Mill         &                77  &              34.13  &   \underline{70.44} &              9583  &              67.64  &             13809  &              65.64  &            640723  &              68.84  &      \textbf{71.24}  \\
        Amazon               &                42  &               6.53  &   \underline{17.55} &               993  &              15.16  &             13123  &              11.79  &           5149683  &              15.37  &      \textbf{17.61}  \\
        \midrule
        Avg. Rank            &                    &               5.2   &    \underline{2.4}  &                    &                3.6  &                    &               4.0   &                    &               2.8   &       \textbf{1.0}   \\
        \bottomrule
    \end{tabular}}
\end{table}

\textbf{Predictive Performance.}
First, we report next-\seqElementText{} prediction accuracy on the test splits of the real-world datasets in \cref{tab:table_real_results}; cross-entropy results in Appendix~\ref{app:subsec_cross_entropy_and_model_sizes} show consistent trends.
\ourMethod{} alone performs competitively on Holidays and Commits, achieving accuracies close to the black-box transformer. 
This indicates that the underlying dynamics of these datasets can be 
well captured by a small number of \ruleName{}s.
On the remaining datasets, \ourMethod{} identifies meaningful structure through compact rule sets, and extending it with a residual network in \ourMethodSideChannel{} achieves performance competitive with the transformer, outperforming all other baselines and ranking second overall.
Next, we examine how interpretable the decisions of \ourMethod{} and \ourMethodSideChannel{} are.

\textbf{Model Interpretability.}
We assess the structural complexity of the models in \cref{tab:table_real_results}, using rule set and antecedent sizes (Appendix~\ref{app:subsec_cross_entropy_and_model_sizes}) as proxies for global interpretability, since smaller models are generally easier to reason about.
Following Appendix~\ref{app:sec_interpreting_models_as_rules}, we treat \textsc{PPM} and tree ensembles as sequential rule sets: for \textsc{PPM}, non-zero-order Markov contexts define rule antecedents, with empirical next-\seqElementText{} distributions as consequents; for trees, each root-to-leaf path forms a conjunctive antecedent, with class logits at the leaf specifying the consequent.
For \textsc{PPM}, we measure antecedent size by context length and count unique contexts observed during training as rules.
For trees, we measure antecedent size by leaf depth and define the rule count as the number of distinct leaf regions. %

Across datasets, \ourMethod{} produces compact, highly interpretable models, typically consisting of only tens of rules with antecedents of size 1--5.
Tree ensembles, while limited in depth, often induce thousands of distinct symbolic conditions, making exhaustive inspection impractical---especially for \textsc{XGBoost}, which builds separate trees for each class.
Even individual trees are hard to interpret, as they induce complex logic over categorical features and do not explicitly tie event-time features to events (Appendices~\ref{app:sec_interpreting_models_as_rules} and~\ref{app:subsec_qualitative_examples}).
\textsc{PPM} performs strongly and produces simple models when short contexts suffice, but its rule set grows rapidly with vocabulary size and context length, as observed on Rolling Mill.
Beyond compactness, the rules learned by \ourMethod{} are semantically meaningful, as demonstrated by our qualitative analysis and comparison to baseline models in Appendix~\ref{app:subsec_qualitative_examples}.
For example, on Amazon, \ourMethod{} recovers a rule similar to our running example, which \textsc{PPM} cannot capture due to its lack of temporal modeling.

Overall, \ourMethod{} provides directly inspectable models, whereas competing methods require additional rule selection (e.g., pruning) to achieve comparable model interpretability. 
As shown in Appendix~\ref{app:subsec_effect_rule_processing}, individual predictions of \ourMethod{} are also easy to explain, typically involving only 0--4 active rules, with traceable additive residual contributions in \ourMethodSideChannel{}.

\begin{figure}[t]
\centering
\begin{minipage}{0.43\linewidth}
    \centering
    \captionof{table}{
        \textit{\ourMethodSideChannel{} corrects errors (\ding{55}) but rarely flips correct predictions (\checkmark) of \ourMethod{}.}
        Prediction agreement is averaged over test splits of real datasets; per-dataset results are in Appendix~\ref{app:subsec_confusion_side_channel_individual}.
        Counts are normalized within rule-activity groups.
    }
    \label{tab:confusion_side_channel_aggregated}
    \resizebox{1.0\linewidth}{!}{\footnotesize
    \setlength{\tabcolsep}{8pt}
    \begin{tabular}{c rr rr}
        \toprule
                         & \multicolumn{4}{c}{\ourMethod{}} \\
        \cmidrule(lr){2-5}
        \ourMethodSideChannel{}     & \multicolumn{2}{c}{Rule Active} & \multicolumn{2}{c}{No Rule Active}\\
        \cmidrule(lr){2-3} \cmidrule(lr){4-5}
                         & \multicolumn{1}{c}{\checkmark} & \multicolumn{1}{c}{\ding{55}} & \multicolumn{1}{c}{\checkmark} & \multicolumn{1}{c}{\ding{55}}  \\
        \midrule
        \checkmark       &  0.40 &  0.17 &  0.19 &  0.36  \\
        \ding{55}        &  0.03 &  0.40 &  0.03 &  0.42  \\
        \midrule
        Data Frac. & \multicolumn{2}{c}{0.64} & \multicolumn{2}{c}{0.36} \\
        \bottomrule
    \end{tabular}}
\end{minipage}
\hfill
\begin{minipage}{0.55\linewidth}
    \centering
    \resizebox{1.0\linewidth}{!}{
    \begin{tikzpicture}
    
        \begin{axis}[
            width=9cm,
            height=3.7cm,
            clip=false,
            axis lines=left,
            axis line style={line width=1pt},
            x axis line style={-},
            tick style={line width=0.8pt, black},
            tick label style={font=\small},
            label style={font=\small},
            ymajorgrids,
            grid style={black!30,line width=0.5pt,dashed},
            ybar,
            ymin=0,
            enlarge x limits=0.15,
            enlarge y limits={upper=0.1},
            yticklabel style={/pgf/number format/fixed},
            xlabel={Binned Symmetric KL, \ourMethod vs. \ourMethodSideChannel},
            ylabel={Bin Probability},
            xtick=data,
            xticklabels={{[0.0, 0.7]},{[0.7, 1.5]},{[1.5, 2.2]},{[2.2, 2.9]},{[2.9, 3.7]},{[3.7, 4.4]},{[4.4, 5.1]},{[5.1, 5.9]}},
            xticklabel style={rotate=30, anchor=east, font=\tiny},
            ymax=1,
            ytick={0.0,0.2,0.4,0.6,0.8,1.0},
            scaled y ticks=false,
            legend columns=5,
            legend style={
                font=\scriptsize,
                legend cell align=left,
                at={(0.45,1.05)},
                anchor=south,
                /tikz/every even column/.append style={column sep=0.5cm}
            },
            legend image code/.code={\draw [#1] (0cm,-0.055cm) rectangle (0.15cm,0.095cm);}
        ]
        
            \addplot+[
                fill=darkgreen, draw=darkgreen, bar width=2pt
            ] coordinates {(0,0.7784)   (1,0.4353)   (2,0.3213)   (3,0.2772)   (4,0.0866)   (5,0.0383)   (6,0.0144)   (7,0.0000)  };

            \addplot+[
                fill=green, draw=green, bar width=2pt
            ] coordinates {(0,0.0437)   (1,0.1959)   (2,0.3062)   (3,0.4346)   (4,0.6699)   (5,0.7883)   (6,0.8993)   (7,1.0000)  };

            \addplot+[
                fill=red, draw=red, bar width=2pt
            ] coordinates {(0,0.0418)   (1,0.0647)   (2,0.0409)   (3,0.0120)   (4,0.0077)   (5,0.0000)   (6,0.0000)   (7,0.0000)  };

            \addplot+[
                fill=blue, draw=blue, bar width=2pt
            ] coordinates {(0,0.0991)   (1,0.1466)   (2,0.0794)   (3,0.0315)   (4,0.0038)   (5,0.0000)   (6,0.0000)   (7,0.0000)  };

            \addplot+[
                fill=orange, draw=orange, bar width=2pt
            ] coordinates {(0,0.0369)   (1,0.1575)   (2,0.2522)   (3,0.2448)   (4,0.2320)   (5,0.1734)   (6,0.0863)   (7,0.0000)  };

\legend{\checkmark\checkmark,\protect\ding{55}$\rightarrow$\checkmark,\checkmark$\rightarrow$\ding{55},\ding{55}\ding{55},\ding{55}$\rightarrow$\ding{55}}
        \end{axis}

    \end{tikzpicture}}
    \captionof{figure}{
        \textit{Unsupervised metrics help identify correct residual changes when rules are active.}
        Rolling Mill: Probability of \ourMethod{} being correct and how \ourMethodSideChannel{} changes ($\to$) predictions on the test split, grouped by KL (e.g., \ding{55}\ding{55} = models agree on a wrong prediction).
    }
    \label{fig:side_channel_rolling_mill}
\end{minipage}
\end{figure}

\textbf{Effect of the Residual Network.}
Finally, we analyze how the residual network in \ourMethodSideChannel{} complements the predictions of \ourMethod{}, excluding Commits, where the residual network does not improve performance.
\cref{tab:confusion_side_channel_aggregated} shows that at least one rule is active for 64\% of test predictions on average (97\% on Holidays, 55\% on Chorales, 88\% on Rolling Mill, 17\% on Amazon),\!%
\footnote{On Commits, at least one rule is active for 57\% of test predictions.}
so a symbolic \ourMethod{} explanation is available for a substantial fraction of cases.
The residual network rarely overturns correct rule-based predictions (3\% of active-rule cases) while correcting many errors (17\%), 
leading to a simple decision rule: \emph{trust the rules when the rule and residual networks agree and incorporate the residual logits otherwise}.
This preserves interpretability when rules are reliable~(40\% of active-rule cases) and leverages black-box residuals only when necessary.
More broadly, \ourMethodSideChannel{} offers an explicit \textbf{accuracy--interpretability trade-off}: users may rely purely on \ourMethod{} for interpretability or include residual logits for high accuracy.

The additive rule and residual logits also directly reveal contributions of active rules and how residuals modify predictions, with symmetric KL divergence and entropy providing practical signals for distinguishing corrective from spurious residual changes (Appendix~\ref{app:subsec_influence_side_channel_individual_predictions}).
For example, on Rolling Mill when rules are active and the residual network changes predictions ($\to$ cases in \cref{fig:side_channel_rolling_mill}), incorrect updates are rare for KL $>3$ but common for KL $<0.7$.
This suggests favoring rules in the low-KL regime, trading off some errors for interpretability, while high KL indicates that rules are likely incorrect.

\section{Conclusion}
\label{sec:conclusion}
We introduced \ourMethod{}, a neuro-symbolic model that learns interpretable rules for next-event prediction through continuous optimization.
Its core contribution is integrating sequential rules into the layers of a neural network, where event- and delay conditions are learned end-to-end to form rule antecedents.
To capture structure beyond rules, \ourMethodSideChannel{} augments \ourMethod with a residual network that corrects errors in rule-based predictions, while maintaining rule-level interpretability via additive rule and residual logits.
On synthetic data, \ourMethod faithfully recovers the ground truth, producing strong predictions driven by symbolic rules alone.
On real-world data, it discovers compact, interpretable rule sets that, when combined with the residual network in \ourMethodSideChannel, achieve competitive performance.

\textbf{Limitations.}
 \ourMethod learns rules differentiably in a neural network, avoiding a costly combinatorial search but introducing randomness and several hyperparameters, such as layer sizes and temperature schedules. 
As with any neural model, this requires tuning, and learned models---and thus extracted rule sets---can vary across runs.
Our differentiable approximations build on prior neuro-symbolic models and are primarily guided by empirical success; future work could further improve results through a more principled analysis of these functions.
Finally, our architecture currently models one categorical event per time step and forms Temporal Conjunctive Prediction Rules via conjunctions over event-time predicates. 
This naturally suggests extensions to richer data types and other classes of rules capturing different structural patterns.

\begin{ack}
This work was funded by DFG grant 389792660 as part of TRR~248 -- CPEC, see \url{https://perspicuous-computing.science}.
Tim Bauerschmidt was supported by the Konrad Zuse School of Excellence in Learning and Intelligent Systems (ELIZA) through the DAAD programme Konrad Zuse Schools of Excellence in Artificial Intelligence, sponsored by the Federal Ministry of Education and Research.
We gratefully acknowledge the stimulating research environment of the GRK 2853/1 ``Neuroexplicit Models of Language, Vision, and Action'', funded by the Deutsche Forschungsgemeinschaft (DFG, German Research Foundation) under project number 471607914.
The views and opinions expressed in this publication are those of the author(s) alone and do not necessarily reflect those of the aforementioned funding agencies and institutions. 
None of the aforementioned parties can be held responsible for them.
\end{ack}

{
\small
\bibliographystyle{plainnat}
\bibliography{paper}
}

\clearpage
\appendix

\section{Convergence Proofs for Temperature Annealing}
\label{app:sec_convergence_proofs}

\begin{figure}[hb]
\hypersetup{linkcolor=black}
\centering

\ref{sharedlegend}
\vspace{0.2cm}

\begin{subfigure}{0.31\linewidth}
    \begin{tikzpicture}[scale=0.83, transform shape]
        \begin{axis}[
            width=5.2cm,
            height=4cm,
            xlabel={$x$},
            ylabel={$\sigma(\frac{x}{\tau})$},
            axis lines=left,
            axis line style={line width=1pt},
            tick style={line width=0.8pt, black},
            grid=both,
            xmin=-4, xmax=4,
            ymin=0, ymax=1.1,
            xtick={-2,0,2},
            ytick={0,0.25,0.5,0.75,1},
            legend to name=sharedlegend,
            legend style={
                legend columns=3,
                font=\small,
                /tikz/every even column/.append style={column sep=0.5cm}
            }
        ]
            \addplot[green, line width=1.2pt, domain=-4:4, samples=500] {1/(1 + exp(-x))};
            \addlegendentry{$\tau = 1.0$};
            \addplot[blue, line width=1.2pt, domain=-4:4, samples=500] {1/(1 + exp(-x/0.3))};
            \addlegendentry{$\tau = 0.3$}
            \addplot[blue, dashed, line width=1.2pt, forget plot, domain=-4:4, samples=500] {1/(1 + exp(-x/0.3 + ln(0.3)))};
            \addplot[orange, line width=1.2pt, domain=-4:4, samples=500] {1/(1 + exp(-x/0.01 + ln(0.01)))};
            \addlegendentry{$\tau = 0.03$}
        \end{axis}
    \end{tikzpicture}
    \caption{
        Temperature-scaled sigmoid function.
    }
\end{subfigure}
\hfill
\begin{subfigure}{0.31\linewidth}
    \begin{tikzpicture}[scale=0.83, transform shape]
        \begin{axis}[
            width=5.2cm,
            height=4cm,
            xlabel={$x$},
            ylabel={$\softInterval(x;a,b,\tau)$},
            axis lines=left,
            axis line style={line width=1pt},
            tick style={line width=0.8pt, black},
            grid=both,
            xmin=0, xmax=8,
            ymin=0, ymax=1.1,
            xtick={2,4,6},
            ytick={0,0.25,0.5,0.75,1},
        ]
            \addplot[green, line width=1.2pt, domain=0:8, samples=500] {exp(2*x - 2)/(exp(x) + exp(2*x - 2) + exp(3*x - 8))};
            \addplot[blue, line width=1.5pt, domain=0:8, samples=500] {exp(1/0.3 * (2*x - 2))/(exp(1/0.3 * x) + exp(1/0.3 * (2*x - 2)) + exp(1/0.3 * (3*x - 8)))};
            \addplot[orange, line width=1.5pt, domain=0:8, samples=500] {exp(1/0.01 * (2*x - 2))/(exp(1/0.01 * x) + exp(1/0.01 * (2*x - 2)) + exp(1/0.01 * (3*x - 8)))};
        \end{axis}
    \end{tikzpicture}
    \caption{
        Temperature-scaled interval membership function.
    }
\end{subfigure}
\hfill
\begin{subfigure}{0.31\linewidth}
    \begin{tikzpicture}[scale=0.83, transform shape]
        \begin{axis}[
            width=5.2cm,
            height=4cm,
            xlabel={$x$},
            ylabel={$\softplus(x;\tau)$},
            axis lines=left,
            axis line style={line width=1pt},
            tick style={line width=0.8pt, black},
            grid=both,
            xmin=-4, xmax=4,
            ymin=0, ymax=4.4,
            xtick={-2,0,2},
            ytick={0,1,2,3,4},
        ]
            \addplot[green, line width=1.2pt, domain=-4:4, samples=500] {ln(1 + exp(x))};
            \addplot[blue, line width=1.2pt, domain=-4:4, samples=500] {0.3 * ln(1 + exp(x/0.3))};
            \addplot[orange, line width=1.2pt, domain=-4:4, samples=500] {0.01 * ln(1 + exp(x/0.01))};
        \end{axis}
    \end{tikzpicture}
    \caption{
        Temperature-scaled softplus function.
    }
\end{subfigure}

    \caption{
        \textit{Illustration of temperature annealing for different temperature-scaled functions.}
        \textbf{(a)}~The sigmoid function $\sigma(x/\tau)$ converges to the Heaviside step function as $\tau \to 0^+$. The dashed curve shows a version with temperature-dependent bias: $\sigma(x/\tau - \log(\tau))$ as used in \cref{eq:ant_act}.
        \textbf{(b)}~The soft interval membership function $\softInterval(x;a,b,\tau)$ converges to the hard interval function $\indicator(x \in [a,b])$, shown here with $a=2$ and $b=6$.
        \textbf{(c)}~The temperature-scaled softplus function $\tau \cdot \log(1 + \exp(x/\tau))$ converges to the ReLU function $\max(0,x)$.
    }
    \label{fig:examples_temp_annealing}
\end{figure}
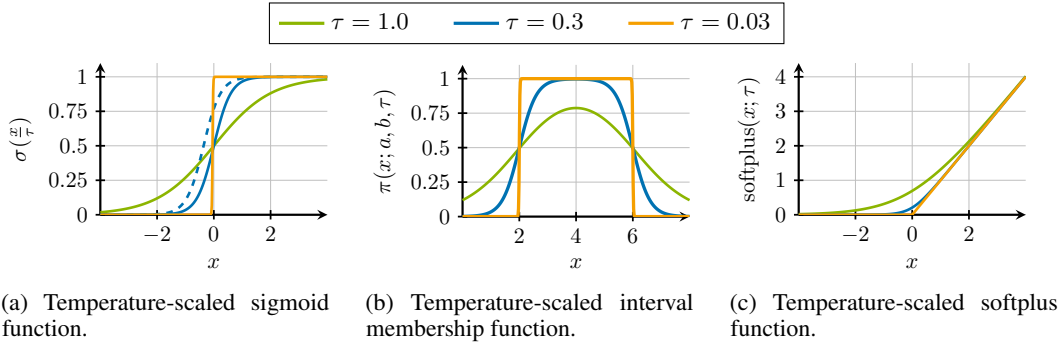

Before presenting formal convergence proofs for the temperature-dependent activation functions used in \ourMethod{}, we first provide illustrative examples in \cref{fig:examples_temp_annealing}.
Each function is smooth at $\tau = 1$, and as $\tau \to 0^+$, it converges to its corresponding hard function.

\subsection{Basic Functions}
\label{app:subsec_temp_annealing_basic}

\paragraph{Sigmoid Function.}
The temperature-controlled sigmoid function (used in \cref{eq:event_act,eq:ant_act}) converges to the Heaviside step function as the temperature approaches zero:
\begin{equation}
    \lim_{\tau\to 0^+}\sigma\paren{x;\tau} = \lim_{\tau\to 0^+} \frac{1}{1+\exp\paren{-\frac{x}{\tau}}} = \begin{cases}
        \frac{1}{1+0}, & \text{if } x > 0 \\
        \frac{1}{1+1}, & \text{if } x = 0 \\
        \frac{1}{\lim_{\tau\to 0^+} \exp\paren{-\frac{x}{\tau}}}, & \text{if } x < 0
    \end{cases} = \begin{cases}
        1, & \text{if } x > 0 \\
        0.5, & \text{if } x = 0 \\
        0, & \text{if } x < 0
    \end{cases}.
\end{equation}

\paragraph{Softplus Function.}
\label{app:par_softplus_convergence}
The temperature-controlled softplus function converges to the ReLU function as the temperature approaches zero:
\begin{align}
    \lim_{\tau\to 0^+} \softplus\paren{x;\tau} = \lim_{\tau\to 0^+} \tau\cdot \log\paren{1+\exp\paren{\frac{x}{\tau}}} &= \begin{cases}
        \lim_{\tau\to 0^+} \tau\cdot \frac{x}{\tau}, & \text{if } x > 0 \\
        0\cdot \log(2), & \text{if } x = 0 \\
        0\cdot \log(1), & \text{if } x < 0
    \end{cases}\\ 
    & = \max(0,x).
\end{align}

\paragraph{LogSumExp.}
\label{app:par_lse_convergence}
For the LogSumExp function (\cref{eq:event_act}), consider an input vector $x = (x_1, x_2, \dots, x_d)\in \RR^d$ and denote $x^* = \max(x_1, x_2, \dots, x_d)$.
For $\tau > 0$, observe that
\begin{align}
    \max(x_1, x_2, \dots, x_d) &= \tau\cdot\log\paren{\exp\paren{\frac{x^*}{\tau}}}\\
    & \leq \tau\cdot \log\paren{\sum_{i=1}^d \exp\paren{\frac{x_i}{\tau}}} && \text{(LogSumExp)} \\
    & \leq \tau\cdot \log\paren{d\cdot \exp\paren{\frac{x^*}{\tau}}}\\
    & = \max(x_1, x_2, \dots, x_d) + \tau\cdot \log(d),
\end{align}
where the inequalities follow from the monotonicity of the logarithm.
Applying the squeeze theorem immediately gives the convergence
\begin{equation}
    \lim_{\tau\to 0^+}\lse\paren{x_1, x_2,\dots, x_d; \tau} = \max\paren{x_1, x_2,\dots, x_d}.
\end{equation}

\paragraph{Softmax and Softmax-Weighted Average.}
\label{app:par_softmax_convergence}
For the temperature-scaled softmax function, consider an input vector $x = (x_1, x_2, \dots, x_d)\in\RR^d$ and denote $i^* = \argmax_i x_i$.
Rewriting the softmax with respect to $x_{i^*}$ gives
\begin{equation}
    \lim_{\tau\to 0^+}\softmax_i\paren{x_1, x_2, \dots, x_d; \tau} = \lim_{\tau\to 0^+} \frac{\exp\paren{\frac{x_i}{\tau}}}{\sum_{j=1}^{d} \exp\paren{\frac{x_j}{\tau}}} = \lim_{\tau\to 0^+} \frac{\exp\paren{\frac{x_i-x_{i^*}}{\tau}}}{1 +\sum_{j\neq i^*} \exp\paren{\frac{x_j - x_{i^*}}{\tau}}}.
\end{equation}

We analyze two cases:
\begin{enumerate}
    \item \textit{Non-maximal elements:} If $x_i < x_{i^*}$, the numerator vanishes in the limit, so the corresponding softmax entry converges to $0$, i.e.,
    \begin{equation}
        \lim_{\tau\to 0^+} \frac{\exp\paren{\frac{x_i-x_{i^*}}{\tau}}}{1 +\sum_{j\neq i^*} \exp\paren{\frac{x_j - x_{i^*}}{\tau}}} = 0.
    \end{equation}
    \item \textit{Maximal elements:} If $x_i = x_{i^*}$, then the softmax converges to $1$ divided by the number of maximal elements:
    \begin{align}
        \lim_{\tau\to 0^+} \frac{\exp\paren{\frac{x_i-x_{i^*}}{\tau}}}{1 +\sum_{j\neq i^*} \exp\paren{\frac{x_j - x_{i^*}}{\tau}}} 
        & = \lim_{\tau\to 0^+} \frac{1}{1 +\sum_{j\neq i^*} \exp\paren{\frac{x_j - x_{i^*}}{\tau}}} \\
        & = \frac{1}{\abs{\set{x_i\mid 1\leq i\leq d: x_{i^*} = x_i}}}.
    \end{align}
    
\end{enumerate}
Thus, when the input has a \textit{unique} maximum, the temperature-scaled softmax converges to a one-hot encoding of the argmax.
Denoting $n = \abs{\set{x_i\mid 1\leq i\leq d: x_{i^*} = x_i}}$ as the number of maximal elements, this also establishes the convergence of the softmax-weighted average (\cref{eq:pred_act}) to the true maximum:
\begin{equation}
   \lim_{\tau\to 0^+}\sum_{i=1}^d x_i \cdot \frac{\exp\paren{\frac{x_i}{\tau}}}{\sum_{j=1}^d \exp\paren{\frac{x_j}{\tau}}} = \sum_{i\in \set{j\mid 1\leq j\leq d: x_j = x_{i^*}}} x_i \cdot \frac{1}{n} = n\cdot \frac{x_{i^*}}{n} = x_{i^*}.
\end{equation}

\subsection{\ourMethod{}: Predicate Layer}
\label{app:subsec_temp_annealing_predicate}
In this section, we analyze the convergence of the predicate layer (\cref{eq:event_act,eq:delay_act,eq:pred_act}) under temperature annealing.
We have already shown in Appendix~\ref{app:par_softmax_convergence} that the softmax-based approximation of the maximum, used to aggregate $\tilde{\predAct}_i$ across all time steps (\cref{eq:pred_act}), converges to the true maximum function and thus accurately models a disjunction (logical OR).
We now turn our attention to the convergence of the soft \seqElementText{} match and the soft interval membership function.

\paragraph{Soft \SeqElementText{} Match.}
\label{app:par_temp_annealing_event_match}
We prove the convergence of the soft \seqElementText{} match (\cref{eq:event_act}) by simplifying notation: let $w = \firstWeights_i$, $w_j = \firstWeights_{i,\seqElement_\seqInd}$, and $\tau = \tau_1$.
Our goal is to show that the function activates primarily when $w_j = \firstWeights_{i,\seqElement_\seqInd}$ is positive and equal to the maximum element, effectively implementing the predicate $\mathbb{I}(j = \argmax_i w_i) = \mathbb{I}(w_j = \max_i w_i)$, which is equivalent to $\mathbb{I}(\seqElement_\seqInd = \argmax_d \firstWeights_{i,d})$ in the main body.
Formally, let $w^* = \max_i w_i$ and denote $n = \abs{\set{w_i\mid 1\leq i\leq d: w^* = w_i}}$ the number of maximal elements. 
We then prove
\begin{equation}
    \begin{aligned}
        & \lim_{\tau\to 0^+}  \sigma\paren{\frac{1}{\tau^2}\cdot \paren{w_j - (1-\tau)\cdot \lse\paren{w ; \tau}}}\\
        & \quad = \begin{cases}
        0, & \text{if } w^* < 0 \text{ or } w_j < w^* \text{ or } w_j = w^* < \log\paren{n}\\
        1, & \text{if } w_j = w^* > \log\paren{n} \\
        \frac{1}{1 + 1/n}, & \text{otherwise}
    \end{cases}.
    \end{aligned}
\end{equation}

In words, the soft \seqElementText{} match evaluates to $0$ in the limit if the maximum weight is negative, if the selected weight $w_j$ is smaller than the maximum, or if $w_j$ is maximal but smaller than the logarithm of the number of maximal elements.
It evaluates to $1$ if $w_j$ is maximal, positive, and larger than this logarithm.
In the edge case $w_j = w^* = \log\paren{\abs{\set{w_i\mid 1\leq i\leq d: w^* = w_i}}} = \log(n)$, the sigmoid converges to $\frac{1}{1 + 1/n}$.
Deactivating the sigmoid for $w^* < 0$ is desirable, as it allows \ourMethod{} to disable unnecessary neurons by assigning negative values to all their weights ($w = \firstWeights_i$).
The other edge cases, where the sigmoid does not converge to $1$ despite $w_j = w^*$, are undesirable; however, they are extremely unlikely in practice, since gradient descent would need to produce a maximum element exactly equal to $w^* = 0$ or multiple maximal elements with identical values.

To analyze convergence more precisely, it is useful to consider the argument of the sigmoid.
From the LogSumExp convergence (Appendix~\ref{app:par_lse_convergence}), we know that for positive temperatures $\tau > 0$:
\begin{equation}
    w^* \leq \lse\paren{w_1, \dots, w_d ;\tau} \leq w^* + \tau\cdot \log(d).
\end{equation}
Consequently, we have 
\begin{equation}
    \begin{aligned}
        \frac{1}{\tau^2}\cdot \paren{w_j - (1-\tau)\cdot w^*} & \geq \frac{1}{\tau^2}\cdot \paren{w_j - (1-\tau)\cdot \lse\paren{w ; \tau}}.
    \end{aligned}
\end{equation}
Since the sigmoid function is monotonically increasing, this inequality is preserved when it is applied to both sides.

For the convergence, we consider two cases for $w_j$:
\begin{enumerate}
    \item $w_j < w^*$: In this case, $w_j - w^* < 0$ and the upper bound diverges to $-\infty$, i.e.,
    \begin{align}
        \lim_{\tau\to 0^+} \frac{1}{\tau^2}\cdot \paren{w_j - (1-\tau)\cdot w^*} & = \lim_{\tau\to 0^+} \frac{1}{\tau^2} \cdot \paren{w_j - w^* + \tau w^*} \\
        & = \lim_{\tau\to 0^+} \frac{1}{\tau^2} \cdot \paren{w_j - w^*} = -\infty.
    \end{align}
    Since the upper bound diverges to $-\infty$, the LogSumExp-term must also diverge to $-\infty$,
    \begin{equation}
        \lim_{\tau\to 0^+} \frac{1}{\tau^2}\cdot \paren{w_j - (1-\tau)\cdot \lse\paren{w ; \tau}} = -\infty,
    \end{equation}
    which implies that the sigmoid converges to $0$:
    \begin{equation}
        \lim_{\tau\to 0^+} \sigma\paren{\frac{1}{\tau_1^2}\cdot \paren{w_j - (1-\tau_1)\cdot \lse\paren{w ; \tau}}} = 0.
    \end{equation}

    \item $ w_j = w^*$: In this case, we get the the upper bound as
    \begin{equation}
        \lim_{\tau\to 0^+} \frac{1}{\tau^2}\cdot \paren{w_j - (1-\tau)\cdot w^*} = \lim_{\tau\to 0^+} \frac{1}{\tau^2} \cdot \paren{\tau w^*} = \begin{cases}
             \infty, & \text{if } w^* > 0 \\
             0, & \text{if } w^* = 0 \\
            -\infty, & \text{otherwise}
        \end{cases}.
    \end{equation}
    So, if the maximum weight is negative ($w^* <0$), the sigmoid converges to $0$, analogous to the first case. 
    This shows that \ourMethod{} can disable predicate neurons by setting all of its weights ($w = \firstWeights_i$) to a negative value. 
    Now we consider the activation in case the maximum weight is non-negative ($w^*\geq 0$):
    {\begin{align}
        &\lim_{\tau\to 0^+} \frac{1}{\tau^2}\cdot \paren{w_j - (1-\tau)\cdot \lse\paren{w ; \tau}} \\
        & = \lim_{\tau\to 0^+} \frac{1}{\tau^2}\cdot \paren{w_j - (1-\tau)\cdot \tau \log\paren{\sum_{i=1}^d \exp\paren{\frac{w_i}{\tau}}}} \\
        & = \lim_{\tau\to 0^+} \frac{1}{\tau^2}\cdot \paren{w^* - (1-\tau)\cdot \paren{w^* + \tau \log\paren{\sum_{i=1}^d \exp\paren{\frac{w_i - w^*}{\tau}}}}} \\ 
        & = \lim_{\tau\to 0^+} \frac{1}{\tau^2}\cdot \paren{\tau w^* - (1-\tau)\cdot \paren{\tau \log\paren{\sum_{i=1}^d \exp\paren{\frac{w_i - w^*}{\tau}}}}} \\ \displaybreak
        & = \lim_{\tau\to 0^+} \frac{1}{\tau}\cdot \paren{w^* - \log\paren{\sum_{i=1}^d \exp\paren{\frac{w_i - w^*}{\tau}}} + \tau \log\paren{\sum_{i=1}^d \exp\paren{\frac{w_i - w^*}{\tau}}}} \\
        & = \lim_{\tau\to 0^+} \frac{1}{\tau}\cdot \paren{w^* - \log\paren{n} + \tau \log\paren{n}} \\
        & = \begin{cases}
            \lim_{\tau\to 0^+} \frac{1}{\tau}\cdot \paren{\tau \log\paren{n}}, & \text{if } w^* = \log\paren{n}\\
            \lim_{\tau\to 0^+} \frac{1}{\tau}\cdot \paren{w^* - \log\paren{n}}, & \text{otherwise}
        \end{cases}\\
        & = \begin{cases}
            \infty, & \text{if } w^* > \log\paren{n} = \log\paren{\abs{\set{w_i\mid 1\leq i\leq d: w^* = w_i}}} \\
            \log(n), & \text{if } w^* = \log\paren{n} = \log\paren{\abs{\set{w_i\mid 1\leq i\leq d: w^* = w_i}}} \\
            -\infty, & \text{otherwise}
        \end{cases}.
    \end{align}}%
    So this gives us the desired behavior in all cases when the maximum element is larger than the logarithm of the number of maximal elements.
    Hence, when $w^*$ is a unique maximum, this behavior is guaranteed for a positive weight since then
    \begin{equation}
        \log\paren{\abs{\set{w_i\mid 1\leq i\leq d: w^* = w}}} = \log(1) = 0.
    \end{equation}
    In summary, for $w_j = w^* > \log\paren{\abs{\set{w_i\mid 1\leq i\leq d: w^* = w_i}}}$, we have 
    \begin{equation}
        \lim_{\tau\to 0^+} \sigma\paren{\frac{1}{\tau_1^2}\cdot \paren{w_j- (1-\tau_1)\cdot \lse\paren{w ; \tau}}} = 1.
    \end{equation}

\end{enumerate}
Note that in the second case, the scaling factor $\frac{1}{\tau^2}$ is necessary to ensure proper convergence of the sigmoid to the desired limit values.

\paragraph{Soft Interval Membership.}
\label{app:par_temp_annealing_interval_membership}
From the convergence of the softmax function in Appendix~\ref{app:par_softmax_convergence}, we can directly establish the convergence of the soft interval membership function (repeated from \cref{eq:delay_act})
\begin{equation}
    \softInterval\paren{x; a, b, \tau} = \frac{\exp\paren{\frac{1}{\tau}\cdot \paren{2x - a}}}{\exp\paren{\frac{1}{\tau}\cdot x} + \exp\paren{\frac{1}{\tau}\cdot \paren{2x - a}} + \exp\paren{\frac{1}{\tau}\cdot \paren{3x - a - b}}}
\end{equation}
to
\begin{equation}
    \lim_{\tau\to 0^+} \softInterval\paren{x; a, b, \tau} = \begin{cases}
        1, & \text{if } a < x < b \\
        1/2, & \text{if either }  x = a \text{ or }  x = b \\
        1/3, &  \text{if } x = a = b\\
        0, & \text{otherwise}
    \end{cases}.
\end{equation}
Here we assume that $a\leq b$.
Observe that $\softInterval$ is the second coordinate of a temperature-scaled softmax applied to the logits $x$, $2x-a$, and $3x-a-b$, i.e.,
\begin{equation}
    \softInterval\paren{x; a, b, \tau} = \softmax_2\paren{x, 2x-a, 3x-a-b; \tau}.
\end{equation}
We now distinguish four cases:
\begin{enumerate}
    \item If $x < a \leq b$, then (I) $-x > - a \Leftrightarrow x > 2x - a$. Similarly, (II) $x < b \Leftrightarrow 0 > x - b$. Summing inequalities (I) and (II), we get 
    \begin{equation}
        x > 2x - a > 3x - a - b.
    \end{equation}
    Hence, $\lim_{\tau\to 0^+} \softInterval\paren{x; a, b, \tau} = 0$, which follows from the fact that the softmax output corresponding to the \textit{non-maximal} logit $2x-a$ vanishes in the limit.
    \item If $a < x < b$, we get $x < 2x - a$ and $2x - a > 3x - a - b$. It follows that $\lim_{\tau\to 0^+} \softInterval\paren{x; a, b, \tau} = 1$ from the convergence of the softmax function for the \textit{maximal} logit $2x-a$.
    \item If $a\leq b < x$, then $x < 2x - a < 3x - a - b$, proving proper convergence for the \textit{non-maximal} logit $2x-a$ analogous to the first case.
    \item When $x = a = b$, all three logits are equal, and the softmax converges to $1/3$.
    When $x = a$ or $x = b$ but not both, two logits tie for the maximum, so the softmax assigns $1/2$ to the second coordinate.
\end{enumerate}

\subsection{\ourMethod{}: Antecedent Layer}
\label{app:subsec_temp_annealing_antecedent}
To show convergence of the antecedent layer, we simplify the notation in \cref{eq:ant_act} by substituting $\predAct_i = \predAct_i\paren{\seq_{<t_{\predInd}}}$ and $\tau = \tau_2$.
We further expand $\secondWeights_{j,i} = \softplus(u_i;\tau)$, since the weights in \cref{eq:ant_act} are parameterized as the post-softplus transformations of underlying real-valued parameters to ensure non-negativity.
With this notation, we get
\begin{align}
    & \lim_{\tau\to 0^+} \sigma\paren{\frac{1}{\tau}\cdot \paren{\sum_{i=1}^\countInNeurons \softplus\paren{u_i; \tau}\cdot \paren{\predAct_i + \frac{\tau}{2} - 1}} - \log\paren{\tau}}\\
    & = \lim_{\tau\to 0^+} \sigma\paren{ \paren{\sum_{i=1}^\countInNeurons \frac{\tau}{\tau}\cdot \log\paren{1+\exp\paren{\frac{u_i}{\tau}}}\cdot \paren{\predAct_i + \frac{\tau}{2} - 1}} - \log\paren{\tau}}\\
    & = \lim_{\tau\to 0^+} \sigma\paren{ \paren{\sum_{i=1}^\countInNeurons \log\paren{1+\exp\paren{\frac{u_i}{\tau}}}\cdot \paren{\predAct_i - 1} + \frac{\softplus\paren{u_i; \tau}}{2}} - \log\paren{\tau}}\\
    & = \sigma\paren{ \lim_{\tau\to 0^+} \paren{\sum_{i=1}^\countInNeurons \log\paren{1+\exp\paren{\frac{u_i}{\tau}}}\cdot \paren{\predAct_i - 1} + \frac{\max(u_i, 0)}{2}} - \log\paren{\tau}}.
\end{align}
Here we used that $\softplus(u_i; \tau)$ converges to the ReLU function $\max(0, u_i)$ as $\tau \to 0^+$ (Appendix~\ref{app:par_softplus_convergence}).
To analyze the limit, we split the sum into two parts: terms with $\predAct_i - 1 = 0$ and terms with $\predAct_i - 1 < 0$, noting that $\predAct_i \in [0,1]$:
\begin{align}
    & \sigma\paren{ \lim_{\tau\to 0^+} \paren{\sum_{i=1}^\countInNeurons \log\paren{1+\exp\paren{\frac{u_i}{\tau}}}\cdot \paren{\predAct_i - 1} + \frac{\max(u_i, 0)}{2}} - \log\paren{\tau}}\\
    & = \sigma\left( \lim_{\tau\to 0^+} \paren{\sum_{i: (\predAct_i - 1) = 0} \frac{\max(u_i, 0)}{2}}\right.\notag \\
    & \qquad + \left.\paren{\sum_{i: (\predAct_i - 1) < 0} (\predAct_i - 1)\cdot \log\paren{1+\exp\paren{\frac{u_i}{\tau}}} + \frac{\max(u_i, 0)}{2}} - \log\paren{\tau}\right)\\
    & = \begin{cases}
        1, & \text{if } \bigwedge_{i=1}^\countInNeurons (u_i > 0)\rightarrow (\predAct_i = 1)  \\
        0, & \text{otherwise}
    \end{cases} = \indicator\paren{\bigwedge_{i=1}^\countInNeurons (u_i > 0)\rightarrow (\predAct_i = 1)}
\end{align}
To justify the last transformation, observe that
\begin{equation}
    \lim_{\tau\to 0^+} \log\paren{1+\exp\paren{\frac{u_i}{\tau}}} = \begin{cases}
        0, & \text{if } u_i < 0  \\
        \log(2), & \text{if } u_i = 0\\
        \infty, & \text{otherwise}
    \end{cases},
\end{equation}
so that these terms either vanish, remain constant, or diverge depending on the sign of $u_i$.
In the limit, the only non-constant contributions are $-\log(\tau)$ and $(\predAct_i - 1)\cdot \log\paren{1+\exp\paren{\frac{u_i}{\tau}}}$ for $\predAct_i < 1$ and $u_i > 0$.
All other terms, including $\tfrac{\max(u_i, 0)}{2}$, are dominated by them.
Finally, observe that $-\log(\tau) \to \infty$ as $\tau \to 0^+$, but it is itself dominated by any negative term of the form $(\predAct_i - 1)\cdot \log\paren{1+\exp\paren{\frac{u_i}{\tau}}}$ with $u_i > 0$, which diverges to $-\infty$ in the limit.

Note that the conjunction $\bigwedge_{i=1}^\countInNeurons (u_i > 0)\rightarrow (\predAct_i = 1)$ also holds for the post-softplus weights after annealing, since $\lim_{\tau\to0^+} \softplus(u_i;\tau) = \max(0,u_i)$ and thus $u_i > 0$ if and only if $\softplus(u_i;\tau) > 0$ in the limit.

\section{Additional Details on \ourMethodTitle{}}
\label{app:sec_rule_network_details}
\subsection{Layer Initialization}
\label{app:subsec_layer_init}
Parameter initialization is an important aspect for effective training.
Here, we focus on the initialization of the non-standard parameters specific to our model, namely $\firstWeights, \minOffset, \maxOffset$, and $\secondWeights$; all other layers use the default initialization of PyTorch.
In the following, $\countInNeurons$ denotes the number of predicate neurons in the network and $\countAntNeurons$ denotes the number of antecedent neurons.

\paragraph{Predicate Layer.}
For the weights of the first layer, $\firstWeights \in \mathbb{R}^{\countInNeurons \times D}$, we use a classical zero-centered uniform initialization \citep{2010_layer_init},
\begin{equation}
    \firstWeights_{i,d} \sim \mathcal{U}\paren{-\frac{1}{\sqrt{D}}, \frac{1}{\sqrt{D}}},
\end{equation}
where $\mathcal{U}$ denotes the uniform distribution.

For the delay intervals $\brackets{\minOffset_i, \maxOffset_i}$, we assume a known maximum lookback window $\maxLookback > 0$ and a small minimum offset $\varepsilon > 0$, ensuring $\minOffset_i \ge \varepsilon > 0$.
If $\maxLookback$ and $\varepsilon$ are not provided, we can set $\maxLookback$ to the largest temporal gap between any two \seqElementText{}s in a sequence and choose $\varepsilon$ no larger than the minimum time between any two consecutive \seqElementText{}s.
Given $\maxLookback$ and $\varepsilon$, we first sample the interval length
\begin{equation}
    \maxOffset_i - \minOffset_i \sim \paren{\maxLookback - \varepsilon}\cdot \mathcal{B}\paren{2.5, 5.0}
\end{equation} 
and then choose the starting point $\minOffset_i$ uniformly at random so that $\varepsilon \leq \minOffset_i \leq \maxOffset_i \leq \maxLookback$:
\begin{align}
    \minOffset_i &\sim \mathcal{U}\paren{\varepsilon, \maxLookback - \paren{\maxOffset_i - \minOffset_i}},\\
    \maxOffset_i & = \minOffset + \paren{\maxOffset_i - \minOffset_i}.
\end{align}
By sampling the initial interval width first, we make it straightforward to control.
The Beta distribution $\mathcal{B}(2.5, 5.0)$ favors moderate-to-narrow intervals, avoids extremely short ones, and still permits wider intervals.
Its overall shape results in a balanced coverage of the \seqElementText{} history.

\paragraph{Antecedent Layer.}
For the second hidden layer, we also initialize the incoming weights to small values, avoiding saturation of the activation function and facilitating training.
Since we use the softplus function to enforce non-negativity of $\secondWeights \in \mathbb{R}_{\geq 0}^{\countAntNeurons \times \countInNeurons}$, we first sample the post-softplus weights
\begin{equation}
    \secondWeights_{j,i} \sim \mathcal{U}\paren{\frac{1}{\sqrt{\countInNeurons}}-\frac{1}{2\sqrt{\countInNeurons}}, \frac{1}{\sqrt{\countInNeurons}}+\frac{1}{2\sqrt{\countInNeurons}}},
\end{equation}
and then apply the inverse softplus (at temperature $\tau = 1$) to obtain the initial underlying parameters $\secondWeights'_{j,i}$:
\begin{equation}
    \secondWeights'_{j,i} = \softplus^{-1}\paren{\secondWeights_{j,i}} = \log\paren{\exp\paren{\secondWeights_{j,i}} - 1}.
\end{equation}
This yields initial weights $\secondWeights_{j,i} = \softplus(\secondWeights'_{j,i}; \tau)$ uniformly distributed around $\frac{1}{\sqrt{\countInNeurons}}$ during the first forward pass, providing a scaling that helps keep the layer outputs well-conditioned at initialization.

\subsection{Interpreting \ourMethodTitle{} as a Symbolic Rule Set}
\label{app:subsec_rule_set_extraction}

After training, we can directly use the symbolic versions of the antecedents within \ourMethod{}, as well as for defining the symbolic rules themselves.
In this section, we present the core idea for deriving symbolic consequents by discretizing the softmax layer of \ourMethod{}, while Appendix~\ref{app:subsec_antecedent_pruning} introduces additional steps for more accurate rule extraction.

\paragraph{Translation Invariance of the Softmax.}
A key property for discretizing the softmax layer into symbolic consequent \seqElementText{}s is its \emph{translation invariance} when adding the same constant $c\in\mathbb{R}$ to each coordinate, i.e.,
\begin{equation}
    \softmax\paren{x + c\cdot \vec{1}} = \softmax\paren{x}.
\end{equation}
Here $\vec{1}$ denotes a vector of ones.
Consequently, for rule logits $O_j \in \mathbb{R}^D$, their influence on the classification depends on the \emph{relative distances} between individual \seqElementText{} logits, rather than their absolute values.
Exploiting this, for any binary (symbolic) antecedent activation vector $\antAct\paren{\seq_{<t_{\predInd}}} \in \set{0,1}^\countAntNeurons$, we can rewrite
\begin{align}
    \softmax\paren{b + \sum_{j=1}^J \antAct_j\paren{\seq_{<\seqTime_{\predInd}}}\cdot \outputWeights_j} & = \softmax\paren{b + \sum_{j: \antAct_j\paren{\seq_{<t_{\predInd}}} = 1} \outputWeights_j} \\
    & = \softmax\paren{b - \mu\paren{b}\cdot \vec{1} + \sum_{j: \antAct_j\paren{\seq_{<t_{\predInd}}} = 1} \outputWeights_j - \mu\paren{\outputWeights_j}\cdot \vec{1}},
\end{align}
since the mean vector $\mu(\cdot)\cdot \vec{1}$ is constant across all coordinates.
The mean-centered logits
\begin{equation}
    \tilde{\outputWeights}_j \coloneq \outputWeights_j - \mu\paren{\outputWeights_j}\cdot \vec{1}
\end{equation}
capture this distance-based interpretation: a large positive logit indicates that the rule increases the likelihood of the associated \seqElementText{}, whereas a large negative logit indicates the opposite.

\paragraph{Algorithmic Discretization.}
Our algorithm for discretizing the softmax layer relies on three user-defined parameters:
\begin{itemize}
    \item The \textit{min-max variation threshold} $\threshold_{\mathrm{var}}\in \RR_{\geq 0}$ filters out rules with near-constant logits, which have negligible influence on the predictions due to the translation invariance of the softmax.
    We require $\max_d\paren{\tilde{\outputWeights}_{j,d}} - \min_d\paren{\tilde{\outputWeights}_{j,d}} > \threshold_{\mathrm{var}}$.
    \item The \textit{quantile threshold} $\threshold_{\mathrm{quant}}\in \brackets{0,1}$ ensures that we only keep rules that strongly favor an \seqElementText{} compared to other learned rules.
    Specifically, we include a rule only if its maximum mean-centered logit exceeds the $\threshold_{\mathrm{quant}}$-quantile, denoted as $Q_{\threshold_{\mathrm{quant}}}$, of all mean-centered logits and mean-centered biases.
    \item The \textit{inclusion threshold} $\threshold_{\mathrm{inc}}\in (0,1]$ determines which \seqElementText{}s are included in the consequent for rules passing the first two criteria.
    An \seqElementText{} $d$ is included if it satisfies $\tilde{\outputWeights}_{j,d} > \threshold_{\mathrm{inc}}\cdot \max_d\paren{\tilde{\outputWeights}_{j,d}}$.
\end{itemize}

If we wish to include rules predicting the \textit{absence} of \seqElementText{}s, we can analogously define quantile and inclusion thresholds using the minimum mean-centered logits.

\paragraph{Formal Rule Set Definition.}

\begin{figure}[ht!]
    \centering
    \begin{tikzpicture}[
        scale=0.84, transform shape,
        heading/.style={rectangle, draw=none, minimum size=7mm, align=center},%
        table/.style={rectangle, draw=none, text=black, minimum size=7mm, align=center},%
    ]   
        \node[heading, text=black] (t1h) at (0,15) {\textbf{Predicate Neurons}};
        \node[table] (t1) [below=0.15cm of t1h] {%
            \begin{tabular}{c rrrr rr}
            \toprule
            & \multicolumn{4}{c}{$\firstWeights$: Input Weights (\SeqElementText{}s)} & & \\
            \cmidrule(lr){2-5}
            Neuron & \textit{computer case}       & \textit{computer processor}  & \textit{motherboard}         & \textit{power supply unit}       & $\minOffset$ & $\maxOffset$  \\
            \midrule
            1      & \textcolor{black}{   $ 3.2$} & \textcolor{black!40}{$ 1.9$} & \textcolor{black!40}{$-1.2$} & \textcolor{black!40}{$-1.5$}     & $3.0$        & $15.0$        \\
            2      & \textcolor{black}{   $ 2.7$} & \textcolor{black!40}{$-2.1$} & \textcolor{black!40}{$ 0.2$} & \textcolor{black!40}{$ 0.7$}     & $1.0$        & $43.0$        \\
            3      & \textcolor{black!40}{$-1.1$} & \textcolor{black}{   $ 3.9$} & \textcolor{black!40}{$-1.0$} & \textcolor{black!40}{$ 0.9$}     & $2.0$        & $21.0$        \\
            4      & \textcolor{black!40}{$-0.4$} & \textcolor{black!40}{$-0.1$} & \textcolor{black!40}{$ 0.5$} & \textcolor{black}   {$ 1.3$}     & $4.0$        & $ 9.0$        \\
            \bottomrule
            \end{tabular}
        };

        \node[heading, text=black] (t2h) [below=0.5cm of t1.south] {\textbf{Antecedent Neurons \& Output Layer}};
        \node[table] (t2) [below=0.15cm of t2h] {%
            \begin{tabular}{c rrrr rrrr}
            \toprule
            &  \multicolumn{4}{c}{$\secondWeights$: Conjunction Weights} & \multicolumn{4}{c}{$\tilde{\outputWeights}$: Centered Output Weights (\SeqElementText{}s)} \\
            \cmidrule(lr){2-5} \cmidrule(lr){6-9}
            Neuron  & \multicolumn{1}{c}{1}       & \multicolumn{1}{c}{2}       & \multicolumn{1}{c}{3}       & \multicolumn{1}{c}{4}       & \textit{computer case}       & \textit{computer processor}  & \textit{motherboard}         & \textit{power supply unit}  \\
            \midrule
            1       & \textcolor{black}{   $1.8$} & \textcolor{black!40}{$0.0$} & \textcolor{black!40}{$0.0$} & \textcolor{black!40}{$0.0$} & \textcolor{black!40}{$-3.4$} & \textcolor{black!40}{$ 0.3$} & \textcolor{black}{   $ 2.9$} & \textcolor{black!40}{$ 0.2$} \\
            2       & \textcolor{black!40}{$0.0$} & \textcolor{black}{   $2.2$} & \textcolor{black}{   $1.9$} & \textcolor{black!40}{$0.0$} & \textcolor{black!40}{$-3.6$} & \textcolor{black!40}{$-1.1$} & \textcolor{black}{   $ 2.3$} & \textcolor{black}{   $ 2.4$} \\
            3       & \textcolor{black!40}{$0.0$} & \textcolor{black!40}{$0.0$} & \textcolor{black!40}{$0.0$} & \textcolor{black}{   $0.6$} & \textcolor{black!40}{$-0.2$} & \textcolor{black!40}{$ 0.2$} & \textcolor{black!40}{$ 0.3$} & \textcolor{black!40}{$-0.3$} \\
            \midrule
            Bias    & & & &                                                                                                                 & \textcolor{black!40}{$ 0.7$} & \textcolor{black!40}{$-0.1$} & \textcolor{black!40}{$-0.4$} & \textcolor{black!40}{$-0.2$} \\
            \bottomrule
            \end{tabular}
        };

        \node[heading, text=black] (t3h) [below=1.75cm of t2.south] {\textbf{Rule Set}};
        \node[table, align=left, draw=black] (t3) [below=0.25cm of t3h]{
            $\set{(\textit{computer case}, [3,15])} \ruleTo \set{\textit{motherboard}}$\\
            $\set{(\textit{computer case}, [1,43]), (\textit{computer processor}, [2,21])} \ruleTo \set{\textit{motherboard}, \textit{power supply unit}}$
        };

        \draw[-{Stealth}, thick] ($(t2.south) - (0.0, 0.1)$) to ($(t3h.north) + (0.0, 0.1)$);
        \node[heading, draw=black, fill=white, rounded corners] (disc) at ($(t2.south)!0.5!(t3h.north)$) {Discretize};

    \end{tikzpicture}
    \caption{
        \textit{Example symbolic rule set derived from the weights of \ourMethod{}.}
        Black highlights retained weights after discretization; \textcolor{black!40}{gray} marks discarded weights.
        \textbf{Top table}:~Event-time predicate neurons encoding $(\textit{computer case}, [3,15])$, $(\textit{computer case}, [1,43])$, $(\textit{computer processor}, [2,21])$, and $(\textit{power supply unit}, [4,9])$.
        \textbf{Bottom table}:~(Post-ReLU) conjunction weights (\textit{left}) and mean-centered output weights (\textit{right}).
        Positive conjunction weights indicate the inclusion of a predicate in the antecedent; for example, antecedent neuron~2 combines predicate neuron~2 $(\textit{computer case}, [1,43])$ and predicate neuron~3 $(\textit{power supply unit}, [2,21])$.
        \textbf{Bottom}:~Symbolic rule set discretized with $\threshold_{\mathrm{var}} = 0.9$, $\threshold_{\mathrm{quant}} = 0.875$, and $\threshold_{\mathrm{inc}} = 0.9$.
        The rule of antecedent neuron 3 is discarded because its maximum logit is not in the top $(1-87.5)$\% (top 2); it also fails to meet the variation threshold $\threshold_{\mathrm{var}} = 0.9$.
        The rule of neuron~2 includes two consequent \seqElementText{}s because its highest logit (headphones) is in the top 2, and its second-highest (phone case) satisfies the inclusion threshold $\threshold_{\mathrm{inc}}$.
    }
    \label{fig:example_rule_set_exrtraction}
\end{figure}

We formally define the set of significant rules encoded by \ourMethod{} after training as
\begin{equation}
    \begin{aligned}
    & \left\{\set{\paren{\argmax_d \firstWeights_{i,d}, \brackets{\minOffset_i, \maxOffset_i}} \ \big\vert\ 1\leq i\leq \countInNeurons: \secondWeights_{j,i} > 0}\right.\\
    & \quad \ruleTo \set{d  \ \big\vert\ 1\leq d\leq \vocabSize: \tilde{\outputWeights}_{j,d} > \threshold_{\mathrm{inc}}\cdot \max_{d'}\paren{\tilde{\outputWeights}_{j,d'}}} \\ 
    & \quad \left. \big\vert\ 1\leq j\leq \countAntNeurons: \paren{\max_d\paren{\tilde{\outputWeights}_{j,d}} - \min_d\paren{\tilde{\outputWeights}_{j,d}} > \threshold_{\mathrm{var}}}\land \paren{\max_d\paren{\tilde{\outputWeights}_{j,d}} > Q_{\threshold_{\mathrm{quant}}}} \right\},
    \end{aligned}
    \label{eq:rule_set_formally}
\end{equation}
where $\countInNeurons$ and $\countAntNeurons$ denote the number of predicate and antecedent neurons, respectively. 
In \cref{fig:example_rule_set_exrtraction}, we illustrate the application of this definition with $\threshold_{\mathrm{var}} = 0.9$, $\threshold_{\mathrm{quant}} = 0.875$, and $\threshold_{\mathrm{inc}} = 0.9$.

The condition 
\begin{equation}
    \paren{\max_d\paren{\tilde{\outputWeights}_{j,d}} - \min_d\paren{\tilde{\outputWeights}_{j,d}} > \threshold_{\mathrm{var}}}\land \paren{\max_d\paren{\tilde{\outputWeights}_{j,d}} > Q_{\threshold_{\mathrm{quant}}}}
\end{equation}
ensures that only rules satisfying both the min-max variation and quantile thresholds are included.
The $\threshold_{\mathrm{quant}}$-quantile $Q_{\threshold_{\mathrm{quant}}}$ is computed over the set 
\begin{equation}
    \set{\tilde{\outputWeights}_{j,d}\mid 1\leq j\leq \countAntNeurons, 1\leq d\leq \vocabSize} \cup \set{b_d - \mu\paren{b}\mid 1\leq d\leq \vocabSize},
\end{equation}
so it includes the mean-centered bias terms alongside the logits.
We include the bias because, although it does not require an antecedent activation, it influences predictions in the same way as rule logits.
In the example, the rule of antecedent neuron~3 is discarded because its maximum mean-centered logit is below the $\threshold_{\mathrm{quant}}$-quantile $Q_{0.875} = 2.3$, and it also fails to meet the min-max variation threshold of $0.9$ for its logits.

The antecedent set
\begin{equation}
    \set{\paren{\argmax_d \firstWeights_{i,d}, \brackets{\minOffset_i, \maxOffset_i}} \ \big\vert\ 1\leq i\leq \countInNeurons: \secondWeights_{j,i} > 0}
\end{equation}
contains all learned \seqElementText{}s and temporal constraints for predicate neurons with positive weights $\secondWeights_{j,i}$, meaning they participate in the conjunction for antecedent neuron~$j$.
In \cref{fig:example_rule_set_exrtraction}, the predicate neurons encode $(\textit{computer case}, [3,15])$, $(\textit{computer case}, [1,43])$, $(\textit{computer processor}, [2,21])$, and $(\textit{power supply unit}, [4,9])$ through their maximum weight and delay interval parameters.
Antecedent neuron~2 combines the second and the third predicate into the antecedent $\{(\textit{computer case}, [1,43]), (\textit{computer processor}, [2,21)\}$, as indicated by its positive weights for these neurons.
Similarly, antecedent neurons~1 and 3 encode $\{(\textit{computer case}, [3,15])\}$ and $\{(\textit{power supply unit}, [4,9])\}$, respectively.

The consequent set
\begin{equation}
    \set{d  \ \big\vert\ 1\leq d\leq D: \tilde{\outputWeights}_{j,d} > \threshold_{\mathrm{inc}}\cdot \max_{d'}\paren{\tilde{\outputWeights}_{j,d'}}}
\end{equation}
includes the \seqElementText{} with the highest mean-centered logit and any others whose mean-centered logits exceed the inclusion threshold $\threshold_{\mathrm{inc}}$ times the maximum logit.
For antecedent neuron~2, this results in including not only the \seqElementText{} with the maximum logit (\textit{power supply unit}) but also the \seqElementText{} with the second-highest logit (\textit{motherboard}), even though the latter does not meet the quantile threshold.

\subsection{Antecedent Pruning for Improved Rule Extraction}
\label{app:subsec_antecedent_pruning}

For accurate discretization of the softmax layer (Appendix~\ref{app:subsec_rule_set_extraction}), it is important to treat \textit{equivalent antecedents} together.
For example, when two antecedent neurons $j_1$ and $j_2$ use exactly the same event-time predicates in their conjunction, i.e., $\forall i:\paren{\secondWeights_{j_1, i} > 0} \leftrightarrow \paren{\secondWeights_{j_2, i} > 0}$, they model the same antecedent.
Moreover, since these antecedents are perfectly co-active, their logits $\outputWeights_{j_1}$ and $\outputWeights_{j_2}$ do not act independently. 
Instead, they always occur together, causing their contributions to combine additively as the sum $\outputWeights_{j_1} + \outputWeights_{j_2}$.

To address this, we apply three pre-processing steps before discretizing the softmax layer. None of these steps change the behavior of the model on the given dataset:
\begin{enumerate}
    \item Prune neurons that are always on or always off.
    \item Prune \textit{subsumed} predicates within antecedents. These predicates are redundant due to stricter temporal constraints on the same \seqElementText{}.
    \item Unify \textit{equivalent} antecedents and their consequents when computing the final rule set.
\end{enumerate}

\paragraph{Step 1: Pruning Neurons.}
For a given dataset, hidden neurons that are never active can be safely pruned, as antecedents with activation $\antAct_j(\seq_{<t_\predInd}) = 0$ do not contribute to the softmax layer. 
Similarly, predicate neurons with always-off activations $\predAct_i(\seq_{<t_\predInd}) = 0$ force any antecedent containing them to be inactive due to the identity $0 \land \predAct = 0$ for any predicate $\predAct$. 
Such always-off neurons may arise if all weights $\firstWeights_{i,d}$ are negative, effectively suppressing activation.

Conversely, always-on neurons with $\predAct_i(\seq_{<t_\predInd}) = 1$ can also be pruned in the first layer, since $1 \land \predAct = \predAct$. 
In the second layer, always-on antecedent neurons $\antAct_j(\seq_{<t_\predInd})$ contribute their logits~$\outputWeights_j$ to every classification, acting like a constant bias. 
Thus, they can be removed by updating the bias vector as $b \leftarrow b + \outputWeights_j$.

Using strict definitions of always-on and always-off neurons, pruning does not alter model performance on the dataset.
However, applying thresholds on activation frequency enables pruning of rare or overly specific neurons at the cost of minor behavioral changes.

\paragraph{Step 2: Removing Subsumed Event-Time Predicates.}
An event-time predicate $\paren{\antElement_{1}, \brackets{\minOffset_{1}, \maxOffset_{1}}}$ \textbf{subsumes} another predicate $\paren{\antElement_{2}, \brackets{\minOffset_{2}, \maxOffset_{2}}}$ if
\begin{equation}
    \paren{\antElement_{1} = \antElement_{2}} \land \paren{\brackets{\minOffset_{1}, \maxOffset_{1}} \subseteq \brackets{\minOffset_{2}, \maxOffset_{2}}}.\label{eq:subsumption}
\end{equation}
When both occur in the same antecedent, the predicate with the wider interval is redundant: the conjunction is active only if the stricter (subset-interval) predicate holds.
We can directly prune subsumed predicates in \ourMethod{} by setting the corresponding weight $\secondWeights_{j,i} = 0$, simplifying rules without altering network outputs.

\paragraph{Step 3: Unifying Equivalent Antecedents.}
Once subsumed predicates are removed, antecedents are equivalent (always activate for the same \seqElementText{}s) if their predicate sets are identical.
Neurons defining equivalent antecedents are functionally redundant since they always activate together.
We can unify them by removing one neuron and adding its logits to the logits of the remaining neuron, without changing model behavior.

\subsection{Symbolic Refinement of Learned Rules}
\label{app:subsec_symbolic_rule_refinement}
Rules with symbolic consequents obtained after discretization (Appendices~\ref{app:subsec_rule_set_extraction} and~\ref{app:subsec_antecedent_pruning}) enable the use of classical algorithms to improve predictive performance or interpretability.
For example, antecedents learned by \ourMethod{} may contain redundant predicates or overly wide delay intervals, both of which we address with symbolic post-processing.
Crucially, refining a small set of learned rules is typically \textbf{much more efficient} than selecting rules from a large pool of candidates.

Our post-processing reduces antecedent complexity and suppresses spurious rule activations while preserving correct activations (support) of rules with symbolic consequents $\cons \subseteq \vocab$.
When desired, the resulting refined antecedents can be re-embedded into the hidden layers of \ourMethod{}, requiring retraining of only the output layer.
As shown in Appendix~\ref{app:subsec_effect_rule_processing}, this procedure reduces the number of rules that contribute to each prediction in \ourMethodSideChannel{}, increasing sparsity and interpretability.
We apply the post-processing by evaluating rules on the training data using the following steps.

\paragraph{Removing Data-Equivalent Predicates.}
We remove predicates from an antecedent when they behave identically on a given (training) dataset, even if they are not logically equivalent.
We identify such pairs by making a single pass over the data and, for each pair of predicates within the same rule, checking whether they have identical truth values for all \seqElementText{}s in each sequence.
If that is the case, we remove one of the two predicates from the antecedent without changing when the rule activates.

\paragraph{Unifying Overlapping Predicates.}
Another factor that contributes to unnecessarily large antecedents is the presence of multiple overlapping temporal predicates that together effectively model a single predicate.
For example, \ourMethod{} might learn $\set{(\antElement, [1,5]), (\antElement, [3,7])}$ instead of a single predicate $(\antElement, [3,5])$ representing the intersection of both intervals.
These (partial) antecedents are not logically equivalent, since two occurrences of $\antElement$, $1$ and $7$ time steps ago, can satisfy the conjunction of the individual predicates but not the intersection.
However, for a given dataset, both can produce equivalent rule activations.

We seek to unify such predicates that share the same \seqElementText{} and have overlapping delay intervals.
Given a subset of predicates $\set{\paren{\antElement, \brackets{\minOffset_i,\maxOffset_i}}\mid 1\leq i\leq K}$ with the same \seqElementText{} $\antElement$, we consider replacing them with the joint predicate
\begin{equation}
    \paren{\antElement, \bigcap_{i=1}^K \brackets{\minOffset_i, \maxOffset_i}},
\end{equation}
where the delay interval is the intersection of the original intervals.
Specifically, we enumerate all joint predicates with non-empty intersections formed by combining two or more predicates in an antecedent.
We then verify on the dataset whether the joint predicate activates whenever the original rule is supported.
If so, we replace the individual predicates with the joint predicate.
If multiple joint predicates are valid, we heuristically select those that unify the most predicates first, ensuring that each original predicate is used in only one intersection.

Importantly, it suffices to consider only the \seqElementText{}s $\seqElement_\seqInd$ in a sequence $\seq$ that support the original rule $\genericRule$---i.e., where the antecedent is active ($\indicatorRule_\ant(\seq, \seqTime_\seqInd)$) and $\seqElement_\seqInd$ is part of the consequent ($\seqElement_\seqInd \in \cons$)---rather than all \seqElementText{}s in the dataset, since the joint predicate subsumes the original predicates. In other words, whenever the joint predicate is true, all original predicates are also true.

Testing joint predicates formed from all subsets of predicates sharing the same \seqElementText{} might seem costly, since the number of subsets grows exponentially with the number of these predicates in an antecedent.
In practice, however, we rarely encounter many overlapping predicates within the same rule, so this step has little impact on runtime.
If needed, the number of predicates considered when forming joint predicates can be further limited to control computational cost.

\paragraph{Tightening Delay Constraints.}
The learned delay intervals $\brackets{\minOffset_i, \maxOffset_i}$ may be wider than necessary.
This can occur when the initial soft model fits the data well with wide intervals, after which temperature annealing sharpens the model and reduces gradient flow to these parameters.
The effect is exacerbated when the model exploits entangled combinations of rule activations to predict certain \seqElementText{}s rather than relying on individual rule consequents.
Temperature annealing further complicates regularization of $\minOffset$ and $\maxOffset$, as gradients at low temperatures become dominated by the regularization term.

To tighten delay intervals for extracted symbolic rules, we consider candidate sub-intervals $\brackets{\minOffset_{*}, \maxOffset_{*}} \subseteq \brackets{\minOffset_i, \maxOffset_i}$ and evaluate their activations on the dataset. 
Two observations make it tractable to efficiently evaluate the otherwise infinite set of possible candidates:\footnote{
    For regularly sampled sequences or integer-valued times, the number of candidates is technically only exponential; for real-valued times, it is infinite.
}
\begin{itemize}
    \item The interval bounds of an optimal event-time predicate must coincide with observed delays of rule occurrences; otherwise, a tighter interval covering the same \seqElementText{}s exists.
    \item Since an interval subsumes another if it is contained within it (\cref{eq:subsumption}), the search space can be efficiently pruned.
\end{itemize}
Among all non-subsumed candidates, we select the smallest interval $\brackets{\minOffset_{*},\maxOffset_{*}}$ and replace the original predicate $\paren{\antElement, \brackets{\minOffset_i,\maxOffset_i}}$ with $\paren{\antElement, \brackets{\minOffset_{*},\maxOffset_{*}}}$.
Because the new interval is contained within the original, it cannot introduce incorrect rule activations.
Consequently, it suffices to consider only the \seqElementText{}s supporting the antecedent and consequent of a rule when tightening delay constraints.

\section{Connections to Temporal Logic}
\label{app:sec_connection_temporal_logic}
In this section, we situate our rule language within existing temporal formalisms.
\ruleName{}s (\cref{subsec:tcp_rules}) are connected to metric temporal logic (MTL) \citep{1990_mtl}, which extends linear temporal logic (LTL) by adding timing constraints to the temporal operators.
Consider the event-time predicate induced by an antecedent element $\paren{\antElement, \brackets{\minOffset, \maxOffset}}$ with $\delayInterval = \brackets{\minOffset, \maxOffset}$. Evaluated at time $\seqTime$ on history $\seq_{<\seqTime}$, the predicate is
\begin{equation}
    \bigvee_{\paren{\seqElement_\seqInd, \seqTime_\seqInd} \in \seq_{<\seqTime}}
           (\seqElement_\seqInd = \antElement) \land \paren{\seqTime - \seqTime_\seqInd \in \delayInterval},
\end{equation}
which is exactly the satisfaction condition of the past-time metric operator ``ONCE $\antElement$ within $\delayInterval$''. 
This is expressible via the metric SINCE operator $\mathcal{S}$ in MTL as $\mathrm{true}\ \mathcal{S}_{\delayInterval}\,\antElement$ (i.e., ``$\antElement$ occurred at some point in $\delayInterval$, and since then $\mathrm{true}$ has held'').
A rule antecedent is a conjunction of such operators, and is therefore a formula in the bounded past-time fragment of MTL.
The only departure is the consequent: we use additive logits rather than a hard implication, which is what allows rules to be combined for prediction. 
Discretizing the consequent into a disjunction of \seqElementText{}s (Appendix~\ref{app:subsec_rule_set_extraction}) recovers a fully symbolic rule that is expressible in MTL.

\section{Scaling Behavior of \ourMethod{}}
\label{app:sec_discussion_complexity}
Training \ourMethod{} on a sequence of length $\seqLen$ requires computing a prediction for each event.
Most operations can be implemented as standard MLP layers, with the LogSumExp in \cref{eq:event_act} adding no asymptotic cost.
The dominant term arises from computing the activations of the predicate neurons $\predAct_i$ (\cref{eq:pred_act}), which depend on all values $\tilde{p}_i(\seqInd,\predInd)$ with $\seqInd < \predInd$.
Evaluating $\predAct_i$ across all $\seqLen-1$ prediction steps and for all $\countInNeurons$ predicate neurons thus requires $\mathcal{O}(\seqLen^2 \countInNeurons)$ time and space.
However, computations can be parallelized across prediction time steps, since \ourMethod{} includes no recurrent dependencies (i.e., neurons do not depend on previously computed states).

\paragraph{Practical Implications.} In practice, the sequence length determines the maximum \textit{event} lookback and thus implicitly constrains the \textit{temporal} lookback window $\maxLookback$.
This is only limiting for long-range rules spanning many events. 
For short-range rules, training on shorter sequences reduces computation without impairing performance or the ability to learn the underlying rule structure.
Specifically, when fixing $\seqLen$ (e.g., by training on subsequences), the time required to process $\seqLen_\mathrm{tot}$ events per epoch scales \emph{linearly} as
\[
\mathcal{O}\left(\frac{\seqLen_\mathrm{tot}}{\seqLen} \seqLen^2 \countInNeurons\right)
= \mathcal{O}(\seqLen_\mathrm{tot} \seqLen \countInNeurons),
\]
where $\frac{\seqLen_\mathrm{tot}}{\seqLen}$ is the number of (sub)sequences per epoch.

We confirm this empirically using the synthetic setup from \cref{subsec:synthetic_data} with vocabulary size $\vocabSize=2048$ and batch size $32$, setting $\countInNeurons=300$ predicate neurons and $\countAntNeurons=64$ antecedent neurons in \ourMethod{}.
We measure the average training step time (forward + backward) and standard deviation over $1024$ sequences after one warm-up epoch on an NVIDIA A100 GPU (40GB), and derive epoch times for $\seqLen_\mathrm{tot} = 2^{16}$ events:
\begin{itemize}
    \item $\seqLen=64$: $0.019 \pm 0.001$s per step $\rightarrow$ $0.608$s per epoch (32 batches),
    \item $\seqLen=256$: $0.267 \pm 0.001$s per step $\rightarrow$ $2.136$s per epoch (8 batches).
\end{itemize}
Increasing $\seqLen$ by a factor of 4 increases epoch time by approximately a factor of 3.5, consistent with the predicted linear scaling in $\seqLen$.
In practice, larger lookback windows often yield limited accuracy gains, as exemplified by the strong performance of \textsc{PPM} on real-world datasets with a maximum 3-event lookback (\cref{tab:table_real_results}).
Thus, increasing $\seqLen$ is often unnecessary to achieve strong performance.

\begin{figure}[ht]
    \centering
    \begin{subfigure}{0.49\linewidth}
    \raggedright
        \includegraphics[width=1.0\linewidth]{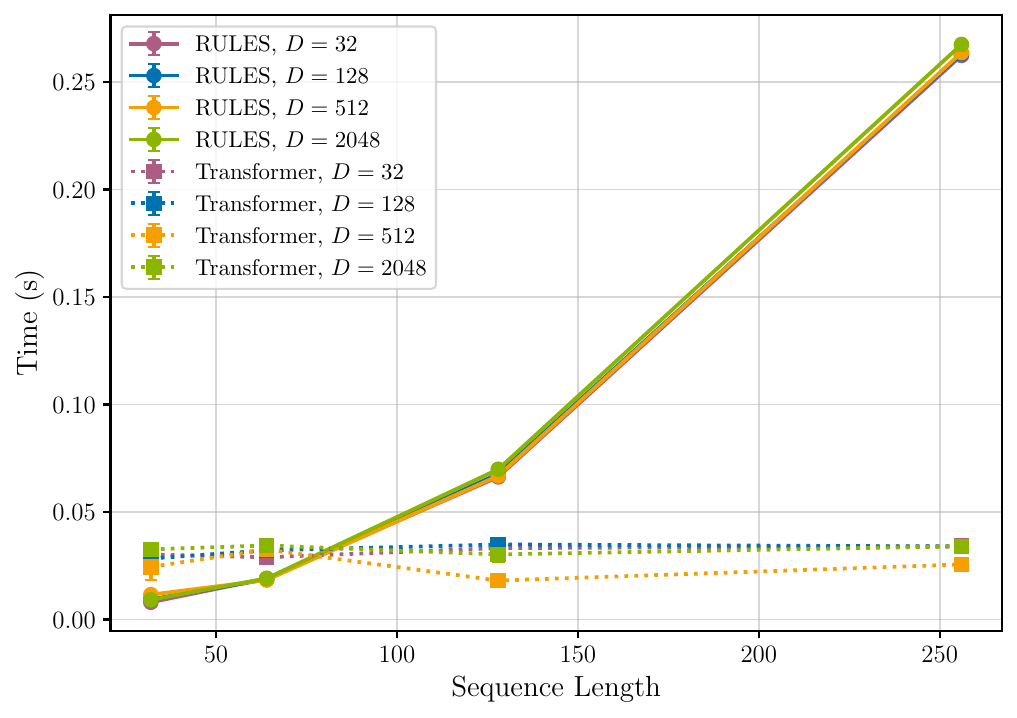}
        \caption{Per-batch training times.}
        \label{subfig:training_times}
    \end{subfigure}
    \hfill
    \begin{subfigure}{0.49\linewidth}
        \raggedleft
        \includegraphics[width=1.0\linewidth]{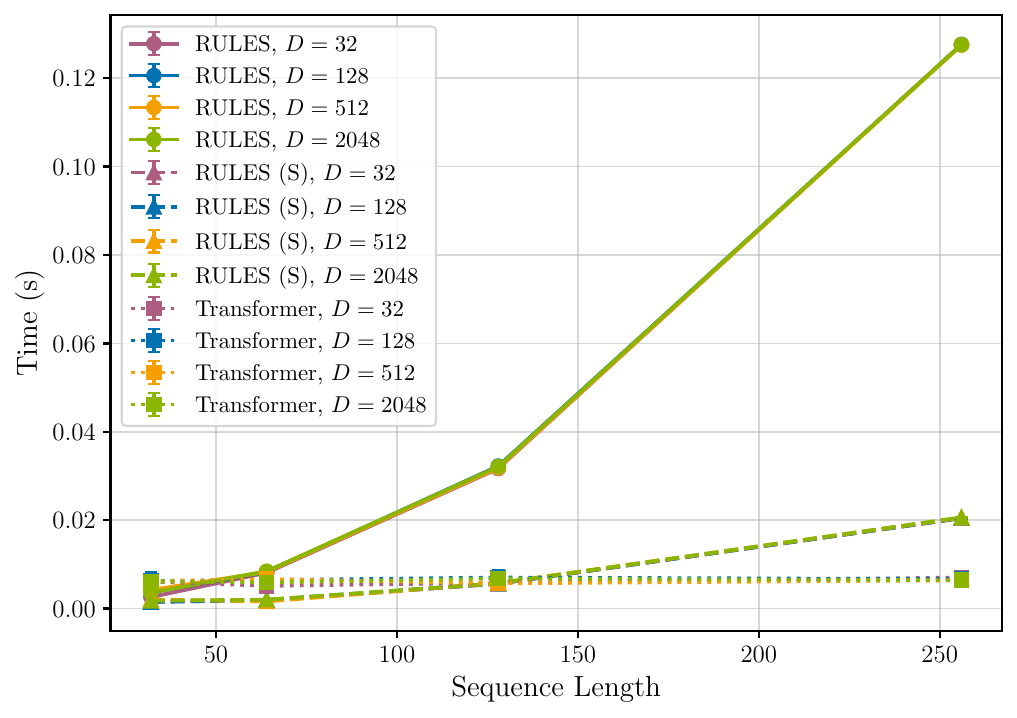}
        \caption{Per-batch inference times.}
        \label{subfig:inference_times}
    \end{subfigure}
    \caption{
        \textit{Per-batch computation times for \ourMethod{} and the transformer across varying sequence lengths $\seqLen$ and vocabulary sizes $\vocabSize$.}
        \ourMethod{}~(S) denotes the model with symbolic (boolean) hidden layers.
        We report mean and one standard deviation over 32 batches of size 32; variability is small, so error bars are barely visible.
        Both models are largely insensitive to the vocabulary size.
        For larger $\seqLen$, the expected quadratic scaling of \ourMethod{} becomes visible, while the symbolic variant achieves substantially faster inference.
    }
     
    \label{fig:runtimes}
\end{figure}

\paragraph{Scalability Study.} 
In \cref{fig:runtimes}, we further evaluate the scaling behavior of \ourMethod{} and compare it to the transformer used in our experiments.
We use the same synthetic setting as above, varying the vocabulary size $\vocabSize\in \set{32, 128, 512, 2048}$ and the sequence length $\seqLen\in \set{32, 64, 128, 256}$, and record average per-batch processing times for both training and inference.
Note that \ourMethodSideChannel{} effectively adds the forward pass of \ourMethod{}~(S) (symbolic inference) to the computations of the transformer.

We find that for both \ourMethod{} and the transformer, computation time is largely independent of the vocabulary size.
For shorter sequences ($\seqLen \leq 64$), \ourMethod{} achieves faster training times per batch than the transformer and maintains comparable inference times.
As $\seqLen$ increases, the expected quadratic dependence on sequence length becomes visible for \ourMethod{}; however, absolute runtimes remain moderate (e.g., around $0.26$s for training and $0.13$s for inference at $\seqLen=256$).
In contrast, transformer runtimes remain stable across sequence lengths, reflecting the parallel structure of attention computations.
While our rule formulation also enables \ourMethod{} to compute predicate and antecedent neurons in parallel for each prediction time step, its dependence on the number of predicate neurons $\countInNeurons$ and the current implementation cause the theoretical scaling to be more directly reflected in the runtime.

Finally, we observe that the symbolic (boolean) formulation of the hidden layers in \ourMethod{} (denoted by \ourMethod{}~(S)) substantially reduces inference time, further improving its efficiency in deployment settings.
\section{Baselines}
\label{app:sec_baseline_details}
\subsection{PPM-C} 
\textsc{PPM-C} \citep{1990_ppm_c,2004_vo_markov_models} is a table-based Markov model that considers a maximum context of $C\in \NN_{\geq0}$ previous sequence elements (\seqElementText{}s).
If a context of a certain length $c\leq C$ has not been encountered in the training data, \textsc{PPM-C} falls back to a lower order context.
Specifically, given the empirical probabilities $P_c$ for a context of length $c$, it uses
\begin{equation}
    \begin{aligned}
        & P(\seqElement_\predInd\vert \seqElement_{\predInd-c}, \dots,  \seqElement_{\predInd-1}) = \\ 
        & \quad \begin{cases}
            P_c(\seqElement_\predInd\vert \seqElement_{\predInd-c}, \dots,  \seqElement_{\predInd-1}), & 
            \begin{aligned}
                & \text{if } \seqElement_{\predInd - c}\dots \seqElement_{\predInd -  1}\seqElement_{\predInd} \\
                & \text{appeared in training}
            \end{aligned} \\
             P_c(\mathrm{fallback} \vert \seqElement_{\predInd-c}, \dots,  \seqElement_{\predInd-1})\cdot P(\seqElement_\predInd\vert \seqElement_{\predInd-c+1}, \dots,  \seqElement_{\predInd-1}), & \text{otherwise}
        \end{cases}.
    \end{aligned}
\end{equation}

In \textsc{PPM-C}, the fallback probability mass is defined using the occurrences of different symbols after context $s = \seqElement_{\predInd - c}\dots \seqElement_{\predInd -  1}$.
Denote by $n(s, \seqElement_{\predInd})$ the number of times $\seqElement_{\predInd}$ occurred after $s$ in training, and let $\vocab(s) = \set{\seqElement_{\predInd} \mid n(s, \seqElement_{\predInd}) > 0}$, then $P_c$ is defined as:
\begin{align}
    P_c(\seqElement_{\predInd} \vert s) & = \frac{n(s, \seqElement_{\predInd})}{\abs{\vocab(s)} + \sum_{\seqElement_{\predInd}' \in \vocab(s)} n(s, \seqElement_{\predInd}')}, \\
     P_c(\mathrm{fallback} \vert s) & = \frac{\abs{\vocab(s)}}{\abs{\vocab(s)} + \sum_{\seqElement_{\predInd}' \in \vocab(s)} n(s, \seqElement_{\predInd}')}.
\end{align}
We use the version of \textsc{PPM-C} that excludes already-matched symbols from probability calculations at lower-order contexts \citep{1990_ppm_c}.
Since \textsc{PPM-C} is not designed to handle sampling times, we disregard them and use the last $C$ ordered \seqElementText{}s as context window for prediction.

\subsection{Gradient-Boosted Decision Trees}
\textsc{XGBoost} \citep{2016_xgboost} and \textsc{CatBoost} \citep{2018_catboost} are tree-based boosting algorithms.
Thus, they need tabular input data, which we achieve by adding one sample (row) for each predicted \seqElementText{} $\seqElement_\predInd$.
A tabular sample contains all \seqElementText{}s $\seqElement_\seqInd$ that lie in the lookback window $\maxLookback$ of \ourMethod ($\seqTime_\predInd - \seqTime_\seqInd \leq \maxLookback$) and the label is the predicted \seqElementText{}~($\seqElement_\predInd$).
Specifically, we create one categorical feature (column) for each of the positional offsets $\predInd - \seqInd$ that takes the value $\seqElement_\seqInd$.
For irregularly sampled data with varying offsets $\seqTime_\predInd - \seqTime_\seqInd$, we additionally include these offsets as separate numerical features.
We ensure that all samples contain the same number of features by introducing a padding class for categorical features, and padding offsets with $\maxLookback$.

\subsection{RoPE Transformer}
Transformers \citep{2017_attention_is_all_you_need} are well suited for sequential prediction tasks.
We adopt a decoder-only architecture with causal (autoregressive) attention masking.
Each decoder layer consists of a multi-head self-attention block followed by a position-wise feed-forward network. 
Both sub-layers are wrapped with residual (skip) connections and layer normalization \citep{2016_layer_norm}. 
Throughout the model, we use GELU activation functions \citep{2016_gelu} and apply dropout \citep{2014_dropout} with rate $0.1$.

To encode temporal information, we employ rotary position embeddings (RoPE) \citep{2024_roformer} within the self-attention mechanism.
RoPE injects relative positional information by applying position-dependent rotations to the query and key vectors.
While the original formulation of RoPE was designed for NLP token sequences and uses index-based positional offsets $(\predInd - \seqInd)$, we generalize this approach to continuous time by instead using temporal offsets $(\seqTime_{\predInd} - \seqTime_{\seqInd})$.
Consistent with the autoregressive formulation, attention is masked to prevent access to future information.
In addition, we restrict attention in each layer to the same fixed context used in \ourMethod{} by masking out \seqElementText{}s $\seqElement_{\seqInd}$ whose temporal distance from the prediction point exceeds the maximum lookback window $\maxLookback$, i.e., $\seqTime_\predInd - \seqTime_\seqInd > \maxLookback$.

\section{Sequential Rule Interpretations and Limitations of Markov Models and Trees}
\label{app:sec_interpreting_models_as_rules}
\begin{figure*}[ht]
    \centering
    \begin{tikzpicture}[
        scale=0.83, transform shape,
        node distance=0.1cm and 0.1cm,
        event/.style n args={1}{rectangle, rounded corners, draw=#1, fill=#1!40, thick, minimum size=4mm, font=\footnotesize},
        feature/.style={below=2pt, font=\scriptsize},
        tree_node/.style={rectangle, draw=black, thick, minimum size=3mm, font=\scriptsize}
    ]

        \node[font=\footnotesize] (sequence) {\SeqElementText{}s:};
        \node[font=\footnotesize] (data) at ($(sequence) + (1.5, 0.8)$) {\textbf{Data}};
        \node[event={blue},   right=of sequence] (e1) {\vphantom{?}a};
        \node[event={orange}, right=of e1]       (e2) {\vphantom{?}b};
        \node[event={blue},   right=of e2]       (e3) {\vphantom{?}a};
        \node[event={gray},   right=of e3]       (e4) {?};
    
        \node[feature] (x1) at (e1.south) {$\seqElement_1$};
        \node[feature] (x2) at (e2.south) {$\seqElement_2$};
        \node[feature] (x3) at (e3.south) {$\seqElement_3$};
        \node[feature] (x3) at (e4.south) {$\seqElement_4$};

        \node[below=0.6cm of sequence, font=\footnotesize] (time) {Time:};
        \node[event={blue},   below=0.6cm of e1] (t1) {1};
        \node[event={orange}, right=of t1]       (t2) {5};
        \node[event={blue},   right=of t2]       (t3) {8};
        \node[event={gray},   right=of t3]       (t4) {9};
    
        \node[feature] (tf1) at (t1.south) {$\seqTime_1$};
        \node[feature] (tf2) at (t2.south) {$\seqTime_2$};
        \node[feature] (tf3) at (t3.south) {$\seqTime_3$};
        \node[feature] (tf4) at (t4.south) {$\seqTime_4$};

        \node[font=\footnotesize] (rules) at ($(time) + (1.8, -1.6)$) {Sequential Rules:};

        \node[font=\footnotesize] (markov_model) at ($(sequence) + (5.4, 0.8)$) {Markov Models};
        \node[below=of markov_model, font=\scriptsize] (markov_table) {
            \begin{tabular}{r ccc}
                \toprule
                          & \multicolumn{3}{c}{Counts} \\ \cmidrule{2-4}
                  Context                                    & a & b & c \\ \midrule
                  \textcolor{orange}{b} \textcolor{blue}{a}  & \textcolor{black!30}{1} & 3 & \textcolor{black!30}{2} \\
                   \textcolor{blue}{a}                       & 5 & 5 & \textcolor{black!30}{2} \\
                \bottomrule
            \end{tabular}
          };

        \node[below=2.8cm of markov_model,font=\scriptsize, style={align=center}] (markov_rules) {
            $(\seqElement_2 = \text{\textcolor{orange}{b}})\land (\seqElement_3 = \text{\textcolor{blue}{a}})\ruleTo (\seqElement_4 = \text{\textcolor{orange}{b}})$ \\
            $(\seqElement_3 = \text{\textcolor{blue}{a}})\ruleTo (\seqElement_2 = \text{\textcolor{blue}{a}})\lor (\seqElement_4 = \text{\textcolor{orange}{b}})$
        };

        \draw[black] ($(sequence) + (3.3, 1.0)$) -- ($(sequence) + (3.3, -3.0)$);

        \node[font=\footnotesize, right=3.3cm of markov_model] (trees) {Trees};
        
        \node[tree_node, below=of trees] (tree1) {$\seqElement_2\in \set{\textcolor{blue}{a},\textcolor{orange}{b}}$};
        \node[tree_node] (tree2) at ($(tree1) - (1.5,0.7)$) {$\seqTime_4 - \seqTime_1 < 7$};
        \node[font=\scriptsize] (tree3) at ($(tree1) - (-1.5,0.85)$) {$\vdots$};
        \node[font=\scriptsize] (tree4) at ($(tree2) - (1.0,0.85)$) {$\vdots$};
        \node[font=\tiny] (tree5) at ($(tree2) - (-1.0,1.0)$) {
            \begin{tabular}{c ccc}
                \toprule
                         & a & b & c \\ \midrule
                  Weights & \textcolor{black!30}{1} & \textcolor{black!30}{2} & 3 \\
                \bottomrule
            \end{tabular}};

        \draw[thick] (tree1.south) -- (tree2.north);
        \draw[thick] (tree1.south) -- (tree3.north);
        \draw[thick] (tree2.south) -- (tree4.north);
        \draw[thick] (tree2.south) -- (tree5.north);

        \node[font=\scriptsize] at ($(tree1)!0.5!(tree2) + (-0.4, 0.1)$) {$\checkmark$};
        \node[font=\scriptsize] at ($(tree1)!0.5!(tree3) + (0.45, 0.15)$) {\ding{55}};

        \node[font=\scriptsize] at ($(tree2)!0.5!(tree4) + (-0.5, 0.1)$) {$\checkmark$};
        \node[font=\scriptsize] at ($(tree2)!0.5!(tree5) + (0.5, 0.2)$) {\ding{55}};

        \node[below=2.8cm of trees,font=\scriptsize] (tree_rules) {
            $(\seqElement_2 \in \set{\textcolor{blue}{a},\text{\textcolor{orange}{b}}})\land (\seqTime_4 - \seqTime_1 \geq 7)\ruleTo (\seqElement_4 = \text{\textcolor{purple}{c}})$
        };

        \draw[black] ($(markov_model)!0.5!(trees) + (-0.3, 0.2)$) -- ($(markov_model)!0.5!(trees) + (-0.3, -3.8)$);

        \node[font=\footnotesize, right=3.5cm of trees] (ours) {Ours};

        \node[circle, black, draw, minimum size=5] (i1) at ($(ours) + (-1.2, -0.6)$) {};
        \node[circle, black, draw, minimum size=5] (i2) at ($(i1) + (0, -0.8)$) {};
        \node[circle, black, draw, minimum size=5, fill=orange!70] (i3) at ($(i2) + (0, -0.8)$) {};

        \node[circle, black, draw, minimum size=5] (h1) at ($(ours) + (0, -1.0)$) {};
        \node[circle, black, draw, minimum size=5] (h2) at ($(h1) + (0, -0.8)$) {};

        \node[circle, black, draw, minimum size=5] (o1) at ($(ours) + (1.2, -0.6)$) {};
        \node[circle, black, draw, minimum size=5] (o2) at ($(o1) + (0, -0.8)$) {};
        \node[circle, black, draw, minimum size=5, fill=purple!70] (o3) at ($(o2) + (0, -0.8)$) {};

        \foreach \i in {1,2,3} {
            \draw[black!50] (i\i.east) -- (h1.west);
        }
        \draw[black!50] (i1.east) -- (h2.west);
        \draw[black!50] (i2.east) -- (h2.west);
        \draw[black,thick] (i3.east) -- (h2.west);

        \foreach \i in {1,2} {
             \foreach \j in {1,2} {
                \draw[black!50] (h\i.east) -- (o\j.west);
             }
        }
        
        \draw[black!50] (h1.east) -- (o3.west);
        \draw[black,thick] (h2.east) -- (o3.west);

        \node[below=2.8cm of ours,font=\scriptsize, style={align=center}] (our_rules) {
            $(\seqElement_2 = \text{\textcolor{orange}{b}})\land (\seqTime_4 - \seqTime_2 \in [3,5])$ \\ $\ruleTo (\seqElement_4 = \text{\textcolor{purple}{c}})$
        };
        \draw[-{Stealth}, black] ($(h2.south) - (0, 0.1)$) -- ($(our_rules.north) - (0, 0.1)$);
        \node[font=\scriptsize, align={center}] at ($(h2.south)!0.5!(our_rules.north) + (0.4,0)$) {read \\ out};

        \draw[black] ($(trees)!0.5!(ours) + (0.4, 0.2)$) -- ($(trees)!0.5!(ours) + (0.4, -3.8)$);
        \draw[black, dashed] ($(data) + (-1.8,-2.9)$) -- ($(ours) + (1.5,-2.9)$);
    
    \end{tikzpicture}
    \caption{
        \textit{Different models using symbolic rules (denoted by $\Rightarrow$) to predict the next \seqElementText{} $\seqElement_4$}.
        Markov models base predictions on empirical event counts following a fixed-order context and do not capture time information beyond order.
        Decision trees define prediction rules through split conditions and leaf weights.
        While they can incorporate time features, these are not explicitly tied to specific \seqElementText{}s (e.g., $\seqElement_2$ vs.\ $\seqTime_1$), and splits for categorical variables often rely on subset partitions (e.g., $\seqElement_2 \in \set{\textcolor{blue}{a},\text{\textcolor{orange}{b}}}$).
        In contrast, \ourMethod learns compact neuro-symbolic rules that directly associate specific events with delay conditions (e.g., $\seqElement_2$ and $\seqTime_2$), enabling transparent and temporally grounded predictions.
    }
    \label{fig:visualization}
\end{figure*}
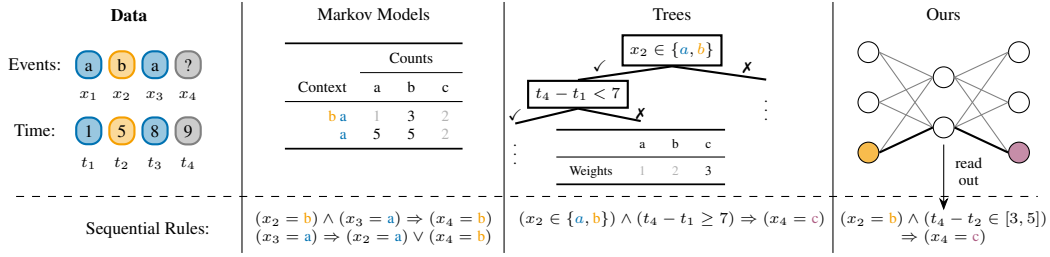

In \cref{fig:visualization}, we illustrate how different models can be interpreted as sequential rule sets.
\textbf{Markov models} are defined by tables that record empirical next-\seqElementText{} frequencies for each observed context of a given maximum length.
At prediction time, the current sequence is matched to a learned context, and the empirical frequency vector defines the predictive distribution.
Accordingly, each event in the context acts as an antecedent element (event condition), while the empirical counts implicitly define the rule consequent.
Unlike \ourMethod{}, Markov models cannot incorporate time information beyond event order and grow exponentially with context length, restricting them to short-range dependencies.

\textbf{Decision trees} define symbolic conditions via their split nodes.
Therefore, each root-to-leaf path forms a conjunction of conditions on the input, constituting a rule antecedent, and the class logits at the leaf define the consequent.
While trees can incorporate event-time features, these are not explicitly tied to specific \seqElementText{}s, and splits for high-cardinality categorical variables are typically implemented as subset partitions, enabling more compact tree representations but reducing interpretability.

\section{Real-World Datasets}
\label{app:sec_real_world_dataset_stats}
In this section, we provide details on the real-world datasets:
\begin{itemize}
    \item \textbf{Holidays} contains a single long sequence of public holidays in a European country with the time steps between them measured in days.
    \item \textbf{Chorales} \citep{1966_bach_chorales,uci_ml_repository} consists of 100 Bach chorales in MIDI format, where events correspond to note pitches and time is measured in 16th notes from the beginning of a chorale (CC BY 4.0 license).
    \item \textbf{Commits} \citep{2018_mdl_peridoic_samba,2024_hopper} lists the authors of commits made to the Samba file system repository (\url{https://git.samba.org/}) between 1996 and 2016. The temporal component represents the order of commits (Apache License 2.0). 
    \item \textbf{Rolling Mill} \citep{2021_rolling_mill_graphs,2024_hopper} comprises 1000 production logs from a steel manufacturing plant, documenting the  order of manufacturing events (public, no explicit license).
    \item \textbf{Amazon} \citep{2024_amazon_dataset} contains over 1.7 million purchases made on Amazon by around 5000 customers. 
    Here we use the product categories as \seqElementText{}s and the purchase days as occurrence times (CC0 1.0 license).
\end{itemize}

In \cref{tab:exp_real_data_stats}, we show additional dataset statistics. 
We refer to a dataset as \textbf{regularly sampled}, when all consecutive \seqElementText{}s have a constant offset $\seqTime_{\seqInd + 1} - \seqTime_{\seqInd}$. 
In this case, event times (sampling times) are effectively equivalent to sequence positions and we set $\seqTime_{\seqInd + 1} - \seqTime_{\seqInd} = 1$ without loss of generality.
To create train, validation, and test data, we split Holidays and Commits via cutoff dates to reflect a forecasting scenario, where models are trained on past data to predict future outcomes.
For the other datasets, we randomly assign sequences to splits.

\begin{table}[hb]
    \centering
    \caption{
        \textit{Statistics of the real-world datasets.}
        Columns show: Number of sequences ($\numSamples$), total number of \seqElementText{}s ($\seqLen_\mathrm{tot}$), minimum and maximum sequence lengths ($\seqLen_\mathrm{min}$ and $\seqLen_\mathrm{max}$), vocabulary size ($\vocabSize$), and whether sequences are regularly sampled.
    }
    \label{tab:exp_real_data_stats}
    {\setlength{\tabcolsep}{10pt}
    \begin{tabular}{l rrrrr c}
        \toprule
        \textbf{Datasets}   & $\numSamples$    & $\seqLen_\mathrm{tot}$ & $\seqLen_\mathrm{min}$ & $\seqLen_\mathrm{max}$ & $\vocabSize$    & Regular \\
        \midrule
        Holidays            &                1 &                   1200 &                     -- &                     -- &              14 & \ding{55}       \\
        Chorales            &              100 &                   4693 &                     26 &                     86 &              20 & \ding{55}       \\
        Commits             &                1 &                  28879 &                     -- &                     -- &             119 & $\checkmark$    \\
        Rolling Mill        &             1000 &                  53788 &                     24 &                    142 &             556 & $\checkmark$    \\
        Amazon              &   4909 &         1760607 &               10 &             5057 &    993 & \ding{55}       \\
        \bottomrule
    \end{tabular}
    }
\end{table}

To efficiently train models on datasets with long sequences (Holidays, Commits, Amazon), we use subsequences to create training samples.
For neural models, this is done randomly during batching, while for the others, we handle it in the pre-processing of the data.
For testing on these datasets, we follow the same strategy, ensuring that all models are tested on the same subsequences.

\paragraph{Pre-processing of the Amazon Dataset.} 
We pre-process the Amazon dataset as follows.
First, we remove sequences (customers) with fewer than $10$ purchases.
We then replace all product categories occurring fewer than $200$ times with a single \texttt{OTHER} category.
To resolve multiple purchases made on the same day, we impose an arbitrary but fixed alphabetical ordering over categories and introduce a small time offset $\varepsilon$ between successive purchases, chosen such that no two purchase days overlap.
Finally, we normalize all timestamps by dividing them by $5$ to facilitate the training of \ourMethod{} and \ourMethodSideChannel{}.

\section{Hardware and Compute Time}
\label{app:sec_hardware_and_compute}
\subsection{Hardware}
All experiments are conducted on a Linux-based compute cluster with NVIDIA A100 (40GB and 80GB) and L40S GPUs, paired with AMD EPYC 7662 and 9654 CPUs. Neural models are implemented in PyTorch with CUDA~12.4. All models are trained on a single GPU, except for the \textsc{PPM-C} baseline, which is trained on CPU.
Most experiments use 32GB of RAM, while rule extraction from tree ensembles requires up to 400GB (\textsc{XGBoost} on Amazon).

\subsection{Compute Time} 
We did not track total compute precisely; instead, we report estimates based on recorded runtimes of individual runs.

\paragraph{Synthetic Data.}
On synthetic data, \ourMethod{} requires about 15--20 minutes per training run (including temperature annealing), with negligible overhead for rule extraction and refinement. Other models typically train in under 2 minutes. 
Each synthetic setting includes 10 datasets, with four random seeds for \ourMethod{}, $\ourMethod^*$, and the transformer; and grid search for \textsc{PPM}, \textsc{XGBoost}, and \textsc{CatBoost}. This results in an approximate upper bound of one day of compute per setting. 
Across 28 synthetic settings, this corresponds to an upper bound of roughly 3--4 weeks of GPU-equivalent compute.

\paragraph{Real Data.}
On Holidays, Chorales, and Commits, all models (including grid search and multiple seeds) complete within about 1 hour. 
On Rolling Mill,  grid search for \textsc{XGBoost} takes $<2$ hours and for \textsc{CatBoost} $<3$ hours; all other models combined complete within 5 hours. 
On Amazon, the most time-consuming models per training run are \textsc{CatBoost} with about 4.5 hours, \textsc{XGBoost} with about 45 minutes, and \ourMethod{} with about 30 minutes.
Excluding tree ensembles, the remaining models together require less than 3 hours.
Exact rule extraction for tree ensembles is computationally expensive on Rolling Mill and Amazon, particularly for \textsc{XGBoost} (multiple hours).

\paragraph{Additional Compute.}
Hyperparameter tuning and preliminary experiments required additional compute beyond the reported runs; we estimate this overhead to be approximately 1--2$\times$ the compute of the reported results.

\section{Hyperparameters} 
\label{app:sec_hyperparameters}

\subsection{Synthetic Datasets}
\label{app:subsec_hyperparams_synthetic}

We set the maximum lookback window for all methods to the known maximum influence of $\maxLookback = 64$.
For \ourMethod{} and the RoPE Transformer, we manually tune hyperparameters on the default setting ($\numSamples = 1024$, $\vocabSize = 128$, $\fracTarg = 0.4$, $\fracCons = 0.85$) and fix them across all experimental settings.
We prioritize training stability, followed by minimizing validation loss, while for \ourMethod{}, we additionally consider ground-truth recovery performance.
For \textsc{PPM}, \textsc{XGBoost}, and \textsc{CatBoost}, we perform a dataset-specific grid search, select the best-performing model per dataset, and aggregate the resulting metrics across datasets within each synthetic setting.

\paragraph{\ourMethodTitle{}.}
For \ourMethod{}, we use $\countInNeurons = 300$ temporal predicate neurons in the first hidden layer and $\countAntNeurons = 64$ antecedent neurons in the second hidden layer, ensuring sufficient capacity to learn the ground-truth rule sets.
When computing the activation of temporal predicates (\cref{eq:pred_act}), we mask out $\tilde{\predAct}_i(\seqInd,\predInd)$ for all events $\seqElement_\seqInd$ outside the maximum lookback window ($\seqTime_\predInd - \seqTime_\seqInd > \maxLookback$).
We optimize the model using Adam \citep{2015_adam} with a learning rate of~$0.03$ and decay parameters $\beta_1 = 0.9$, $\beta_2 = 0.999$ on minibatches of size $64$.
We tested learning rate schedules, but since temperature annealing already reduces gradients in most regions, this did not improve training.
Finally, we clip all $\maxOffset_i$ after gradient updates to enforce a minimum interval width of 1 and $\maxOffset_i \leq \maxLookback$.

We initialize $\tau_1 =\tau_2 = 1$, and begin annealing $\tau_1$, the temperature of the temporal predicate neurons, after a fixed number of warm-up epochs.
For the default $1024$ training sequences, we use $60$ warm-up epochs, scaling them proportionally to the number of training sequences in other settings.
After warm-up, we update the first-layer temperature, $\tau_1$, every $32$ optimization steps using the multiplicative decay $\tau_1 \gets 0.9 \cdot \tau_1$.
When $\tau_1$ falls below $1 \cdot 10^{-4}$, we begin decaying the antecedent layer temperature $\tau_2$ at the same rate and step interval.
To avoid numerical instability, we stop annealing $\tau_1$ once it drops below $1 \cdot 10^{-6}$, and we terminate training when $\tau_2$ falls below $1 \cdot 10^{-5}$.
After training, we use the symbolic version of the hidden layers and prune always-on neurons that activate for more than $99.9\%$ of target \seqElementText{}s and always-off neurons that activate for less than $0.1\%$ of target \seqElementText{}s (see Appendix~\ref{app:subsec_antecedent_pruning}, Step 1).

For each training dataset, we train \ourMethod{} using four random seeds and select the best-performing model.
Because accurate rule recovery depends on faithfully fitting the training data, we select models based on training loss rather than validation loss.
Empirically, models with lower training loss also tend to generalize better.

\paragraph{Rule Sets.}
To discretize the softmax layer of the best-performing \ourMethod{} model into a rule set (Appendices \ref{app:subsec_rule_set_extraction} and~\ref{app:subsec_antecedent_pruning}), we set the min-max variation threshold $\threshold_{\mathrm{var}} = 0.9$, the quantile threshold $\threshold_{\mathrm{quant}} = 0.95$, and the inclusion threshold $\threshold_{\mathrm{inc}} = 0.95$.
Except for the sanity check on random data, we discard rules that are either unreliable (confidence below $0.05$ on the training data) or extremely rare (frequency below $5\cdot 10^{-4}$ on the training data) before refining rule sets (Appendix~\ref{app:subsec_symbolic_rule_refinement}) and computing rule-set–related evaluation metrics.
This simple filtering step focuses the evaluation on rules that both apply non-trivially to the data and make consistent predictions, providing a more realistic aggregation of rule quality across multiple datasets.

\paragraph{\ourMethodTitle{} with Embedded Ground Truth ($\text{\ourMethod{}}^*$).}
For \ourMethod{} with embedded ground-truth rules, hidden neurons are determined by the rules, so we only optimize the softmax output layer.
We optimize with Adam (learning rate $0.03$), using the same hyperparameters as the standard \ourMethod{} model, and apply a staged schedule: $10$ warm-up epochs, followed by learning-rate decay by a factor of $0.3$ whenever the training loss fails to improve by more than $0.01$ over $10$ epochs.
We apply early stopping if the training loss does not improve by at least $1\cdot 10^{-3}$ within $31$ epochs. 
Models are checkpointed after each epoch, and we select the checkpoint with the lowest training loss.
Training is repeated for four random seeds, and we select the best resulting model.

\paragraph{RoPE Transformer.}
We use a transformer with four decoder layers, each with four attention heads and layer normalization \citep{2016_layer_norm}.
We fix the \seqElementText{} embedding dimension and the dimensions of query, key, and value vectors to $128$.
The feedforward network in each decoder layer has one hidden layer of size $256$.
The attention modules use rotary position embeddings \citep{2024_roformer} based on the relative temporal distances between \seqElementText{}s.
In the attention matrix, we mask out \seqElementText{}s outside the maximum lookback window $\maxLookback$ used by \ourMethod{}.
The final feedforward network predicting the next \seqElementText{} from the decoder embeddings has one hidden layer of size $128$.
We use the GELU activation function \citep{2016_gelu} in all hidden layers of the feedforward networks.
During training, we apply dropout of $0.1$ in the attention matrix, on attention outputs, within each decoder-layer feedforward network, and in the hidden layer of the output network.

We optimize the transformer using Adam \citep{2015_adam} with an initial learning rate of~$3\cdot 10^{-4}$ and decay parameters $\beta_1 = 0.9$, $\beta_2 = 0.999$.
We set the batch size to $64$ and use the following learning rate schedule:
We start with a learning rate of $3\cdot 10^{-4}$, multiplying it by $0.1$ whenever the training loss fails to improve by more than $0.01$ over ten epochs.
Early stopping is applied by monitoring the validation loss, stopping training if it does not decrease within $31$ epochs.
We record the model after each epoch and select the version with the lowest validation loss.
As with \ourMethod{}, we repeat training with four different random seeds and then select the best-performing model based on the validation cross-entropy loss.

\paragraph{PPM-C.}
For \textsc{PPM}, we try maximum Markov context lengths $C \in\{0,1,2,3,4,5\}$ on each dataset and then select the best model according to the validation cross-entropy loss.

\paragraph{\textsc{XGBoost}.}
We run \textsc{XGBoost} on GPU for up to $300$ boosting rounds, applying early stopping if the validation cross-entropy loss does not improve for $30$ rounds and selecting the iteration with the lowest validation loss.
The subsampling ratio of training instances is set to $0.6$ for each boosting round.
We perform a grid search over the maximum tree depth~$\{4,6\}$, L2 leaf weight regularization $\{0,1,3,5\}$, and learning rate $\{0.03,0.3\}$, repeating the search over two random seeds and selecting the model with the overall lowest validation loss.

\paragraph{\textsc{CatBoost}.}
We run \textsc{CatBoost} on GPU with a subsampling ratio of $0.6$, which is applied per split rather than per tree as in \textsc{XGBoost}, and otherwise use the same hyperparameters and grid search strategy as for \textsc{XGBoost}.

\subsection{Real-World Datasets}
\label{app:subsec_hyperparams_real}
For real-world datasets, we broadly follow the same training protocol as in the synthetic experiments.
We fix the lookback window $\maxLookback$ across models by tuning it for \ourMethod{} and applying the selected value to all baselines.

\paragraph{\ourMethodTitle{}.}
Unless stated otherwise, we use the same learning rates and annealing schedules as in the synthetic experiments (Appendix~\ref{app:subsec_hyperparams_synthetic}), which we found to work well across most real datasets.
One exception is the number of warm-up epochs before temperature annealing, which we tune per dataset.
On Holidays, where we observe a pronounced mismatch between the soft and symbolic hidden layers, we anneal the second-layer temperature $\tau_2$ until it reaches $1\cdot 10^{-6}$ rather than $1\cdot 10^{-5}$.
On Amazon, we anneal the first-layer temperature $\tau_1$ with a factor of $0.92$ and the second-layer temperature with a factor of $0.95$.
To counteract excessive deactivation of antecedent neurons during annealing on Amazon, we further add a regularization term after the warm-up epochs that subtracts the batch-wise mean antecedent activation from the loss.

The remaining hyperparameters are reported in \cref{tab:exp_real_data_hyperparams}.
For Chorales, we use a batch size of $30$ due to the limited number of training sequences ($60$).
For Holidays, Commits, and Amazon, training is performed on randomly sampled subsequences; we define one epoch as $1024$ samples for Holidays and Commits and $2048$ samples for Amazon.
For Amazon, event times are divided by $5$, resulting in an effective lookback window of $100$ days.
We set the min-max variation threshold to $\threshold_{\mathrm{var}} = 0.9$ across all datasets.

\begin{table}[!ht]
    \centering
    \caption{
        \textit{Selected hyperparameters of \ourMethod{} on real-world datasets.}  
        Columns show: $\countInNeurons$ (number of temporal predicate neurons),  
        $\countAntNeurons$ (number of antecedent neurons),  
        $\maxLookback$ (maximum lookback window),  
        batch size,  
        warm-up epochs,  
        $\threshold_{\mathrm{quant}}$ (quantile threshold), and  
        $\threshold_{\mathrm{inc}}$~(inclusion threshold).
    }
    \label{tab:exp_real_data_hyperparams}
    {\setlength{\tabcolsep}{10pt}
    \begin{tabular}{l rrr rr rr}
        \toprule
        \textbf{Datasets}   & $\countInNeurons$ & $\countAntNeurons$ & $\maxLookback$   & Batch Size & Warm-up & $\threshold_{\mathrm{quant}}$ & $\threshold_{\mathrm{inc}}$ \\
        \midrule
        Holidays            &               200 &                 45 &              365 &         64 &     100 &                          0.85 &                       0.85  \\
        Chorales            &               200 &                 64 &               48 &         30 &     300 &                          0.85 &                       0.85  \\
        Commits             &               300 &                100 &               24 &         64 &      60 &                          0.95 &                       0.90  \\
        Rolling Mill        &               500 &                200 &               64 &         64 &     100 &                          0.90 &                       0.90  \\
        Amazon              &               500 &                200 &               20 &         64 &     200 &                          0.80 &                       0.90  \\    
        \bottomrule
    \end{tabular}
    }
\end{table}

After fixing hyperparameters, we train \ourMethod{} with four random seeds and, unlike in the synthetic experiments, select the best model based on validation loss.
As described in \cref{subsec:experiments_real}, we extract and post-process the learned rules (Appendices~\ref{app:subsec_rule_set_extraction} to~\ref{app:subsec_symbolic_rule_refinement}), re-embed the refined antecedents into the hidden layers, and retrain only the output layer.
Details of the retraining procedure are provided in the next paragraph.

\paragraph{\ourMethodTitle{} with Processed Rules.}
After embedding symbolic antecedents in \ourMethod{}, we retrain only the softmax output layer.
We use the same strategy as for \ourMethod{} with embedded ground-truth antecedents ($\text{\ourMethod{}}^*$) in the synthetic experiments, but select the best model based on validation loss:
We optimize with Adam (learning rate $0.03$), using the same batch size as the end-to-end \ourMethod{} model, and apply a staged schedule: $10$ warm-up epochs, followed by learning-rate decay by a factor of $0.3$ whenever the training loss fails to improve by more than $0.01$ over $10$ epochs.
We apply early stopping if the validation loss does not improve by at least $1\cdot 10^{-3}$ within $31$ epochs. 
Models are checkpointed after each epoch, and we select the checkpoint with the lowest validation loss.
Training is repeated for four random seeds, and we select the best resulting model.

\paragraph{RoPE Transformer.}
We use the same hyperparameters and training schedule as in the synthetic setting, adapting only the batch size and maximum lookback window to match \ourMethod{}.

\paragraph{\ourMethodSideChannelTitle{}.}
For \ourMethodSideChannel{}, we combine the best (frozen) \ourMethod{} model with processed rule antecedents and a transformer residual network.
The transformer uses the same architecture and optimization protocol as the standalone RoPE transformer, but the softmax logits of \ourMethod{} are added to the transformer logits (see \cref{subsec:architecture}).

\paragraph{PPM-C.}
For \textsc{PPM}, we follow the same protocol as in the synthetic experiments, evaluating context lengths $C \in \{0,1,2,3,4,5\}$ and selecting the best model based on validation cross-entropy loss.

\paragraph{\textsc{XGBoost} and \textsc{CatBoost}.}
For datasets in which \ourMethod{} is trained on randomly sampled subsequences (Holidays, Commits, Amazon), we augment the tree-ensemble training and validation data with \seqElementText{}s (samples) by randomly truncating the sequence history beyond the maximum lookback window $\maxLookback$, matching the incomplete context of some \seqElementText{}s in test subsequences.
We use the same hyperparameters and grid-search protocol as for synthetic data, except on Amazon.
Due to its scale, we restrict the Amazon grid search to maximum tree depth $\{6\}$, L2 leaf regularization $\{0,3,5\}$, and learning rate $\{0.003,0.03\}$.
We further reduce the subsampling ratio to $0.1$, early stopping rounds to $6$, and run grid search with a single random seed.

\clearpage
\section{Additional Results on Synthetic Data}
\label{app:sec_results_synthetic}
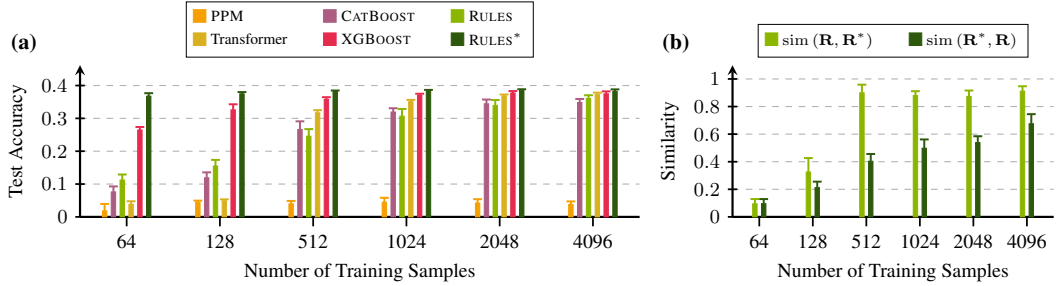
\begin{figure*}[hb]
    \centering
    \begin{tikzpicture}[scale=0.83, transform shape]
    
        \begin{axis}[
            width=10.5cm,
            height=4cm,
            clip=false,
            axis lines=left,
            axis line style={line width=1pt},
            x axis line style={-},
            tick style={line width=0.8pt, black},
            tick label style={font=\small},
            label style={font=\small},
            ymajorgrids,
            grid style={black!30,line width=0.5pt,dashed},
            ybar,
            ymin=0,
            enlarge x limits=0.1,
            enlarge y limits={upper=0.1},
            yticklabel style={/pgf/number format/fixed},
            xlabel={Number of Training Samples},
            ylabel={Test Accuracy},
            xtick=data,
            xticklabels={},
            xticklabels={64,128,512,1024,2048,4096},
            xticklabel style={font=\small},
            ymax=0.42,
            ytick={0.0,0.1,0.2,0.3,0.4},
            scaled y ticks=false,
            legend columns=3,
            legend style={
                font=\scriptsize,
                legend cell align=left,
                at={(0.5,1.05)},
                anchor=south,
                /tikz/every even column/.append style={column sep=0.5cm}
            },
            legend image code/.code={\draw [#1] (0cm,-0.075cm) rectangle (0.2cm,0.125cm);}
        ]

            \addplot+[
                fill=orange, draw=orange, bar width=2pt, error bars/.cd, y dir=plus, y explicit, error mark=|,
                error bar style={draw=orange,line width=0.9pt}, error mark options={draw=orange,line width=0.9pt}
            ] coordinates {(0,0.0185) +- (0,0.0210) (1,0.0442) +- (0,0.0053) (2,0.0398) +- (0,0.0088) (3,0.0446) +- (0,0.0135) (4,0.0422) +- (0,0.0117) (5,0.0380) +- (0,0.0092)};

            \addplot+[
                fill=purple, draw=purple, bar width=2pt, error bars/.cd, y dir=plus, y explicit, error mark=|,
                error bar style={draw=purple,line width=0.9pt}, error mark options={draw=purple,line width=0.9pt}
            ] coordinates {(0,0.0761) +- (0,0.0167) (1,0.1190) +- (0,0.0166) (2,0.2663) +- (0,0.0247) (3,0.3193) +- (0,0.0116) (4,0.3443) +- (0,0.0131) (5,0.3483) +- (0,0.0108)};

            \addplot+[
                fill=green, draw=green, bar width=2pt, error bars/.cd, y dir=plus, y explicit, error mark=|,
                error bar style={draw=green,line width=0.9pt}, error mark options={draw=green,line width=0.9pt}
            ] coordinates {(0,0.1123) +- (0,0.0168) (1,0.1548) +- (0,0.0188) (2,0.2458) +- (0,0.0215) (3,0.3067) +- (0,0.0220) (4,0.3395) +- (0,0.0164) (5,0.3602) +- (0,0.0102)};

            \addplot+[
                fill=yellow, draw=yellow, bar width=2pt, error bars/.cd, y dir=plus, y explicit, error mark=|,
                error bar style={draw=yellow,line width=0.9pt}, error mark options={draw=yellow,line width=0.9pt}
            ] coordinates {(0,0.0383) +- (0,0.0093) (1,0.0487) +- (0,0.0047) (2,0.3169) +- (0,0.0081) (3,0.3507) +- (0,0.0056) (4,0.3678) +- (0,0.0048) (5,0.3721) +- (0,0.0061)};

            \addplot+[
                fill=red, draw=red, bar width=2pt, error bars/.cd, y dir=plus, y explicit, error mark=|,
                error bar style={draw=red,line width=0.9pt}, error mark options={draw=red,line width=0.9pt}
            ] coordinates {(0,0.2646) +- (0,0.0092) (1,0.3259) +- (0,0.0170) (2,0.3582) +- (0,0.0063) (3,0.3700) +- (0,0.0050) (4,0.3772) +- (0,0.0059) (5,0.3748) +- (0,0.0074)};

            \addplot+[
                fill=darkgreen, draw=darkgreen, bar width=2pt, error bars/.cd, y dir=plus, y explicit, error mark=|,
                error bar style={draw=darkgreen,line width=0.9pt}, error mark options={draw=darkgreen,line width=0.9pt}
            ] coordinates {(0,0.3676) +- (0,0.0091) (1,0.3747) +- (0,0.0054) (2,0.3816) +- (0,0.0032) (3,0.3854) +- (0,0.0013) (4,0.3858) +- (0,0.0029) (5,0.3827) +- (0,0.0056)};

            \legend{\textsc{PPM},\textsc{CatBoost},\ourMethod{},Transformer,\textsc{XGBoost},$\text{\ourMethod{}}^*$}
            \node at (axis cs:\pgfkeysvalueof{/pgfplots/xmin},\pgfkeysvalueof{/pgfplots/ymax}) [anchor=east, xshift=-16pt, yshift=10pt] {\textbf{(a)}};
        \end{axis}

        \begin{axis}[
            at={(10.4cm,0)},
            width=6.7cm,
            height=4cm,
            clip=false,
            axis lines=left,
            axis line style={line width=1pt},
            x axis line style={-},
            tick style={line width=0.8pt, black},
            tick label style={font=\small},
            label style={font=\small},
            ymajorgrids,
            grid style={black!30,line width=0.5pt,dashed},
            ybar,
            ymin=0,
            enlarge x limits=0.1,
            enlarge y limits={upper=0.1},
            yticklabel style={/pgf/number format/fixed},
            xlabel={Number of Training Samples},
            ylabel={Similarity},
            xtick=data,
            xticklabels={},
            xticklabels={64,128,512,1024,2048,4096},
            xticklabel style={font=\small},
            ymax=1,
            ytick={0.0,0.2,0.4,0.6,0.8,1.0},
            scaled y ticks=false,
            legend columns=2,
            legend style={
                font=\scriptsize,
                legend cell align=left,
                at={(0.5,1.05)},
                anchor=south,
                /tikz/every even column/.append style={column sep=0.5cm}
            },
            legend image code/.code={\draw [#1] (0cm,-0.075cm) rectangle (0.2cm,0.125cm);}
        ]
        
           \addplot+[
                fill=green, draw=green, bar width=2pt, error bars/.cd, y dir=plus, y explicit, error mark=|,
                error bar style={draw=green,line width=0.9pt}, error mark options={draw=green,line width=0.9pt}
            ] coordinates {(0,0.0950) +- (0,0.0345) (1,0.3257) +- (0,0.1009) (2,0.8997) +- (0,0.0589) (3,0.8792) +- (0,0.0322) (4,0.8737) +- (0,0.0425) (5,0.9116) +- (0,0.0355)};

            \addplot+[
                fill=darkgreen, draw=darkgreen, bar width=2pt, error bars/.cd, y dir=plus, y explicit, error mark=|,
                error bar style={draw=darkgreen,line width=0.9pt}, error mark options={draw=darkgreen,line width=0.9pt}
            ] coordinates {(0,0.0962) +- (0,0.0331) (1,0.2124) +- (0,0.0432) (2,0.4034) +- (0,0.0528) (3,0.4981) +- (0,0.0635) (4,0.5388) +- (0,0.0460) (5,0.6755) +- (0,0.0693)};

            \legend{$\simRulesets\paren{\ruleset,\ruleset^*}$,$\simRulesets\paren{\ruleset^*,\ruleset}$}
            \node at (axis cs:\pgfkeysvalueof{/pgfplots/xmin},\pgfkeysvalueof{/pgfplots/ymax}) [anchor=east, xshift=-16pt, yshift=10pt] {\textbf{(b)}};
        \end{axis}

    \end{tikzpicture}
    \caption{
        \textit{\ourMethod{} improves with an increasing number of training samples $\numSamples$.} 
        We show the mean results over 10 rule sets and corresponding datasets per synthetic setting. 
        Error bars mark the empirical standard deviation.
        \textbf{(a)}~Test accuracy compared to baseline models.
        \textbf{(b)}~Similarity between the rule set $\ruleset$ discovered by \ourMethod{} and the ground truth $\ruleset^*$. 
    }
    \label{fig:synthetic_num_samples}
\end{figure*}

\begin{figure*}[hb]
    \centering
    \begin{tikzpicture}[scale=0.94, transform shape]
    
        \begin{axis}[
            width=15cm,
            height=5cm,
            clip=false,
            axis lines=left,
            axis line style={line width=1pt},
            x axis line style={-},
            tick style={line width=0.8pt, black},
            tick label style={font=\small},
            label style={font=\small},
            ymajorgrids,
            grid style={black!30,line width=0.5pt,dashed},
            ybar,
            ymin=0,
            enlarge x limits=0.1,
            enlarge y limits={upper=0.1},
            yticklabel style={/pgf/number format/fixed},
            xlabel={Signal-to-Noise Ratio $(\fracTarg, \fracCons)$},
            ylabel={Test Accuracy},
            xtick=data,
            xticklabels={(0.05, 0.1),(0.05, 0.3),(0.05, 0.6),(0.05, 0.85),(0.05, 1.0),(0.1, 0.3),(0.1, 0.6),(0.1, 0.85),(0.1, 1.0),(0.4, 0.6),(0.4, 0.85),(0.4, 1.0),(0.9, 1.0)},
            xticklabel style={rotate=45, anchor=east, font=\tiny},
            ymax=1,
            ytick={0.0,0.2,0.4,0.6,0.8,1.0},
            scaled y ticks=false,
            legend columns=3,
            legend style={
                font=\small,
                legend cell align=left,
                at={(0.5,1.05)},
                anchor=south,
                /tikz/every even column/.append style={column sep=0.5cm}
            },
            legend image code/.code={\draw [#1] (0cm,-0.075cm) rectangle (0.2cm,0.125cm);}
        ]
        
            \addplot+[
                fill=orange, draw=orange, bar width=2pt, error bars/.cd, y dir=plus, y explicit, error mark=|,
                error bar style={draw=orange,line width=0.9pt}, error mark options={draw=orange,line width=0.9pt}
            ] coordinates {(0,0.0101) +- (0,0.0009) (1,0.0123) +- (0,0.0015) (2,0.0132) +- (0,0.0012) (3,0.0138) +- (0,0.0017) (4,0.0158) +- (0,0.0022) (5,0.0163) +- (0,0.0019) (6,0.0143) +- (0,0.0026) (7,0.0169) +- (0,0.0023) (8,0.0166) +- (0,0.0026) (9,0.0371) +- (0,0.0089) (10,0.0446) +- (0,0.0135) (11,0.0412) +- (0,0.0083) (12,0.0818) +- (0,0.0254)};

            \addplot+[
                fill=purple, draw=purple, bar width=2pt, error bars/.cd, y dir=plus, y explicit, error mark=|,
                error bar style={draw=purple,line width=0.9pt}, error mark options={draw=purple,line width=0.9pt}
            ] coordinates {(0,0.0105) +- (0,0.0008) (1,0.0166) +- (0,0.0030) (2,0.0339) +- (0,0.0061) (3,0.0496) +- (0,0.0070) (4,0.0592) +- (0,0.0066) (5,0.0303) +- (0,0.0038) (6,0.0559) +- (0,0.0050) (7,0.0723) +- (0,0.0066) (8,0.0707) +- (0,0.0062) (9,0.2595) +- (0,0.0230) (10,0.3193) +- (0,0.0116) (11,0.3260) +- (0,0.0207) (12,0.7421) +- (0,0.0332)};

            \addplot+[
                fill=green, draw=green, bar width=2pt, error bars/.cd, y dir=plus, y explicit, error mark=|,
                error bar style={draw=green,line width=0.9pt}, error mark options={draw=green,line width=0.9pt}
            ] coordinates {(0,0.0102) +- (0,0.0008) (1,0.0233) +- (0,0.0021) (2,0.0477) +- (0,0.0046) (3,0.0625) +- (0,0.0063) (4,0.0723) +- (0,0.0074) (5,0.0354) +- (0,0.0062) (6,0.0737) +- (0,0.0048) (7,0.0828) +- (0,0.0061) (8,0.0890) +- (0,0.0040) (9,0.2417) +- (0,0.0396) (10,0.3067) +- (0,0.0220) (11,0.3177) +- (0,0.0207) (12,0.7425) +- (0,0.0251)};

            \addplot+[
                fill=yellow, draw=yellow, bar width=2pt, error bars/.cd, y dir=plus, y explicit, error mark=|,
                error bar style={draw=yellow,line width=0.9pt}, error mark options={draw=yellow,line width=0.9pt}
            ] coordinates {(0,0.0087) +- (0,0.0008) (1,0.0119) +- (0,0.0014) (2,0.0211) +- (0,0.0164) (3,0.0630) +- (0,0.0066) (4,0.0756) +- (0,0.0081) (5,0.0169) +- (0,0.0018) (6,0.0740) +- (0,0.0048) (7,0.0897) +- (0,0.0036) (8,0.0930) +- (0,0.0044) (9,0.3124) +- (0,0.0074) (10,0.3507) +- (0,0.0056) (11,0.3625) +- (0,0.0088) (12,0.7804) +- (0,0.0127)};

            \addplot+[
                fill=red, draw=red, bar width=2pt, error bars/.cd, y dir=plus, y explicit, error mark=|,
                error bar style={draw=red,line width=0.9pt}, error mark options={draw=red,line width=0.9pt}
            ] coordinates {(0,0.0209) +- (0,0.0019) (1,0.0463) +- (0,0.0036) (2,0.0604) +- (0,0.0040) (3,0.0729) +- (0,0.0045) (4,0.0813) +- (0,0.0063) (5,0.0837) +- (0,0.0034) (6,0.0934) +- (0,0.0041) (7,0.0976) +- (0,0.0031) (8,0.0975) +- (0,0.0036) (9,0.3462) +- (0,0.0100) (10,0.3700) +- (0,0.0050) (11,0.3744) +- (0,0.0112) (12,0.8153) +- (0,0.0125)};

            \addplot+[
                fill=darkgreen, draw=darkgreen, bar width=2pt, error bars/.cd, y dir=plus, y explicit, error mark=|,
                error bar style={draw=darkgreen,line width=0.9pt}, error mark options={draw=darkgreen,line width=0.9pt}
            ] coordinates {(0,0.0438) +- (0,0.0006) (1,0.0555) +- (0,0.0010) (2,0.0661) +- (0,0.0042) (3,0.0783) +- (0,0.0056) (4,0.0863) +- (0,0.0070) (5,0.0985) +- (0,0.0015) (6,0.1020) +- (0,0.0016) (7,0.1038) +- (0,0.0013) (8,0.1044) +- (0,0.0015) (9,0.3686) +- (0,0.0046) (10,0.3854) +- (0,0.0013) (11,0.3861) +- (0,0.0076) (12,0.8342) +- (0,0.0053)};

            \legend{\textsc{PPM},\textsc{CatBoost},\ourMethod{},Transformer,\textsc{XGBoost},$\text{\ourMethod{}}^*$}
            \node at (axis cs:\pgfkeysvalueof{/pgfplots/xmin},\pgfkeysvalueof{/pgfplots/ymax}) [anchor=east, xshift=-16pt, yshift=10pt] {\textbf{(a)}};
        \end{axis}

         \begin{axis}[
            at={(0,-5.5cm)},
            width=15cm,
            height=4cm,
            clip=false,
            axis lines=left,
            axis line style={line width=1pt},
            x axis line style={-},
            tick style={line width=0.8pt, black},
            tick label style={font=\small},
            label style={font=\small},
            ymajorgrids,
            grid style={black!30,line width=0.5pt,dashed},
            ybar,
            ymin=0,
            enlarge x limits=0.1,
            enlarge y limits={upper=0.1},
            yticklabel style={/pgf/number format/fixed},
            xlabel={Signal-to-Noise Ratio $(\fracTarg, \fracCons)$},
            ylabel={Similarity},
            xtick=data,
            xticklabels={(0.05, 0.1),(0.05, 0.3),(0.05, 0.6),(0.05, 0.85),(0.05, 1.0),(0.1, 0.3),(0.1, 0.6),(0.1, 0.85),(0.1, 1.0),(0.4, 0.6),(0.4, 0.85),(0.4, 1.0),(0.9, 1.0)},
            xticklabel style={rotate=45, anchor=east, font=\tiny},
            ymax=1,
            ytick={0.0,0.2,0.4,0.6,0.8,1.0},
            scaled y ticks=false,
            legend columns=2,
            legend style={
                font=\small,
                legend cell align=left,
                at={(0.5,1.05)},
                anchor=south,
                /tikz/every even column/.append style={column sep=0.5cm}
            },
            legend image code/.code={\draw [#1] (0cm,-0.075cm) rectangle (0.2cm,0.125cm);}
        ]
        
            \addplot+[
                fill=green, draw=green, bar width=2pt, error bars/.cd, y dir=plus, y explicit, error mark=|,
                error bar style={draw=green,line width=0.9pt}, error mark options={draw=green,line width=0.9pt}
            ] coordinates {(0,0.0000) +- (0,0.0000) (1,0.7473) +- (0,0.0972) (2,0.7763) +- (0,0.0412) (3,0.8115) +- (0,0.0435) (4,0.8315) +- (0,0.0440) (5,0.7592) +- (0,0.0704) (6,0.7783) +- (0,0.0384) (7,0.7958) +- (0,0.0550) (8,0.8366) +- (0,0.0354) (9,0.8596) +- (0,0.0561) (10,0.8792) +- (0,0.0322) (11,0.8941) +- (0,0.0332) (12,0.9320) +- (0,0.0306)};

            \addplot+[
                fill=darkgreen, draw=darkgreen, bar width=2pt, error bars/.cd, y dir=plus, y explicit, error mark=|,
                error bar style={draw=darkgreen,line width=0.9pt}, error mark options={draw=darkgreen,line width=0.9pt}
            ] coordinates {(0,0.0000) +- (0,0.0000) (1,0.1386) +- (0,0.0411) (2,0.3354) +- (0,0.0739) (3,0.4289) +- (0,0.0851) (4,0.4644) +- (0,0.0643) (5,0.1290) +- (0,0.0554) (6,0.3763) +- (0,0.0862) (7,0.4260) +- (0,0.0701) (8,0.4945) +- (0,0.0589) (9,0.4009) +- (0,0.0813) (10,0.4981) +- (0,0.0635) (11,0.5397) +- (0,0.0793) (12,0.5806) +- (0,0.0895)};

            \legend{$\simRulesets\paren{\ruleset,\ruleset^*}$,$\simRulesets\paren{\ruleset^*,\ruleset}$}
            \node at (axis cs:\pgfkeysvalueof{/pgfplots/xmin},\pgfkeysvalueof{/pgfplots/ymax}) [anchor=east, xshift=-16pt, yshift=10pt] {\textbf{(b)}};
        \end{axis}

    \end{tikzpicture}
    \caption{
        \textit{Extended results on synthetic data for varying signal-to-noise parameters $(\fracTarg, \fracCons)$.} 
        We show the mean results over 10 rule sets and corresponding datasets per synthetic setting. 
        Error bars mark the empirical standard deviation.
        \textbf{(a)}~Test accuracy compared to baseline models.
        \textbf{(b)}~Similarity between the rule set $\ruleset$ discovered by \ourMethod{} and the ground truth $\ruleset^*$. 
    }
    \label{fig:synthetic_noise_app}
\end{figure*}
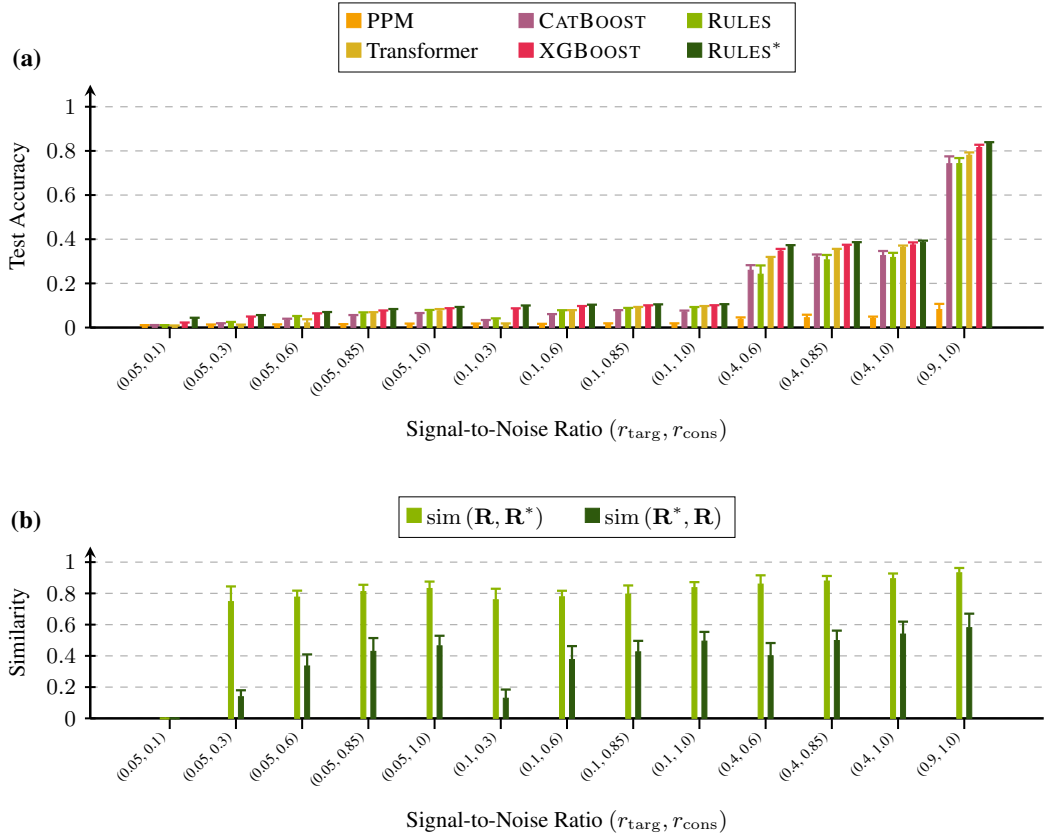

\paragraph{Settings.}
In \cref{fig:synthetic_num_samples,fig:synthetic_vocab_size}, we show the test accuracy and the rule set similarity to the ground-truth rule set $\ruleset^*$ for varying dataset sizes and vocabulary sizes, respectively.
In both cases, we fix the signal-to-noise parameters $\fracCons = 0.85$ and $\fracTarg = 0.4$.
The test datasets contain as many sequences ($\numSamples$) as the training datasets, with the default value being $\numSamples = 1024$.
The default value for the vocabulary size is $\vocabSize = 128$.
For completeness, \cref{fig:synthetic_noise_app} reports results over a wider range of signal-to-noise parameters than \cref{fig:synthetic_noise}.
As before, $\text{\ourMethod{}}^*$ denotes a \ourMethod{} model with embedded ground-truth antecedents.

\paragraph{Results.}
We find that an increasing number of training sequences improves both the ability of  \ourMethod to reconstruct the ground truth and its prediction accuracy.
Regarding the vocabulary size $\vocabSize$, \ourMethod works well for medium to large vocabularies ($\vocabSize\in [16,1024]$).
Low-dimensional data ($\vocabSize\in \set{4,8}$) makes it hard to find structure beyond the base frequencies.
In contrast, high-dimensional data significantly increases the space of possible rule sets, but at the same time permits more sparse rules with smaller antecedents for prediction.
This is reflected by increasing $\simRulesets(\ruleset,\ruleset^*)$ but decreasing $\simRulesets(\ruleset^*,\ruleset)$, with the test accuracy remaining in the same region.

\begin{figure*}[hb]
    \centering
    \begin{tikzpicture}[scale=0.83, transform shape]
    
        \begin{axis}[
            width=10.5cm,
            height=4cm,
            clip=false,
            axis lines=left,
            axis line style={line width=1pt},
            x axis line style={-},
            tick style={line width=0.8pt, black},
            tick label style={font=\small},
            label style={font=\small},
            ymajorgrids,
            grid style={black!30,line width=0.5pt,dashed},
            ybar,
            ymin=0,
            enlarge x limits=0.1,
            enlarge y limits={upper=0.1},
            yticklabel style={/pgf/number format/fixed},
            xlabel={Vocabulary Size},
            ylabel={Test Accuracy},
            xtick=data,
            xticklabels={},
            xticklabels={4,8,16,32,128,256,512,1024},
            xticklabel style={font=\small},
            ymax=0.42,
            ytick={0.0,0.1,0.2,0.3,0.4},
            scaled y ticks=false,
            legend columns=3,
            legend style={
                font=\scriptsize,
                legend cell align=left,
                at={(0.5,1.05)},
                anchor=south,
                /tikz/every even column/.append style={column sep=0.5cm}
            },
            legend image code/.code={\draw [#1] (0cm,-0.075cm) rectangle (0.2cm,0.125cm);}
        ]

            \addplot+[
                fill=orange, draw=orange, bar width=2pt, error bars/.cd, y dir=plus, y explicit, error mark=|,
                error bar style={draw=orange,line width=0.9pt}, error mark options={draw=orange,line width=0.9pt}
            ] coordinates {(0,0.2907) +- (0,0.0203) (1,0.1755) +- (0,0.0222) (2,0.1036) +- (0,0.0169) (3,0.0777) +- (0,0.0172) (4,0.0446) +- (0,0.0135) (5,0.0376) +- (0,0.0090) (6,0.0261) +- (0,0.0096) (7,0.0188) +- (0,0.0063)};

            \addplot+[
                fill=purple, draw=purple, bar width=2pt, error bars/.cd, y dir=plus, y explicit, error mark=|,
                error bar style={draw=purple,line width=0.9pt}, error mark options={draw=purple,line width=0.9pt}
            ] coordinates {(0,0.3123) +- (0,0.0100) (1,0.2309) +- (0,0.0124) (2,0.2322) +- (0,0.0215) (3,0.2601) +- (0,0.0193) (4,0.3193) +- (0,0.0116) (5,0.3182) +- (0,0.0194) (6,0.3229) +- (0,0.0154) (7,0.3149) +- (0,0.0162)};

            \addplot+[
                fill=green, draw=green, bar width=2pt, error bars/.cd, y dir=plus, y explicit, error mark=|,
                error bar style={draw=green,line width=0.9pt}, error mark options={draw=green,line width=0.9pt}
            ] coordinates {(0,0.3009) +- (0,0.0101) (1,0.2608) +- (0,0.0210) (2,0.3299) +- (0,0.0152) (3,0.3541) +- (0,0.0180) (4,0.3067) +- (0,0.0220) (5,0.2788) +- (0,0.0220) (6,0.2828) +- (0,0.0175) (7,0.3089) +- (0,0.0179)};

            \addplot+[
                fill=yellow, draw=yellow, bar width=2pt, error bars/.cd, y dir=plus, y explicit, error mark=|,
                error bar style={draw=yellow,line width=0.9pt}, error mark options={draw=yellow,line width=0.9pt}
            ] coordinates {(0,0.3062) +- (0,0.0122) (1,0.2261) +- (0,0.0124) (2,0.2501) +- (0,0.0132) (3,0.3091) +- (0,0.0120) (4,0.3507) +- (0,0.0056) (5,0.3502) +- (0,0.0085) (6,0.3582) +- (0,0.0065) (7,0.3589) +- (0,0.0043)};

            \addplot+[
                fill=red, draw=red, bar width=2pt, error bars/.cd, y dir=plus, y explicit, error mark=|,
                error bar style={draw=red,line width=0.9pt}, error mark options={draw=red,line width=0.9pt}
            ] coordinates {(0,0.3289) +- (0,0.0081) (1,0.2812) +- (0,0.0139) (2,0.3046) +- (0,0.0210) (3,0.3359) +- (0,0.0168) (4,0.3700) +- (0,0.0050) (5,0.3708) +- (0,0.0085) (6,0.3780) +- (0,0.0045) (7,0.3793) +- (0,0.0035)};

            \addplot+[
                fill=darkgreen, draw=darkgreen, bar width=2pt, error bars/.cd, y dir=plus, y explicit, error mark=|,
                error bar style={draw=darkgreen,line width=0.9pt}, error mark options={draw=darkgreen,line width=0.9pt}
            ] coordinates {(0,0.3586) +- (0,0.0120) (1,0.3609) +- (0,0.0277) (2,0.3887) +- (0,0.0157) (3,0.3891) +- (0,0.0090) (4,0.3854) +- (0,0.0013) (5,0.3802) +- (0,0.0052) (6,0.3804) +- (0,0.0039) (7,0.3793) +- (0,0.0018)};

            \legend{\textsc{PPM},\textsc{CatBoost},\ourMethod{},Transformer,\textsc{XGBoost},$\text{\ourMethod{}}^*$}
            \node at (axis cs:\pgfkeysvalueof{/pgfplots/xmin},\pgfkeysvalueof{/pgfplots/ymax}) [anchor=east, xshift=-16pt, yshift=10pt] {\textbf{(a)}};
        \end{axis}

        \begin{axis}[
            at={(10.45cm,0)},
            width=6.7cm,
            height=4cm,
            clip=false,
            axis lines=left,
            axis line style={line width=1pt},
            x axis line style={-},
            tick style={line width=0.8pt, black},
            tick label style={font=\small},
            label style={font=\small},
            ymajorgrids,
            grid style={black!30,line width=0.5pt,dashed},
            ybar,
            ymin=0,
            enlarge x limits=0.1,
            enlarge y limits={upper=0.1},
            yticklabel style={/pgf/number format/fixed},
            xlabel={Vocabulary Size},
            ylabel={Similarity},
            xtick=data,
            xticklabels={},
            xticklabels={4,8,16,32,128,256,512,1024},
            xticklabel style={font=\small},
            ymax=1,
            ytick={0.0,0.2,0.4,0.6,0.8,1.0},
            scaled y ticks=false,
            legend columns=2,
            legend style={
                font=\scriptsize,
                legend cell align=left,
                at={(0.5,1.05)},
                anchor=south,
                /tikz/every even column/.append style={column sep=0.5cm}
            },
            legend image code/.code={\draw [#1] (0cm,-0.075cm) rectangle (0.2cm,0.125cm);}
        ]
        
           \addplot+[
                fill=green, draw=green, bar width=2pt, error bars/.cd, y dir=plus, y explicit, error mark=|,
                error bar style={draw=green,line width=0.9pt}, error mark options={draw=green,line width=0.9pt}
            ] coordinates {(0,0.1794) +- (0,0.0572) (1,0.6329) +- (0,0.0656) (2,0.7918) +- (0,0.0514) (3,0.8516) +- (0,0.0330) (4,0.8792) +- (0,0.0322) (5,0.8974) +- (0,0.0423) (6,0.9508) +- (0,0.0264) (7,0.9287) +- (0,0.0540)};

            \addplot+[
                fill=darkgreen, draw=darkgreen, bar width=2pt, error bars/.cd, y dir=plus, y explicit, error mark=|,
                error bar style={draw=darkgreen,line width=0.9pt}, error mark options={draw=darkgreen,line width=0.9pt}
            ] coordinates {(0,0.2694) +- (0,0.0665) (1,0.2102) +- (0,0.0515) (2,0.5268) +- (0,0.1058) (3,0.7040) +- (0,0.0829) (4,0.4981) +- (0,0.0635) (5,0.4410) +- (0,0.0881) (6,0.3822) +- (0,0.0761) (7,0.4138) +- (0,0.0626)};

            \legend{$\simRulesets\paren{\ruleset,\ruleset^*}$,$\simRulesets\paren{\ruleset^*,\ruleset}$}
            \node at (axis cs:\pgfkeysvalueof{/pgfplots/xmin},\pgfkeysvalueof{/pgfplots/ymax}) [anchor=east, xshift=-16pt, yshift=10pt] {\textbf{(b)}};
        \end{axis}

    \end{tikzpicture}
    \caption{
        \textit{\ourMethod{} performs well on synthetic data for varying vocabulary sizes $\vocabSize$.} 
        We show the mean results over 10 rule sets and corresponding datasets per synthetic setting. 
        Error bars mark the empirical standard deviation.
        \textbf{(a)}~Test accuracy compared to baseline models.
        \textbf{(b)}~Similarity between the rule set $\ruleset$ discovered by \ourMethod{} and the ground truth $\ruleset^*$.
    }
    \label{fig:synthetic_vocab_size}
\end{figure*}
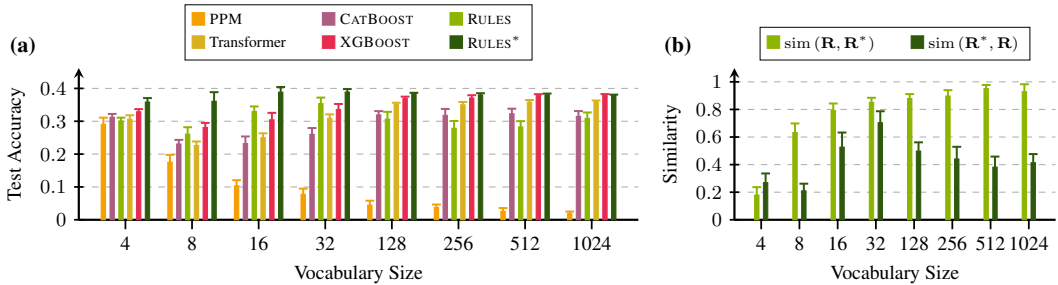

\clearpage
\section{Additional Results on Real Data}
\label{app:sec_results_real}

\subsection{Test Loss and Model Sizes on Real Datasets}
\label{app:subsec_cross_entropy_and_model_sizes}
In \cref{tab:table_real_ce_loss}, we report test cross-entropy loss for the models in \cref{tab:table_real_results}, with trends mirroring those results.
In \cref{tab:table_real_model_sizes}, we further report the sizes of all models that can be fully decomposed into symbolic rules, including their average antecedent lengths.
\begin{table}[hb]
    \centering
    \caption{
        \textit{Test cross-entropy on real-world datasets} (lower is better).
        The best results are in \textbf{bold}, the second best are \underline{underlined}.
    }
    \label{tab:table_real_ce_loss}
    \setlength{\tabcolsep}{8pt}
    \begin{tabular}{l rrr r rrr rrr rrr r}
        \toprule
        \textbf{Datasets}    & \multicolumn{1}{c}{\ourMethod{}} & \multicolumn{1}{c}{\ourMethodSideChannel{}} & \multicolumn{1}{c}{\textsc{PPM}} & \multicolumn{1}{c}{\textsc{CatBoost}} & \multicolumn{1}{c}{\textsc{XGBoost}} & Transformer \\
        \midrule
        Holidays             &              0.053  &      \textbf{0.000} &              0.022  &      \textbf{0.000} &              0.001  &              0.004   \\
        Chorales             &              2.404  &              1.897  &              1.993  &              1.967  &   \underline{1.880} &      \textbf{1.820}  \\
        Commits              &      \textbf{4.161} &              4.212  &              5.459  &   \underline{4.200} &              4.547  &              4.330   \\
        Rolling Mill         &              2.581  &   \underline{1.174} &              1.321  &              1.399  &              1.247  &      \textbf{1.072}  \\
        Amazon               &              5.799  &   \underline{4.829} &              5.318  &              5.429  &              5.270  &      \textbf{4.817}  \\
        \midrule
        Avg. Rank            &                5.0  &       \textbf{2.2}  &                4.8  &                3.4  &                3.2  &         \textbf{2.2}   \\
        \bottomrule
    \end{tabular}
\end{table}

\begin{table}[hb]
    \centering
    \caption{
        \textit{Number of rules ($\abs{\ruleset}$), average antecedent size ($\abs{\ant}$), and test accuracy (Acc.) in \% on real-world datasets for models that can be fully decomposed into symbolic rules.}
        \ourMethod{} consistently produces compact rule sets with small antecedents.
    }
    \label{tab:table_real_model_sizes}
    \resizebox{1.0\linewidth}{!}{{\small
    \begin{tabular}{l rrr rrr rrr rrr}
        \toprule
                            & \multicolumn{3}{c}{\ourMethod{}} & \multicolumn{3}{c}{\textsc{PPM}} & \multicolumn{3}{c}{\textsc{CatBoost}} & \multicolumn{3}{c}{\textsc{XGBoost}} \\
        \cmidrule(lr){2-4} \cmidrule(lr){5-7} \cmidrule(lr){8-10} \cmidrule(lr){11-13}
        \textbf{Datasets}    & $\abs{\ruleset}$   & $\abs{\ant}$       & Acc. $\uparrow$     & $\abs{\ruleset}$   & $\abs{\ant}$       & Acc. $\uparrow$     & $\abs{\ruleset}$   & $\abs{\ant}$       & Acc. $\uparrow$     & $\abs{\ruleset}$   & $\abs{\ant}$       & Acc. $\uparrow$     \\
        \midrule
        Holidays             &                26  &               1.2  &              97.57  &                14  &               1.0  &             100.00  &              1129  &          
     4.0  &             100.00  &              1455  &               3.2  &             100.00   \\
        Chorales             &                33  &               4.2  &              21.00  &               180  &               1.9  &              31.40  &             11596  &          
     6.0  &              37.10  &             34462  &               3.7  &              37.72   \\
        Commits              &                51  &               2.6  &              19.60  &                80  &               1.0  &              18.11  &               380  &          
     6.0  &              15.55  &             83965  &               3.8  &              17.03   \\
        Rolling Mill         &                77  &               4.1  &              34.13  &              9583  &               2.6  &              67.64  &             13809  &          
     6.0  &              65.64  &            640723  &               3.4  &              68.84   \\
        Amazon               &                42  &               1.7  &               6.53  &               993  &               1.0  &              15.16  &             13123  &          
     6.0  &              11.79  &           5149683  &               4.9  &              15.37   \\
        \bottomrule
    \end{tabular}}}
\end{table}

\clearpage

\subsection{Influence of the Residual Network for Individual Datasets}
\label{app:subsec_confusion_side_channel_individual}

In \cref{tab:confusion_side_channel_individual}, we show the individual results for each dataset that when averaged produce  \cref{tab:confusion_side_channel_aggregated}.
For averaging, we sum up the different table entries for each dataset and then divide by the number of datasets.
The Commits dataset is excluded, as \ourMethod already performs well, causing early stopping to select a \ourMethodSideChannel model in which the residual network remains effectively untrained.
\begin{table}[hb]
\centering
{\footnotesize
\caption{
        \textit{Influence of the Residual Network for Individual Datasets.}
        Counts are normalized within the rule activity groups. 
        The bottom rows show the fraction of predictions that fall within the respective groups and the total number of testing \seqElementText{}s.
    }
\label{tab:confusion_side_channel_individual}
\begin{subtable}{0.48\linewidth}
    \centering
    \caption{
        Holidays
    }
    \begin{tabular}{l rr rr}
        \toprule
                         & \multicolumn{4}{c}{\ourMethod{}} \\
        \cmidrule(lr){2-5}
        \ourMethodSideChannel{}     & \multicolumn{2}{c}{Rule Active} & \multicolumn{2}{c}{No Rule Active}\\
        \cmidrule(lr){2-3} \cmidrule(lr){4-5}
                         & right & wrong & right & wrong  \\
        \midrule
        right            &  0.99 &  0.01 &  0.50 &  0.50  \\
        wrong            &  0.00 &  0.00 &  0.00 &  0.00  \\
        \midrule
        Fraction of Data & \multicolumn{2}{c}{0.97} & \multicolumn{2}{c}{0.03} \\ \cmidrule(lr){2-5}
        Total Count      & \multicolumn{4}{c}{11264} \\
        \bottomrule
    \end{tabular}
\end{subtable}
\hfill
\begin{subtable}{0.48\linewidth}
    \centering
    \caption{
        Chorales
    }
   \begin{tabular}{l rr rr}
        \toprule
                         & \multicolumn{4}{c}{\ourMethod{}} \\
        \cmidrule(lr){2-5}
        \ourMethodSideChannel{} & \multicolumn{2}{c}{Rule Active} & \multicolumn{2}{c}{No Rule Active}\\
        \cmidrule(lr){2-3} \cmidrule(lr){4-5}
                         & right & wrong & right & wrong  \\
        \midrule
        right            &  0.16 &  0.15 &  0.10 &  0.35  \\
        wrong            &  0.07 &  0.63 &  0.10 &  0.46  \\
        \midrule
        Fraction of Data & \multicolumn{2}{c}{0.55} & \multicolumn{2}{c}{0.45} \\ \cmidrule(lr){2-5}
        Total Count      & \multicolumn{4}{c}{981} \\
        \bottomrule
    \end{tabular}
\end{subtable}

\vspace{0.5cm}

\begin{subtable}{0.48\linewidth}
    \centering
    \caption{
        Rolling Mill
    }
    \begin{tabular}{l rr rr}
        \toprule
                         & \multicolumn{4}{c}{\ourMethod{}} \\
        \cmidrule(lr){2-5}
        \ourMethodSideChannel{}     & \multicolumn{2}{c}{Rule Active} & \multicolumn{2}{c}{No Rule Active}\\
        \cmidrule(lr){2-3} \cmidrule(lr){4-5}
                         & right & wrong & right & wrong  \\
        \midrule
        right            &  0.34 &  0.37 &  0.14 &  0.48  \\
        wrong            &  0.03 &  0.26 &  0.01 &  0.38  \\
        \midrule
        Fraction of Data & \multicolumn{2}{c}{0.88} & \multicolumn{2}{c}{0.12} \\ \cmidrule(lr){2-5}
        Total Count      & \multicolumn{4}{c}{10507} \\
        \bottomrule
    \end{tabular}
\end{subtable}
\hfill
\begin{subtable}{0.48\linewidth}
    \centering
    \caption{
        Amazon
    }
    \begin{tabular}{l rr rr}
        \toprule
                         & \multicolumn{4}{c}{\ourMethod{}} \\
        \cmidrule(lr){2-5}
        \ourMethodSideChannel{}     & \multicolumn{2}{c}{Rule Active} & \multicolumn{2}{c}{No Rule Active}\\
        \cmidrule(lr){2-3} \cmidrule(lr){4-5}
                         & right & wrong & right & wrong  \\
        \midrule
        right            &  0.10 &  0.16 &  0.05 &  0.11  \\
        wrong            &  0.01 &  0.72 &  0.01 &  0.83  \\
        \midrule
        Fraction of Data & \multicolumn{2}{c}{0.17} & \multicolumn{2}{c}{0.83} \\ \cmidrule(lr){2-5}
        Total Count      & \multicolumn{4}{c}{121498} \\
        \bottomrule
    \end{tabular}
\end{subtable}
}
\end{table}

\subsection{Influence of the Residual Network for Individual Predictions}
\label{app:subsec_influence_side_channel_individual_predictions}

\ourMethodSideChannel{} trains the residual network on the logit residuals of \ourMethod{} and combines both of their logits additively.
This not only yields a transparent decomposition into rule-based and residual contributions, but also provides informative signals even when the ground-truth label is unavailable.

\begin{figure*}[hb]
    \centering
    
    \begin{subfigure}{1.0\linewidth}
        \centering
        \includegraphics[width=1.0\linewidth]{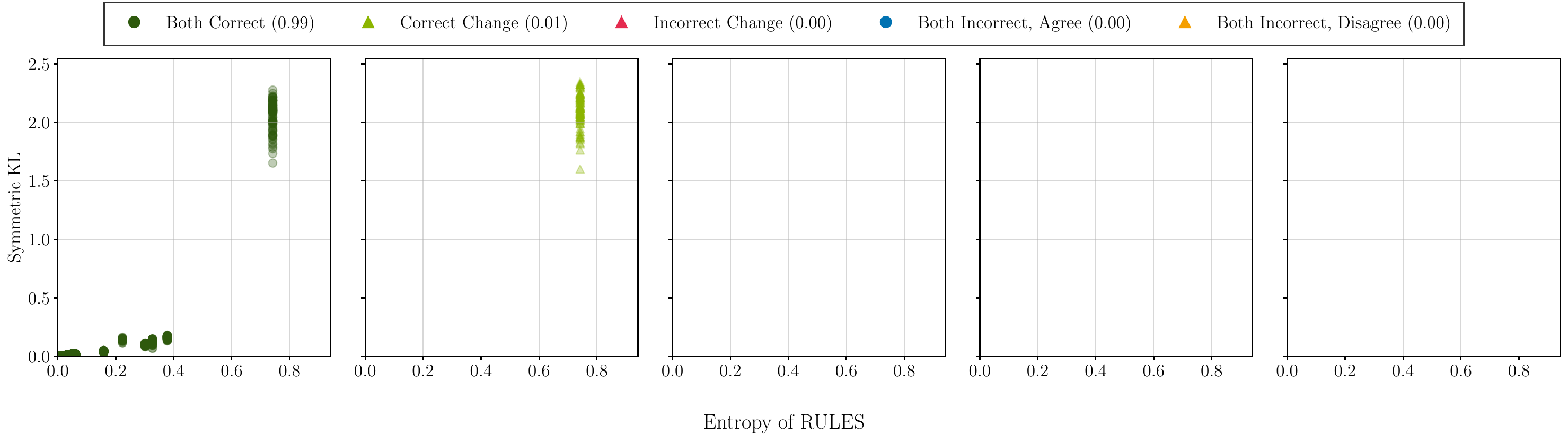}
        \caption{Holidays, 10890 total \seqElementText{}s with active rules.}
        \label{subfig:side_channel_individual_holidays}
    \end{subfigure}
    \vspace{0.02cm}

    \begin{subfigure}{1.0\linewidth}
        \centering
        \includegraphics[width=1.0\linewidth]{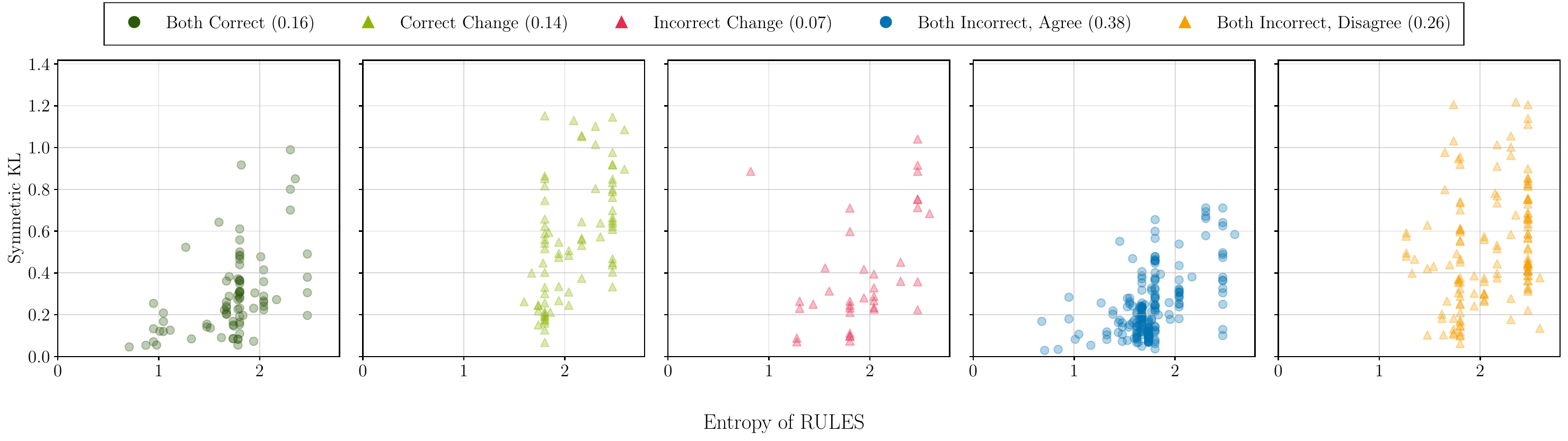}
        \caption{Chorales, 540 total \seqElementText{}s with active rules.}
        \label{subfig:side_channel_individual_chorales}
    \end{subfigure}
    \vspace{0.02cm}

    \begin{subfigure}{1.0\linewidth}
        \centering
        \includegraphics[width=1.0\linewidth]{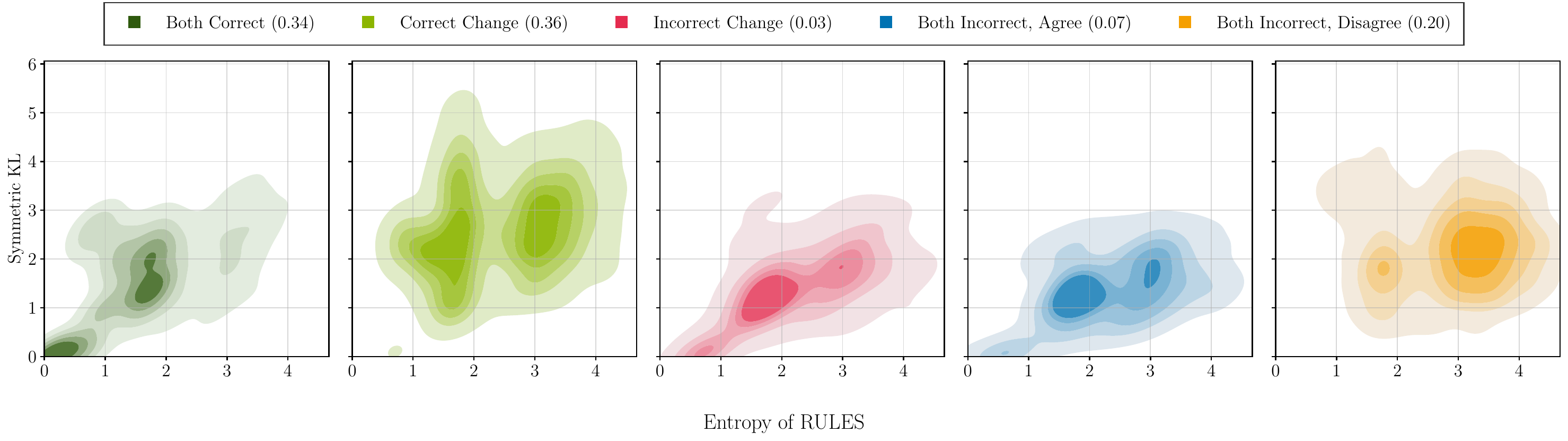}
        \caption{Rolling Mill, 9198 total \seqElementText{}s with active rules.}
        \label{subfig:side_channel_individual_rolling_mill}
    \end{subfigure}
    \vspace{0.02cm}

    \begin{subfigure}{1.0\linewidth}
        \centering
        \includegraphics[width=1.0\linewidth]{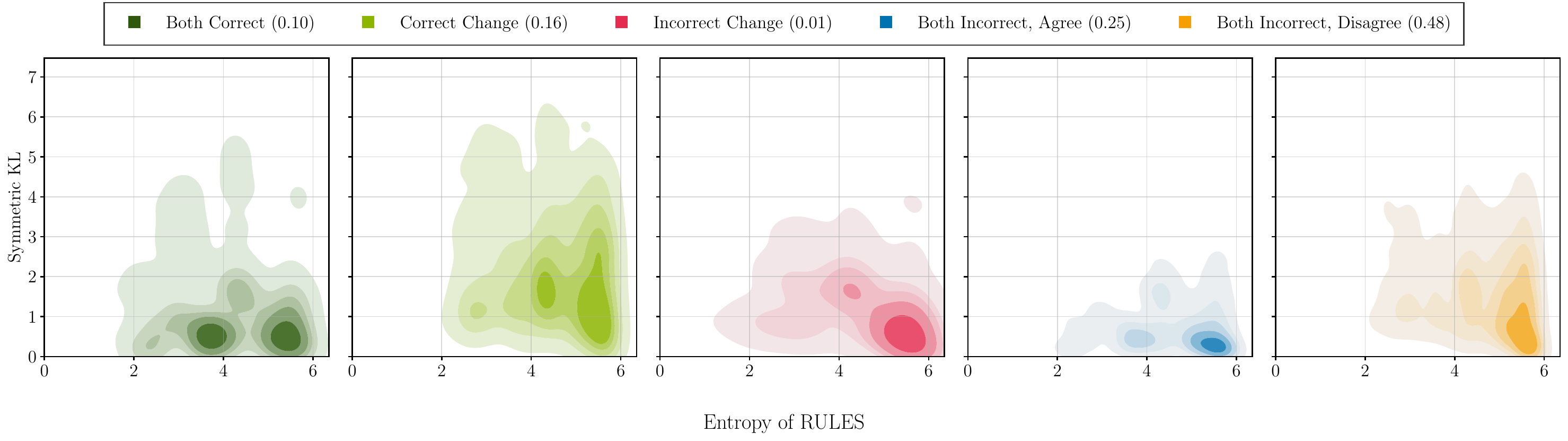}
        \caption{Amazon, 20386 total \seqElementText{}s with active rules.}
        \label{subfig:side_channel_individual_amazon}
    \end{subfigure}
    \caption{
        \textit{Entropy of the output distribution of \ourMethod{} (x-axis) versus the symmetric KL divergence to \ourMethodSideChannel{} (y-axis) on test splits of real-world datasets, considering \seqElementText{}s with at least one active rule.}
        Plots are grouped by prediction outcome: both models correct (\checkmark\checkmark), only \ourMethodSideChannel{} correct (correct change, \ding{55}$\to$\checkmark), only \ourMethod{} correct (incorrect change, \checkmark$\to$\ding{55}), both incorrect (\ding{55}\ding{55}) with the same prediction, and both incorrect with different predictions (\ding{55}$\to$\ding{55}).
        For clarity, density plots are shown for Rolling Mill and Amazon.
        Subcaptions report the total number of \seqElementText{}s per dataset, and the legends indicate the fraction corresponding to each case.
    }
    \label{fig:kl_entropy_side_channel}
\end{figure*}

To illustrate this, we plot the entropy of the output distribution of \ourMethod{} against the symmetric KL divergence between \ourMethod{} and \ourMethodSideChannel{} for individual predictions in \cref{fig:kl_entropy_side_channel}, additionally distinguishing correctness and agreement between the models.
We observe that (i) the residual network usually induces larger shifts (high KL) when \ourMethod{} is uncertain (high entropy), and (ii) large entropy and KL values combined with disagreement often correspond either to correct adjustments or to noisy \seqElementText{}s where neither model predicts correctly.

For example, on Rolling Mill, a logit shift corresponding to a symmetric KL divergence greater than $3$ strongly correlates with correct prediction changes when the models disagree, while incorrect prediction changes almost never occur in this region, providing a practical signal for deciding whether to rely only on the rule-based or also the residual component.
Similarly, on Chorales, when the entropy of \ourMethod{} is below $1.5$, correct changes do not occur while incorrect ones are frequent, indicating that rules should be preferred in this regime.

\clearpage

\subsection{Effect of Symbolic Rule Processing}
\label{app:subsec_effect_rule_processing}
\cref{tab:statistics_rule_processing} summarizes the effect of symbolic rule post-processing (Appendices~\ref{app:subsec_rule_set_extraction}--\ref{app:subsec_symbolic_rule_refinement}) on the antecedent complexity and test accuracy of \ourMethod{} for real-world data.
Post-processing accuracy refers to the accuracy of \ourMethod{} after embedding the processed antecedents into its hidden layers and retraining only the output layer (i.e., the \ourMethod{} model that we use in our experiments).

\paragraph{Results.}
Post-processing substantially reduces both the number of predicates per antecedent and the width of the remaining delay intervals.
Its effect on the test accuracy is mixed: accuracy improves on some datasets, likely due to more precise rule activations, while it decreases on others.
\cref{fig:histograms_rule_activitiy} shows that post-processing also reduces the number of active rules per predicted \seqElementText{}, increasing sparsity.
We hypothesize that, prior to post-processing, antecedent neurons can be entangled for downstream predictions rather than representing individual, meaningful rules. 
The increased sparsity induced by symbolic refinement likely promotes disentanglement into more self-contained rules, improving interpretability while sometimes reducing predictive performance.

\begin{table}[hb]
    \centering
    \caption{
        \textit{Symbolic processing of rules extracted from \ourMethod{} reduces antecedent complexity.}
        The table shows \ourMethod{} before and after symbolic post-processing of rules (Appendices~\ref{app:subsec_rule_set_extraction}--\ref{app:subsec_symbolic_rule_refinement}).
        It reports test accuracy for next-\seqElementText{} prediction (Acc.) in \%, average antecedent size ($\abs{\ant}$), and average delay interval width ($\abs{\Delta_i} = \abs{[\minOffset, \maxOffset]}$) across rules.
        Standard deviations for antecedent sizes and delay intervals are included.
        The final column shows the absolute change after post-processing.
        Post-processing accuracy refers to \ourMethod{} after embedding the processed antecedents into its hidden layers and retraining only the output layer.
    }
    \resizebox{1.0\linewidth}{!}{{\footnotesize
    \label{tab:statistics_rule_processing}
    \begin{tabular}{l rrr rrr rrr}
        \toprule
                             & \multicolumn{3}{c}{Before} & \multicolumn{3}{c}{After} & \multicolumn{3}{c}{Difference} \\
        \cmidrule(lr){2-4} \cmidrule(lr){5-7} \cmidrule(lr){8-10}
        \textbf{Datasets}    & Acc.         & Avg. $\abs{\ant}$     & Avg. $\abs{\Delta_i}$            & Acc.         & Avg. $\abs{\ant}$     & Avg. $\abs{\Delta_i}$            & Acc.                   & Avg. $\abs{\ant}$     & Avg. $\abs{\Delta_i}$  \\
        \midrule
        Holidays             &        96.12 &       3.1 $\pm$   1.8 &     107.6 $\pm$             51.9 &        97.57 &       1.2 $\pm$   0.4 &       8.5 $\pm$             14.0 &  {\scriptsize $+$}1.45 & {\scriptsize $-$}1.99 & {\scriptsize $-$}99.03  \\
        Chorales             &        22.22 &       6.5 $\pm$   3.6 &      16.7 $\pm$   \phantom{0}7.5 &        21.00 &       4.2 $\pm$   2.3 &       5.1 $\pm$   \phantom{0}6.3 &  {\scriptsize $-$}1.22 & {\scriptsize $-$}2.28 & {\scriptsize $-$}11.59  \\
        Commits              &        11.72 &       3.2 $\pm$   2.2 &       6.9 $\pm$   \phantom{0}4.9 &        19.60 &       2.6 $\pm$   1.8 &       2.8 $\pm$   \phantom{0}3.4 &  {\scriptsize $+$}7.88 & {\scriptsize $-$}0.54 & {\scriptsize $-$}\phantom{0}4.13  \\
        Rolling Mill         &        43.10 &       7.0 $\pm$   4.0 &      25.2 $\pm$   \phantom{0}9.7 &        34.13 &       4.1 $\pm$   2.2 &       4.1 $\pm$   \phantom{0}4.9 &  {\scriptsize $-$}8.97 & {\scriptsize $-$}2.95 & {\scriptsize $-$}21.04  \\
        Amazon               &         7.55 &       1.7 $\pm$   0.9 &       6.2 $\pm$   \phantom{0}6.7 &         6.53 &       1.7 $\pm$   0.9 &       3.3 $\pm$   \phantom{0}5.3 &  {\scriptsize $-$}1.02 & {\scriptsize $-$}0.00 & {\scriptsize $-$}\phantom{0}2.88  \\
        \bottomrule
    \end{tabular}
    }}
\end{table}
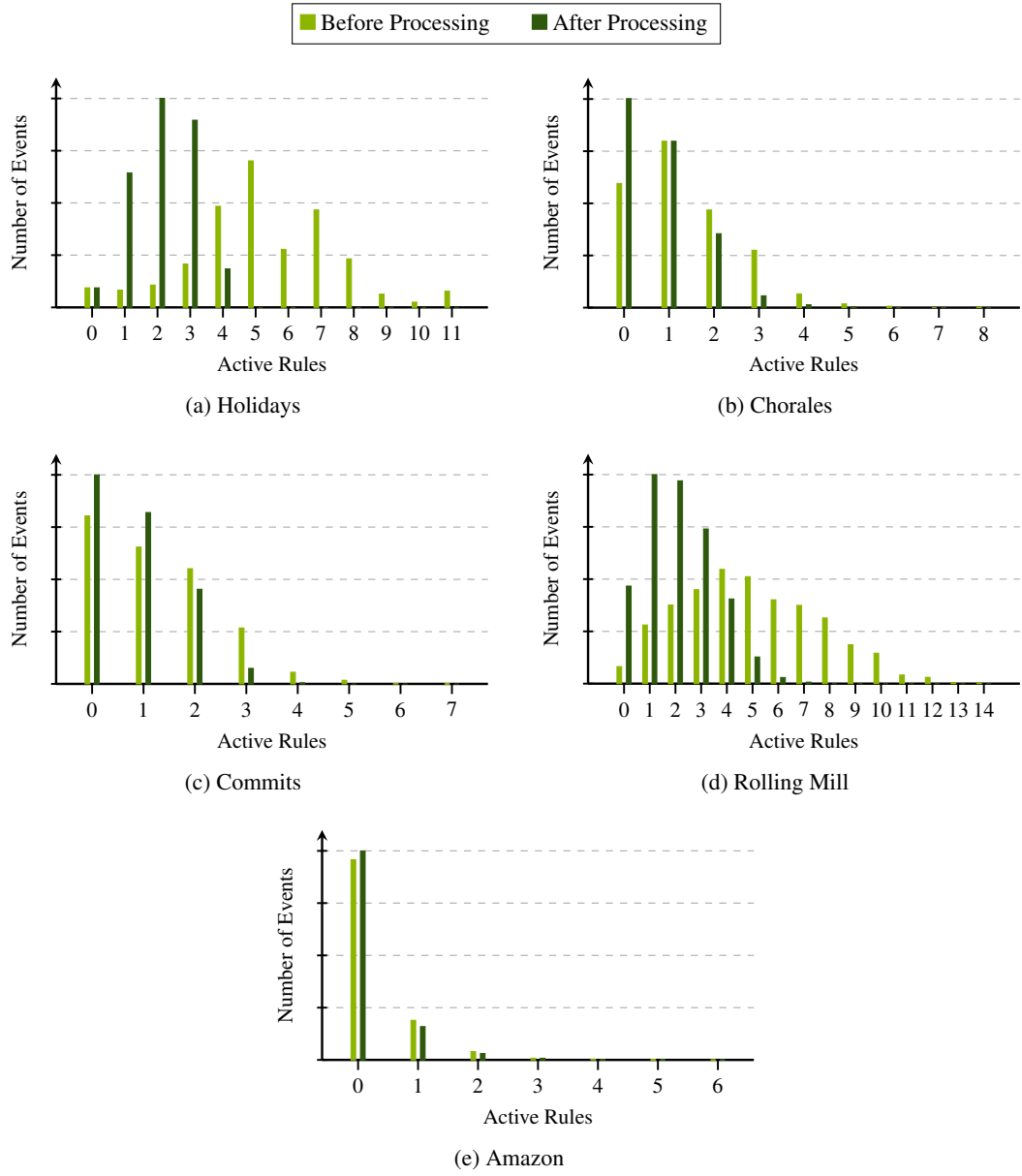
\begin{figure}[hb]
\centering

\ref{sharedlegend_2}
\vspace{0.4cm}
    
\begin{subfigure}{0.48\linewidth}
    \resizebox{1.0\linewidth}{!}{
    \begin{tikzpicture}
    
        \begin{axis}[
            width=8cm,
            height=5cm,
            clip=false,
            axis lines=left,
            axis line style={line width=1pt},
            x axis line style={-},
            tick style={line width=0.8pt, black},
            tick label style={font=\small},
            label style={font=\small},
            ymajorgrids,
            grid style={black!30,line width=0.5pt,dashed},
            ybar,
            ymin=0,
            enlarge x limits=0.1,
            enlarge y limits={upper=0.1},
            yticklabel style={/pgf/number format/fixed},
            xlabel={Active Rules},
            ylabel={Number of Events},
            xtick=data,
            xticklabels={0,1,2,3,4,5,6,7,8,9,10,11},
            ymax=3999,
            ytick={0,999,1998,2997,3996},
            yticklabels={},
            scaled y ticks=false,
            legend to name=sharedlegend_2,
            legend columns=2,
            legend style={
                font=\small,
                text=black,
                legend cell align=left,
                at={(0.5,1.05)},
                anchor=south,
                /tikz/every even column/.append style={column sep=0.5cm}
            },
            legend image code/.code={\draw [#1] (0cm,-0.075cm) rectangle (0.2cm,0.125cm);}
        ]
        
            \addplot+[
                fill=green, draw=green, bar width=2pt
            ] coordinates {(0,374.0000) (1,331.0000) (2,427.0000) (3,830.0000) (4,1934.0000) (5,2803.0000) (6,1110.0000) (7,1864.0000) (8,928.0000) (9,253.0000) (10,100.0000) (11,310.0000)};

            \addplot+[
                fill=darkgreen, draw=darkgreen, bar width=2pt
            ] coordinates {(0,374.0000) (1,2573.0000) (2,3999.0000) (3,3581.0000) (4,737.0000) (5,0.0000) (6,0.0000) (7,0.0000) (8,0.0000) (9,0.0000) (10,0.0000) (11,0.0000)};

        \legend{Before Processing,After Processing}
        \end{axis}

    \end{tikzpicture}}
    \caption{
        Holidays
    }
\end{subfigure}
\hfill
\begin{subfigure}{0.48\linewidth}
    \resizebox{1.0\linewidth}{!}{
    \begin{tikzpicture}
    
        \begin{axis}[
            width=8cm,
            height=5cm,
            clip=false,
            axis lines=left,
            axis line style={line width=1pt},
            x axis line style={-},
            tick style={line width=0.8pt, black},
            tick label style={font=\small},
            label style={font=\small},
            ymajorgrids,
            grid style={black!30,line width=0.5pt,dashed},
            ybar,
            ymin=0,
            enlarge x limits=0.1,
            enlarge y limits={upper=0.1},
            yticklabel style={/pgf/number format/fixed},
            xlabel={Active Rules},
            ylabel={Number of Events},
            xtick=data,
            xticklabels={0,1,2,3,4,5,6,7,8},
            yticklabels={},
            ymax=441,
            ytick={0,110,220,330,440},
            scaled y ticks=false,
        ]
        
            \addplot+[
                fill=green, draw=green, bar width=2pt
            ] coordinates {(0,262.0000) (1,351.0000) (2,206.0000) (3,121.0000) (4,29.0000) (5,8.0000) (6,3.0000) (7,0.0000) (8,1.0000)};

            \addplot+[
                fill=darkgreen, draw=darkgreen, bar width=2pt
            ] coordinates {(0,441.0000) (1,351.0000) (2,156.0000) (3,25.0000) (4,6.0000) (5,2.0000) (6,0.0000) (7,0.0000) (8,0.0000)};

        \end{axis}

    \end{tikzpicture}}
    \caption{
        Chorales
    }
\end{subfigure}

\vspace{0.4cm}

\begin{subfigure}{0.48\linewidth}
    \resizebox{1.0\linewidth}{!}{
    \begin{tikzpicture}
    
        \begin{axis}[
            width=8cm,
            height=5cm,
            clip=false,
            axis lines=left,
            axis line style={line width=1pt},
            x axis line style={-},
            tick style={line width=0.8pt, black},
            tick label style={font=\small},
            label style={font=\small},
            ymajorgrids,
            grid style={black!30,line width=0.5pt,dashed},
            ybar,
            ymin=0,
            enlarge x limits=0.1,
            enlarge y limits={upper=0.1},
            yticklabel style={/pgf/number format/fixed},
            xlabel={Active Rules},
            ylabel={Number of Events},
            xtick=data,
            xticklabels={0,1,2,3,4,5,6,7},
            ymax=27420,
            ytick={0,6855,13710,20565,27420},
            yticklabels={},
            scaled y ticks=false,
        ]
        
            \addplot+[
                fill=green, draw=green, bar width=2pt
            ] coordinates {(0,22046.0000) (1,17947.0000) (2,15114.0000) (3,7339.0000) (4,1534.0000) (5,483.0000) (6,42.0000) (7,7.0000)};

            \addplot+[
                fill=darkgreen, draw=darkgreen, bar width=2pt
            ] coordinates {(0,27420.0000) (1,22482.0000) (2,12421.0000) (3,2040.0000) (4,149.0000) (5,0.0000) (6,0.0000) (7,0.0000)};

        \end{axis}

    \end{tikzpicture}}
    \caption{
        Commits
    }
\end{subfigure}
\hfill
\begin{subfigure}{0.48\linewidth}
    \resizebox{1.0\linewidth}{!}{
    \begin{tikzpicture}
    
        \begin{axis}[
            width=8cm,
            height=5cm,
            clip=false,
            axis lines=left,
            axis line style={line width=1pt},
            x axis line style={-},
            tick style={line width=0.8pt, black},
            tick label style={font=\small},
            label style={font=\small},
            ymajorgrids,
            grid style={black!30,line width=0.5pt,dashed},
            ybar,
            ymin=0,
            enlarge x limits=0.1,
            enlarge y limits={upper=0.1},
            yticklabel style={/pgf/number format/fixed},
            xlabel={Active Rules},
            ylabel={Number of Events},
            xtick=data,
            xticklabels={0,1,2,3,4,5,6,7,8,9,10,11,12,13,14},
            ymax=2804,
            ytick={0,701,1402,2103,2804},
            yticklabels={},
            scaled y ticks=false,
        ]
        
            \addplot+[
                fill=green, draw=green, bar width=2pt
            ] coordinates {(0,229.0000) (1,787.0000) (2,1054.0000) (3,1260.0000) (4,1533.0000) (5,1435.0000) (6,1125.0000) (7,1050.0000) (8,881.0000) (9,522.0000) (10,408.0000) (11,116.0000) (12,85.0000) (13,13.0000) (14,9.0000)};

            \addplot+[
                fill=darkgreen, draw=darkgreen, bar width=2pt
            ] coordinates {(0,1309.0000) (1,2804.0000) (2,2721.0000) (3,2075.0000) (4,1136.0000) (5,358.0000) (6,83.0000) (7,21.0000) (8,0.0000) (9,0.0000) (10,0.0000) (11,0.0000) (12,0.0000) (13,0.0000) (14,0.0000)};

        \end{axis}

    \end{tikzpicture}}
    \caption{
        Rolling Mill
    }
\end{subfigure}

\vspace{0.4cm}

\begin{subfigure}{0.48\linewidth}
    \resizebox{1.0\linewidth}{!}{
    \begin{tikzpicture}
    
        \begin{axis}[
            width=8cm,
            height=5cm,
            clip=false,
            axis lines=left,
            axis line style={line width=1pt},
            x axis line style={-},
            tick style={line width=0.8pt, black},
            tick label style={font=\small},
            label style={font=\small},
            ymajorgrids,
            grid style={black!30,line width=0.5pt,dashed},
            ybar,
            ymin=0,
            enlarge x limits=0.1,
            enlarge y limits={upper=0.1},
            yticklabel style={/pgf/number format/fixed},
            xlabel={Active Rules},
            ylabel={Number of Events},
            xtick=data,
            xticklabels={0,1,2,3,4,5,6},
            ymax=101112,
            ytick={0,25278,50556,75834,101112},
            yticklabels={},
            scaled y ticks=false,
        ]
        
            \addplot+[
                fill=green, draw=green, bar width=2pt
            ] coordinates {(0,96833.0000) (1,19175.0000) (2,4075.0000) (3,840.0000) (4,380.0000) (5,190.0000) (6,5.0000)};

            \addplot+[
                fill=darkgreen, draw=darkgreen, bar width=2pt
            ] coordinates {(0,101112.0000) (1,16115.0000) (2,3106.0000) (3,827.0000) (4,238.0000) (5,99.0000) (6,1.0000)};

        \end{axis}

    \end{tikzpicture}}
    \caption{
        Amazon
    }
\end{subfigure}

    \caption{
        \textit{Symbolic processing of rules extracted from \ourMethod{} reduces the number of active rules per prediction.}
        The figure shows histograms of the number of active rules per next-\seqElementText{} prediction on the test split of real-world datasets.
    }
    \label{fig:histograms_rule_activitiy}
\end{figure}

\clearpage

\subsection{Variability of \ourMethod, \ourMethodSideChannel, and the Transformer}
\label{app:subsec_variability}

For \ourMethod{} and \ourMethodSideChannel{} (and the transformer), we report results in \cref{tab:table_real_results,tab:table_real_ce_loss} using the best model (selected by validation loss; Appendix~\ref{app:subsec_hyperparams_real}) across four random seeds. 
This ensures consistent reporting across the three-stage training pipeline of \ourMethodSideChannel: (i)~end-to-end training of \ourMethod{}, (ii)~training of \ourMethod{} with symbolically refined rules, and (iii)~training of the residual network, where each stage selects the best model from the previous one.
In \cref{tab:table_mean_std_ce,tab:table_mean_std_acc}, we additionally report the mean and standard deviation on real-world datasets for each stage and the standalone transformer, using the same four seeds as for model selection.
Overall, variability is moderate during rule learning and small in later stages.

\begin{table}[hb]
    \centering
    \caption{
        \textit{Mean and standard deviation of the test cross-entropy (lower is better) across training stages of \ourMethodSideChannel{} on real-world datasets.}
        ``No refinement'' refers to \ourMethod{} without symbolically processed rules.
    }
    \label{tab:table_mean_std_ce}
    \small
    \begin{tabular}{l r r r r}
        \toprule
                            & \multicolumn{1}{c}{\ourMethod{} (no refinement)} & \multicolumn{1}{c}{\ourMethod{}} & \multicolumn{1}{c}{\ourMethodSideChannel{}} & \multicolumn{1}{c}{Transformer} \\
        \midrule
        Holidays             &  0.147 $\pm$  0.082 &  0.054 $\pm$  0.001 &  0.000 $\pm$  0.000 &  0.005 $\pm$  0.000 \\
        Chorales             &  2.627 $\pm$  0.065 &  2.404 $\pm$  0.002 &  1.931 $\pm$  0.029 &  1.841 $\pm$  0.024 \\
        Commits              &  8.107 $\pm$  0.671 &  4.186 $\pm$  0.018 &  4.191 $\pm$  0.017 &  4.250 $\pm$  0.073 \\
        Rolling Mill         &  2.468 $\pm$  0.104 &  2.580 $\pm$  0.001 &  1.183 $\pm$  0.010 &  1.069 $\pm$  0.003 \\
        Amazon               &  5.811 $\pm$  0.045 &  5.801 $\pm$  0.002 &  4.850 $\pm$  0.016 &  4.834 $\pm$  0.015 \\
        \bottomrule
    \end{tabular}
\end{table}

\begin{table}[hb]
    \centering
    \caption{
        \textit{Mean and standard deviation of the test accuracy in \% (higher is better) across training stages of \ourMethodSideChannel{} on real-world datasets.}
        ``No refinement'' refers to \ourMethod{} without symbolically processed rules.
    }
    \label{tab:table_mean_std_acc}
    \small
    \begin{tabular}{l r r r r}
        \toprule
                            & \multicolumn{1}{c}{\ourMethod{} (no refinement)} & \multicolumn{1}{c}{\ourMethod{}} & \multicolumn{1}{c}{\ourMethodSideChannel{}} & \multicolumn{1}{c}{Transformer} \\
        \midrule
        Holidays             &  90.98 $\pm$   4.05 &  97.57 $\pm$   0.00 & 100.00 $\pm$   0.00 & 100.00 $\pm$   0.00 \\
        Chorales             &  21.41 $\pm$   0.59 &  21.15 $\pm$   0.10 &  37.16 $\pm$   0.72 &  39.17 $\pm$   0.49 \\
        Commits              &  12.85 $\pm$   1.27 &  17.20 $\pm$   1.60 &  17.07 $\pm$   0.99 &  20.58 $\pm$   0.85 \\
        Rolling Mill         &  37.79 $\pm$   3.74 &  34.13 $\pm$   0.00 &  70.20 $\pm$   0.34 &  71.49 $\pm$   0.33 \\
        Amazon               &   6.72 $\pm$   0.65 &   6.52 $\pm$   0.04 &  17.36 $\pm$   0.14 &  17.49 $\pm$   0.10 \\
        \bottomrule
    \end{tabular}
\end{table}

\subsection{Externally Mined Rules vs. End-to-End Learning}
\label{app:subsec_pattern_mining}

Existing rule- and pattern-mining methods are largely unsuitable for direct next-\seqElementText{} prediction: (i)~they often lack temporal information, making it unclear how long a rule should remain active; (ii)~many cannot handle irregular event times; (iii)~they lack a principled way to combine rules for prediction; and (iv)~frequent pattern mining risks large rule sets that hinder interpretability, while set mining scales poorly. 
For example, regarding (iv), \citet{2026_seqret} report that their rule \emph{set} miner requires 22 hours on a smaller Rolling Mill dataset (18k \seqElementText{}s), while \ourMethod{} runs on a larger version in 1 hour; moreover, they observe that frequent rule miners run out of 64GB memory due to the large number of candidates.
Nonetheless, \textbf{mined patterns can be incorporated into \ourMethod{}} by converting them into antecedents of Temporal Conjunctive Prediction Rules, enabling predictive use.

We evaluate the sequential pattern set miner \textsc{Sqs} \citep{2012_sqs} in this setting, leveraging its compact pattern sets. 
\textsc{Sqs} is tested on four of the five real-world datasets, since it does not finish within 24 hours on Amazon, whereas \ourMethod{} trains in approximately 30 minutes. 
As \textsc{Sqs} assumes regularly sampled sequences, we leverage integer timestamps in the irregularly sampled Holidays and Chorales datasets and insert padding \seqElementText{}s between actual \seqElementText{}s to induce equal spacing.
To enable prediction, we heuristically convert mined patterns into rules with delay intervals and apply symbolic refinement as in \ourMethod{} (Appendix~\ref{app:subsec_symbolic_rule_refinement}). 
We select the top rules based on the product of confidence and support to match the maximum capacity of \ourMethod{} ($\countAntNeurons$ in \cref{tab:exp_real_data_hyperparams}) and embed their antecedents into the model. 
The output layer is then trained as in the standard pipeline for \ourMethod{} with processed rules (Appendix~\ref{app:subsec_hyperparams_real}).

\paragraph{Results.}
The results in \cref{tab:table_pattern_mining_comparison} show that, beyond the efficiency gap on Amazon, \ourMethod{} (with end-to-end learning and symbolic refinement) outperforms the \textsc{Sqs}-based pipeline on three of the remaining four datasets, while \textsc{SQS} helps with rule discovery on Rolling Mill.
 
\begin{table}[ht]
    \centering
    \caption{
        \textit{Comparison of externally mined rules and end-to-end learning.}
        We report test cross-entropy (NLL) and accuracy (Acc.) in \%.
        \textsc{Sqs}~$\to$~\ourMethod{} denotes \ourMethod{} with antecedents mined via \textsc{Sqs}.
    }
    \label{tab:table_pattern_mining_comparison}
    \setlength{\tabcolsep}{8pt}
    \begin{tabular}{l rr rr}
        \toprule
                             & \multicolumn{2}{c}{\ourMethod{}} & \multicolumn{2}{c}{\textsc{Sqs}~$\to$~\ourMethod{}} \\ 
        \cmidrule(lr){2-3} \cmidrule(lr){4-5}
        \textbf{Datasets}    & NLL $\downarrow$     & Acc. $\uparrow$      & NLL $\downarrow$     & Acc. $\uparrow$      \\  
        \midrule
        Holidays             &       0.053 &       97.57 &                0.643 &                71.29 \\
        Chorales             &       2.404 &       21.00 &                2.518 &                12.95 \\
        Commits              &       4.161 &       19.60 &                4.344 &                13.67 \\
        Rolling Mill         &       2.581 &       34.13 &                1.987 &                54.21 \\
        \bottomrule
    \end{tabular}
\end{table}

\subsection{Independent Training of Rule and Residual Networks}
\label{app:subsec_independent_training}
We study whether adding the logits of the frozen \ourMethod{} network during training of the residual network, as in \ourMethodSideChannel{}, provides benefits over training both models independently.
To this end, for each dataset, we fix the corresponding \ourMethod{} model from \cref{tab:table_real_results} and compare \ourMethodSideChannel{} against a variant that combines \ourMethod{} with an independently trained transformer residual network.
This corresponds to training the transformer in \cref{tab:table_real_results} and combining it with the respective \ourMethod{} model.

In \cref{tab:table_independent_training}, we report mean results for residual network training over 4 random seeds.
We find that adding rule logits during training improves performance by enabling the residual network to learn corrective logits relative to rule-based predictions.

\begin{table}[hb]
    \centering
    \caption{
        \textit{\ourMethodSideChannel{} benefits from training the residual network with added rule logits.}
        We report mean and standard deviation of test cross-entropy (NLL) and accuracy (Acc.) in \% on real-world datasets over 4 random seeds for residual network training.
        ``Independent'' denotes \ourMethodSideChannel{} with independently trained rule and residual networks.
        Best results within one standard deviation are in \textbf{bold}.
    }
    \label{tab:table_independent_training}
    \setlength{\tabcolsep}{7pt}
    \begin{tabular}{l rr rr}
        \toprule
                             & \multicolumn{2}{c}{\ourMethodSideChannel{}} & \multicolumn{2}{c}{\ourMethodSideChannel{} (independent)} \\
        \cmidrule(lr){2-3} \cmidrule(lr){4-5}
        \textbf{Datasets}    & \multicolumn{1}{c}{NLL $\downarrow$} & \multicolumn{1}{c}{Acc. $\uparrow$}     & \multicolumn{1}{c}{NLL $\downarrow$} & \multicolumn{1}{c}{Acc. $\uparrow$}     \\
        \midrule
        Holidays             &  \textbf{0.000 $\pm$  0.000} & \textbf{100.00 $\pm$   0.00} &  \textbf{0.000 $\pm$  0.000} & \textbf{100.00 $\pm$   0.00} \\
        Chorales             &  \textbf{1.931 $\pm$  0.029} &  \textbf{37.16 $\pm$   0.72} &  2.234 $\pm$  0.035 &  33.69 $\pm$   0.61 \\
        Commits              &  \textbf{4.191 $\pm$  0.017} &  17.07 $\pm$   0.99 &  5.655 $\pm$  0.124 &  \textbf{20.71 $\pm$   1.12} \\
        Rolling Mill         &  \textbf{1.183 $\pm$  0.010} &  \textbf{70.20 $\pm$   0.34} &  1.386 $\pm$  0.004 &  \textbf{70.00 $\pm$   0.30} \\
        Amazon               &  \textbf{4.850 $\pm$  0.016} &  \textbf{17.36 $\pm$   0.14} &  5.760 $\pm$  0.015 &  15.05 $\pm$   0.05 \\
        \bottomrule
    \end{tabular}
\end{table}

\subsection{Ablation on the Lookback Window}
\label{app:subsec_ablation_lookback_window}
On Holidays and Chorales, we use a maximum lookback window $\maxLookback$ with natural meaning (one year and three 4/4 bars or four 3/4 bars, respectively). 
For the other real-world datasets, we tune the lookback window for \ourMethod, then apply it to all other methods, so they operate on the same data (Appendix~\ref{app:subsec_hyperparams_real}). 
To assess sensitivity to this choice, we ablate the lookback window size on Commits, where \ourMethod{} leads the symbolic baselines, using the same protocol as in \cref{tab:table_real_results}.

\begin{table}[ht]
    \centering
    \caption{
        \textit{Test accuracy in \% for different lookback windows $\maxLookback$ on Commits.}
    }
    \label{tab:table_ablation_lookback_window}
    \begin{tabular}{c c c c c}
        \toprule
        \multicolumn{1}{c}{Lookback Window}  & \multicolumn{1}{c}{\ourMethod{}} & \multicolumn{1}{c}{\textsc{CatBoost}} & \multicolumn{1}{c}{\textsc{XGBoost}} & \multicolumn{1}{c}{Transformer} \\
        \midrule
        12 &  18.32 & 17.81 & 16.68 & 19.71 \\
        24 &  19.60 & 15.55 & 17.03 & 20.10 \\
        36 &  13.65 & 15.46 & 16.26 & 20.23 \\
        48 &  12.68 & 17.34 & 16.24 & 20.14 \\
        \bottomrule
    \end{tabular}
\end{table}

\cref{tab:table_ablation_lookback_window} shows that the performance of \ourMethod drops for larger window sizes, likely because the hyperparameters (tuned for $\maxLookback=24$) transfer poorly to wider windows, where good \seqElementText{} delay intervals are harder to learn. Nonetheless, at both its selected window ($\maxLookback = 24$) and the shorter window ($\maxLookback = 12$), \ourMethod outperforms every configuration of baselines with symbolic decomposition in the table.\footnote{
    The best configuration of \textsc{PPM} on Commits uses a lookback window (context) of size~$1$ and achieves a test accuracy of 18.11\% (see \cref{tab:table_real_results}).
}

\subsection{Qualitative Results}
\label{app:subsec_qualitative_examples}
In this section, we use case studies to qualitatively evaluate the rules discovered by \ourMethod{} and baseline models. 

\subsubsection{Comparison across Models}
We first compare rules discovered by different models.
On Amazon, \ourMethod identifies $\{(\textit{physical movie}, [\varepsilon, 68])\}\ruleTo \{\textit{physical movie}\}$,\!\footnote{$\varepsilon$ is a small time step for resolving same-day purchases.} indicating that customers often purchase another movie within 68 days of a prior movie purchase.
A first-order Markov model (PPM) also captures that movies tend to follow movies, but lacks detailed temporal information, failing to use this dependency in 38\% of test cases where the rule is correctly active in \ourMethod{}.

Similar to \textsc{PPM}, the first tree in \textsc{XGBoost} for \textit{physical movie} (illustrated later in this section) uses the most recent \seqElementText{} at the root split, and it incorporates its associated time feature in subsequent nodes.
However, the smaller partition of the root split is already defined by a subset of 49 possible events, and subsequent splits further condition on other events, making the logic difficult to interpret.
\textsc{CatBoost} further obscures interpretability by internally mapping categorical variables to continuous representations that cannot be easily inverted to recover explicit rules.

Returning to our running example, \ourMethod discovers a similar rule on Amazon: 
``\textbf{If} \textit{a customer purchased a `computer case' within the last 43 days, a `computer processor' within the last 21 days}, \textbf{and} \textit{a `computer drive' within the last 69 days} \textbf{then} \textit{the next \seqElementText{} is likely to be the purchase of a `motherboard'}, \textit{a `power supply unit'} \textbf{or} \textit{a `graphics card'}.''
This rule occurs 114 times in the training data, covers 13\% of \seqElementText{}s when the antecedent is active, and generalizes to 30 test cases.
In contrast, \textsc{PPM} with order 3 cannot model temporal gaps and relies on exact event orderings.
As a result, among the 27 training occurrences where the three antecedent events appear contiguously, it observes our consequent events only twice and entirely misses \textit{power supply unit}, despite this item being purchased 39 times in the training data when our antecedent is active. 
These examples help explain the consistent performance gains of \ourMethod{} over \textsc{PPM} for synthetic data, which specifically incorporates such long-range dependencies.

We next illustrate the representations learned by different models on the Amazon dataset, focussing on their degree of direct inspectability and interpretability.

\paragraph{\ourMethodTitle{}.}
First, we give a \textit{complete} list of the rules extracted from \ourMethod{}, where consequent sets define \textbf{disjunctions}.
The resulting rule set is sufficiently compact to be fully enumerated and inspected by a human, without post-hoc rule selection:
{\small

\begin{enumerate}
    \item $\{(\textit{physical movie}, [\varepsilon, 68])\} \ruleTo \{\textit{physical movie}\}$
    \item $\{(\textit{computer processor}, [\varepsilon, 21]), (\textit{computer drive or storage}, [\varepsilon, 69]), (\textit{computer chassis}, [\varepsilon, 43])\}$ \\ $ \ruleTo \{\textit{video card}, \textit{system power device}, \textit{motherboard}\}$
    \item $\{(\textit{pet toy}, [\varepsilon, \varepsilon])\} \ruleTo \{\textit{pet toy}\}$
    \item $\{(\textit{shirt}, [\varepsilon, \varepsilon])\} \ruleTo \{\textit{sweatshirt}, \textit{sweater}, \textit{shirt}, \textit{skirt}, \textit{tunic}, \textit{shorts}, \textit{suspender}\}$
	\item $\{(\textit{food}, [\varepsilon, \varepsilon]), (\textit{grocery}, [\varepsilon, \varepsilon]), (\textit{fruit}, [\varepsilon, \varepsilon])\} \ruleTo \{\textit{swimwear}\}$
	\item $\{(\textit{health personal care}, [\varepsilon, \varepsilon]), (\textit{physical video game software}, [9, 9])\}$ \\ $ \ruleTo \{\textit{video game console}, \textit{console video games}, \textit{single board computer}\}$
	\item $\{(\textit{skin moisturizer}, [\varepsilon, \varepsilon])\} \ruleTo \{\textit{skin serum}, \textit{skin moisturizer}, \textit{sunscreen}\}$
	\item $\{(\textit{water}, [34, 34]), (\textit{camera digital}, [33, 33])\} \ruleTo \{\textit{water}, \textit{electronic device docking station}\}$
	\item $\{(\textit{pet food}, [28, 37]), (\textit{necktie}, [87, 97])\} \ruleTo \{\textit{pet food}\}$
	\item $\{(\textit{file folder}, [61, 61]), (\textit{hair coloring agent}, [7, 7])\}$ \\ $  \ruleTo \{\textit{hair coloring agent}, \textit{furniture floor protector}\}$
	\item $\{(\textit{body deodorant}, [62, 62]), (\textit{pajamas}, [87, 87])\} \ruleTo \{\textit{ce accessory}\}$
	\item $\{(\textit{health personal care}, [\varepsilon, \varepsilon]), (\textit{toy figure}, [47, 47]), (\textit{toys and games}, [32, 32]),$ \\ \quad $ (\textit{tabletop game}, [\varepsilon, \varepsilon]), (\textit{non riding toy vehicle}, [\varepsilon, \varepsilon]),(\textit{figurine}, [57, 57])\} $ \\ $  \ruleTo \{\textit{figurine}, \textit{toy figure playset}\}$
	\item $\{(\textit{pants}, [\varepsilon, \varepsilon])\} \ruleTo \{\textit{pants}, \textit{sweater}\}$
	\item $\{(\textit{fruit}, [\varepsilon, \varepsilon])\} \ruleTo \{\textit{meat}, \textit{poultry}, \textit{salad dressing}, \textit{fruit}, \textit{vegetable}, \textit{legume}, \textit{milk substitute}\}$
	\item $\{(\textit{sticker decal}, [\varepsilon, \varepsilon])\} \ruleTo \{\textit{sticker decal}\}$
	\item $\{(\textit{grocery}, [\varepsilon, \varepsilon])\} \ruleTo \{\textit{pantry}\}$
	\item $\{(\textit{earring}, [\varepsilon, 2])\} \ruleTo \{\textit{jewelry set}, \textit{earring}, \textit{jewelry}\}$
	\item $\{(\textit{gift card}, [13, 16]), (\textit{muscle roller}, [3, 6])\} \ruleTo \{\textit{clothes pin}, \textit{artificial tree}, \textit{ottoman}, \textit{radio}\}$
	\item $\{(\textit{laundry detergent}, [20, 21]), (\textit{contact lens cleaner}, [29, 31])\} \ruleTo \{\textit{snow pant}, \textit{hair roller}\}$
	\item $\{(\textit{collectible card}, [\varepsilon, 17])\} \ruleTo \{\textit{collectible card}\}$
	\item $\{(\textit{dairy based cheese}, [\varepsilon, 35]), (\textit{bread}, [\varepsilon, 3])\}$ \\ \quad $ \ruleTo \{\textit{dairy based cheese}, \textit{dairy based drink}, \textit{dairy based cream}, \textit{dairy based yogurt}, \textit{egg}\}$
	\item $\{(\textit{food}, [\varepsilon, \varepsilon])\} \ruleTo \{\textit{food}, \textit{poultry}, \textit{fruit}, \textit{fruit snack}, \textit{pizza}\}$
	\item $\{(\textit{other}, [\varepsilon, \varepsilon])\} \ruleTo \{\textit{utility knife}, \textit{saw blade}, \textit{splash pool}, \textit{sport helmet}, \textit{puppet}, \textit{utility magnet},\dots\}$
	\item $\{(\textit{gift card}, [7, 30]), (\textit{baby formula}, [12, 33])\} \ruleTo \{\textit{baby formula}\}$
	\item $\{(\textit{thermoplastic filament}, [\varepsilon, 84])\} \ruleTo \{\textit{thermoplastic filament}\}$
	\item $\{(\textit{dress}, [\varepsilon, \varepsilon])\} \ruleTo \{\textit{dress}\}$
	\item $\{(\textit{chewing gum}, [26, 44]), (\textit{protein supplement powder}, [\varepsilon, 94])\} \ruleTo \{\textit{chewing gum}, \textit{snack food bar}\}$
	\item $\{(\textit{false nail}, [\varepsilon, 9])\} \ruleTo \{\textit{nail polish}, \textit{nail file}, \textit{nail polish remover}, \textit{nail drill}, \textit{false nail}\}$
	\item $\{(\textit{mineral supplement}, [26, 92]), (\textit{leavening agent}, [22, 41])\} $ \\ $ \ruleTo \{\textit{mechanical components}, \textit{medical test strip}, \textit{leavening agent},$ \\ $ \textit{blood oxygen monitor},\textit{dietary supplements}, \textit{undergarment slip}\}$
	\item $\{(\textit{envelope}, [7, 70]), (\textit{envelope}, [81, 96]), (\textit{label}, [2, 94])\}$ \\ $  \ruleTo \{\textit{envelope}, \textit{cutting board}, \textit{thread cord}, \textit{camera cleaner}\}$
	\item $\{(\textit{protein drink}, [25, 42]), (\textit{physical tv series}, [25, 39])\} $ \\ $ \ruleTo \{\textit{protein drink}, \textit{two way radio}, \textit{snow pant}, \textit{bladed food peeler}, \textit{keycap}\}$
	\item $\{(\textit{computer processor}, [1, 24]), (\textit{drink flavored}, [27, 32])\} $ \\ $  \ruleTo \{\textit{fork}, \textit{furniture edge bumper}, \textit{overalls}, \textit{baby rattle}\}$
	\item $\{(\textit{battery}, [\varepsilon, 14]), (\textit{food bib}, [83, 90])\} \ruleTo \{\textit{guild home}, \textit{party favor}\}$
	\item $\{(\textit{shirt}, [4, 81]), (\textit{physical movie}, [5, 5])\} \ruleTo \{\textit{ladder}, \textit{saw}, \textit{gun cleaning kit}, \textit{games}\}$
	\item $\{(\textit{culinary salt}, [53, 56]), (\textit{medication}, [25, 45])\} \ruleTo \{\textit{flash card}, \textit{walking stick}\}$
	\item $\{(\textit{bra}, [\varepsilon, \varepsilon])\} \ruleTo \{\textit{bra}\}$
	\item $\{(\textit{dairy based yogurt}, [\varepsilon, \varepsilon]), (\textit{abis music}, [5, 97])\}$ \\ \quad $ \ruleTo \{\textit{flavored drink concentrate}, \textit{dairy based yogurt}, \textit{console video games}, \textit{egg}\}$
	\item $\{(\textit{sandal}, [\varepsilon, 3])\} \ruleTo \{\textit{sandal}\}$
	\item $\{(\textit{air compressor}, [4, 27])\} $ \\ $  \ruleTo \{\textit{sport helmet}, \textit{irrigation sprinkler}, \textit{single board computer}, \textit{unmanned aerial vehicle}\}$
	\item $\{(\textit{abis gift card}, [\varepsilon, 17])\} \ruleTo \{\textit{abis gift card}\}$
	\item $\{(\textit{health personal care}, [\varepsilon, \varepsilon])\} \ruleTo \{\textit{medication}, \textit{skin protectant}, \textit{nail drill}, \textit{wound dressing}, $ \\ $ \textit{medical compression garment}, \textit{personal care appliance}, \textit{rolling paper}\}$
	\item $\{(\textit{electronic gift card}, [\varepsilon, 48])\} \ruleTo \{\textit{electronic gift card}\}$
\end{enumerate} 

}

\clearpage

\paragraph{PPM.}
Second, we show a small subset---far less than $0.1\%$---of the empirical event counts used by a \textsc{PPM} Markov model of order $2$ or higher to generate next-event distributions.
Enumerating the full model would be infeasible without substantial pruning or summarization, preventing complete inspection of the learned statistics:
{\scriptsize
\begin{center}
    \begin{tabular}{ll rrrrrrr}
	\toprule
	& &\multicolumn{7}{c}{\textbf{Counts}} \\ 
	\cmidrule(lr){3-9}
	\multicolumn{2}{c}{\textbf{Context}} & abis book & other & pet food & gift card & shirt & nutritional supplement & \dots \\ 
	\midrule
	& abis book & 21622     & 883       & 363       & 530       & 291       & 359      & \dots \\ 
	& other & 1069      & 4541      & 716       & 237       & 491       & 254      & \dots \\ 
	abis book & abis book & 11476     & 316       & 110       & 182       & 100       & 108      & \dots \\ 
	& pet food & 567       & 248       & 5429      & 177       & 211       & 209      & \dots \\ 
	& gift card & 539       & 364       & 217       & 8626      & 98        & 149      & \dots \\ 
	& shirt & 466       & 194       & 114       & 87        & 5056      & 109      & \dots \\ 
	& nutritional supplement & 425       & 705       & 637       & 121       & 186       & 2428     & \dots \\ 
	& electronic cable & 260       & 503       & 141       & 153       & 88        & 123      & \dots \\ 
	& pants & 212       & 73        & 159       & 48        & 1045      & 52       & \dots \\ 
	& health personal care & 206       & 388       & 241       & 62        & 96        & 401      & \dots \\ 
	& cellular phone case & 219       & 209       & 95        & 101       & 54        & 107      & \dots \\ 
	& medication & 218       & 352       & 373       & 75        & 91        & 536      & \dots \\ 
	gift card & gift card & 160       & 95        & 60        & 6042      & 26        & 37       & \dots \\ 
	& skin moisturizer & 305       & 118       & 100       & 70        & 38        & 110      & \dots \\ 
	& shoes & 301       & 149       & 68        & 74        & 61        & 68       & \dots \\ 
	& food & 45        & 63        & 48        & 45        & 11        & 60       & \dots \\ 
	& vegetable & 144       & 46        & 54        & 25        & 11        & 21       & \dots \\ 
	& toy figure & 484       & 119       & 65        & 91        & 39        & 44       & \dots \\ 
	& toys and games & 426       & 144       & 43        & 79        & 45        & 52       & \dots \\ 
	& grocery & 69        & 129       & 103       & 24        & 21        & 69       & \dots \\ 
	& skin cleaning agent & 216       & 76        & 79        & 75        & 32        & 86       & \dots \\ 
	& headphones & 260       & 292       & 88        & 78        & 83        & 122      & \dots \\ 
	& pet supplies & 207       & 105       & 194       & 35        & 79        & 63       & \dots \\ 
	& battery & 124       & 214       & 99        & 59        & 34        & 86       & \dots \\ 
	& coffee & 110       & 135       & 163       & 81        & 36        & 120      & \dots \\ 
	& vitamin & 309       & 121       & 99        & 71        & 31        & 243      & \dots \\ 
	& cleaning agent & 66        & 179       & 105       & 47        & 26        & 76       & \dots \\ 
	& socks & 266       & 94        & 61        & 38        & 46        & 55       & \dots \\ 
	& screen protector & 239       & 122       & 46        & 90        & 95        & 65       & \dots \\ 
	pet food & pet food & 110       & 45        & 1872      & 29        & 47        & 51       & \dots \\ 
	& beauty & 81        & 105       & 78        & 50        & 38        & 98       & \dots \\ 
	& writing instrument & 442       & 120       & 70        & 53        & 45        & 51       & \dots \\ 
	& drink flavored & 67        & 74        & 116       & 77        & 23        & 76       & \dots \\ 
	shirt & shirt & 120       & 58        & 29        & 28        & 2029      & 32       & \dots \\ 
	& fruit & 6         & 25        & 14        & 21        & 3         & 10       & \dots \\ 
	& charging adapter & 126       & 147       & 63        & 68        & 48        & 57       & \dots \\ 
	& drinking cup & 129       & 173       & 53        & 72        & 66        & 42       & \dots \\ 
	& dress & 92        & 86        & 36        & 32        & 171       & 29       & \dots \\ 
	& portable electronic device cover & 162       & 68        & 25        & 35        & 83        & 39       & \dots \\ 
	& herbal supplement & 111       & 123       & 130       & 19        & 37        & 648      & \dots \\ 
	 \vdots &  \vdots &  \vdots &  \vdots &  \vdots &  \vdots &  \vdots &  \vdots &  $\ddots$ \\ 
	 \bottomrule 
\end{tabular}

\end{center}
}

\clearpage

\paragraph{Decision Trees.}
Third, we display the conditions along a \textit{single root-to-leaf path} from the first decision tree in \textsc{XGBoost} that predicts the event \textit{physical movie} ($\seqElement_{\predInd}$), corresponding to one of thousands of unique paths induced by the model.
This path exemplifies how predictions are formed through a conjunction of feature split conditions.
Even for relatively shallow trees, the use of subset conditions for categorical variables can make the semantic interpretation of a single root-to-leaf path nontrivial, let alone that of the full ensemble.
Paths through trees learned by \textsc{CatBoost} would appear similar once the internal continuous encodings of categorical variables are inverted back to their original categories:
{\small
\begin{align*}
        &\seqElement_{\predInd-1} \not\in &&\{\textit{headphones}, \textit{shaving agent}, \textit{computer add on}, \textit{computer processor}, \textit{amazon tablet}, \\
        & && \quad \textit{apparel belt}, \textit{portable electronic device stand}, \textit{other}, \textit{input mouse}, \textit{cellular phone case}, \\
        & && \quad \textit{handbag}, \textit{underpants}, \textit{amazon tablet accessory}, \textit{tape measure}, \\
        & && \quad \textit{computer drive or storage}, \textit{mouthwash}, \textit{toilet paper}, \textit{tooth cleaning agent}, \\
        & && \quad \textit{electronic cable}, \textit{gift card}, \textit{sim card}, \textit{bed frame}, \textit{herb}, \textit{measuring cup}, \\
        & && \quad \textit{dairy based butter}, \textit{dairy based cheese}, \textit{jerky}, \textit{meat}, \textit{sexual stimulation device}, \\
        & && \quad \textit{file folder}, \textit{vegetable}, \textit{showerhead}, \textit{screen protector}, \textit{portable electronic device cover}, \\
        & && \quad \textit{inkjet printer ink}, \textit{chair}, \textit{cellular phone}, \textit{consumer electronics}, \textit{video game console}, \\
        & && \quad \textit{flat screen display mount}, \textit{food}, \textit{grocery}, \textit{poultry},\\
        & && \quad \textit{wireless accessory}, \textit{baking mix}, \textit{greeting card}, \textit{body deodorant}, \textit{hanging rod}, \textit{skirt}\}  \\
        & \land\quad \seqElement_{\predInd-8} \not\in &&\{\textit{dishware bowl}, \textit{skin cleaning wipe}, \textit{showerhead}, \textit{cotton pad}, \textit{furniture cover}, \\ 
        & && \quad\textit{juice and juice drink}, \textit{envelope}, \textit{measuring cup}, \textit{pitcher}\} \\
        & \land\quad \seqElement_{\predInd-39} \in &&\{\textit{flash memory}, \textit{bathwater additive}, \textit{sugar substitute}, \textit{skirt}\} \\
        & \land\quad \seqElement_{\predInd-3} \not\in &&\{\textit{headphones}, \textit{dishware bowl}, \textit{shaving agent}, \textit{computer add on}, \textit{computer processor}, \\
        & && \quad \textit{amazon tablet}, \textit{apparel belt}, \textit{portable electronic device stand}, \textit{other}, \textit{input mouse}, \\
        & && \quad \textit{cellular phone case}, \textit{handbag}, \textit{underpants}, \textit{amazon tablet accessory}, \textit{tape measure}, \\
        & && \quad \textit{computer drive or storage}, \textit{mouthwash}, \textit{toilet paper}, \textit{tooth cleaning agent}, \\
        & && \quad \textit{electronic cable}, \textit{gift card}, \textit{sim card}, \textit{bed frame}, \textit{herb}, \textit{dairy based butter}, \\
        & && \quad \textit{dairy based cheese}, \textit{jerky}, \textit{meat}, \textit{sexual stimulation device}, \textit{pet food}\} \\
        & \land \quad \seqElement_{\predInd-13} \not\in &&\{\textit{headphones}, \textit{dishware bowl}, \textit{shaving agent}, \textit{computer add on}, \textit{computer processor}, \\
        & && \quad \textit{amazon tablet}, \textit{apparel belt}, \textit{portable electronic device stand}, \textit{other}, \textit{input mouse}, \\
        & && \quad \textit{cellular phone case}, \textit{body positioner}, \textit{handbag}, \textit{underpants}, \textit{amazon tablet accessory}, \\
        & && \quad \textit{tape measure}, \textit{computer drive or storage}, \textit{mouthwash}, \textit{toilet paper}, \\
        & && \quad \textit{tooth cleaning agent}, \textit{electronic cable}, \textit{screwdriver}, \textit{gift card}, \textit{edible oil vegetable}, \\
        & && \quad \textit{sim card}, \textit{bed frame}, \textit{herb}, \textit{dairy based butter}, \textit{dairy based cheese}, \textit{jerky}, \textit{meat}, \\
        & && \quad \textit{sexual stimulation device}, \textit{file folder}, \textit{screen protector}, \textit{memory card reader}, \\ 
        & && \quad \textit{portable electronic device cover}\} \\
        & \land \quad \seqTime_{\predInd} - \seqTime_{\predInd-1} < && 10
\end{align*}

}

\subsubsection{Analysis of \ourMethod{} across Datasets}
Having discussed Amazon above, we now examine representative rules discovered by \ourMethod{} on the remaining datasets, omitting Rolling Mill, since is derived from proprietary data and contains only integer \seqElementText{}s.

\paragraph{Holidays.} 
Holidays has known ground-truth temporal structure, letting us verify rule correctness directly: some holidays occur on a fixed date (e.g., New Year's Day), others at a fixed offset from Easter Sunday. 
Most extracted rules are exact with perfect confidence, i.e., the antecedent activates only when the consequent holds. 
Examples include:
\begin{itemize}
    \item $\{(\textit{Good Friday}, [3,3])\} \ruleTo \{\textit{Easter Monday}\}$
    \item $\{(\textit{Ascension Day}, [11,11])\} \ruleTo \{\textit{Whit Monday}\}$.
\end{itemize}
The remaining rules such as $\{(\textit{New Year's Day}, [80, 112])\} \ruleTo \{\textit{Good Friday}\}$, are also correct but below 100\% confidence (here Easter Monday can fall in the activation window). 
Each of the 14 holidays is covered by at least one rule, with some redundancy where multiple rules predict the same event. 

\paragraph{Commits.} 
On this dataset, rules capture common author-succession patterns of Git commits, e.g.,
\begin{itemize}
    \item $\{(\textit{tridge}, [1,1])\} \ruleTo \{\textit{vlendec}, \textit{derrell}, \textit{vl}\}$ (support: $539$, confidence: $34\%$)
    \item $\{(\textit{sahlberg}, [3,16])\} \ruleTo \{\textit{sahlberg}\}$ (support: $110$, confidence: $10\%$)
    \item $\{(\textit{jra}, [2,8]), (\textit{kseeger}, [1,3]), (\textit{martin}, [1,7])\} \ruleTo \{\textit{sahlberg}, \textit{obnox}, \textit{martin}\}$ (support: $27$, confidence: $59\%$),
\end{itemize}
where the parentheses report training-set statistics. 
These rules exhibit non-trivial temporal relations, although we lack domain expertise to confirm whether they match true workflows. Nonetheless, the rules are interpretable to an outsider and form the basis for the predictions of \ourMethod.

\paragraph{Chorales.}
The two rules with the highest training support
\begin{itemize}
    \item $\{(\textit{A4}, [1,4])\} \ruleTo \{\textit{G4}, \textit{B4}, \textit{G\#/Ab4}, \textit{A\#/Bb4}\}$ (support: 308, confidence: 68\%)
    \item $\{(\textit{C\#/Db5}, [2,8])\} \ruleTo \{\textit{D5}, \textit{B4}, \textit{C\#/Db5}\}$ (support: 254, confidence: 71\%),
\end{itemize}
capture expected melodic tendencies: the first predicts movement to a note's immediate whole-tone and semitone neighbors; the second similarly captures whole-tone ($\textit{C\#5}\to\textit{B4}$) and semitone ($\textit{C\#5}\to\textit{D5}$) motion, with the remaining consequent (note repetition) reflecting another common continuation consistent with its 71\% confidence. 
The rule
\begin{itemize}
    \item $\{(\textit{C5}, [6, 18]), (\textit{A4}, [1, 8])\} \ruleTo \{\textit{G4}, \textit{F4}\}$ (support: 92, confidence: 43\%)
\end{itemize}
captures a descending line, which requires two distinct delay intervals to express. 

\paragraph{Rule Artifacts.}
Alongside rules with meaningful support and confidence, such as the ones discussed above, \ourMethod{} also surfaces some less reliable ones: high-confidence, low-support rules that overfit individual occurrences, as well as some low-confidence rules.
For example, on Amazon
\begin{itemize}
    \item $\{(\textit{body deodorant}, [62, 62]), (\textit{pajamas}, [87, 87])\} \ruleTo \{\textit{ce accessory}\}$ 
\end{itemize}
has confidence 100\% but only a support of 2 on the train set.
Such rules are easy to identify and filter by their low support or confidence, if desired.

\newpage
\section*{NeurIPS Paper Checklist}
\colorlet{blue}{storeblue}
\colorlet{orange}{storeorange}

\begin{enumerate}

\item {\bf Claims}
    \item[] Question: Do the main claims made in the abstract and introduction accurately reflect the paper's contributions and scope?
    \item[] Answer: \answerYes{} %
    \item[] Justification: As stated, we introduce a neuro-symbolic model that learns and applies sequential rules for next-\seqElementText{} prediction.
    We empirically support the claims on performance and interpretability made in the abstract and introduction through results in \cref{sec:experiments} and Appendices~\ref{app:sec_results_synthetic} and~\ref{app:sec_results_real}.

\item {\bf Limitations}
    \item[] Question: Does the paper discuss the limitations of the work performed by the authors?
    \item[] Answer: \answerYes{} %
    \item[] Justification: We discuss the limitations of our work in \cref{sec:conclusion}, particularly regarding runtime and scalability in Appendix~\ref{app:sec_discussion_complexity}, and acknowledge that our approach is based on interpretable rule structures that may not fully capture all data dependencies.

\item {\bf Theory assumptions and proofs}
    \item[] Question: For each theoretical result, does the paper provide the full set of assumptions and a complete (and correct) proof?
    \item[] Answer: \answerYes{} %
    \item[] Justification: We use temperature annealing to obtain continuous relaxations of Boolean logic and show in Appendix~\ref{app:sec_convergence_proofs} that these functions converge to their corresponding logical operators as temperature approaches zero.

    \item {\bf Experimental result reproducibility}
    \item[] Question: Does the paper fully disclose all the information needed to reproduce the main experimental results of the paper to the extent that it affects the main claims and/or conclusions of the paper (regardless of whether the code and data are provided or not)?
    \item[] Answer: \answerYes{} %
    \item[] Justification: We describe our models in \cref{sec:methodology} and Appendix~\ref{app:sec_rule_network_details}, and our evaluation setup in \cref{sec:experiments} and Appendices~\ref{app:sec_real_world_dataset_stats}, \ref{app:sec_hyperparameters}.
    We further provide our source code as supplemental material to facilitate reproducibility.

\item {\bf Open access to data and code}
    \item[] Question: Does the paper provide open access to the data and code, with sufficient instructions to faithfully reproduce the main experimental results, as described in supplemental material?
    \item[] Answer: \answerYes{} %
    \item[] Justification: We provide the source code with instructions in the supplemental material. We include references for accessing the external datasets used in our experiments.

\item {\bf Experimental setting/details}
    \item[] Question: Does the paper specify all the training and test details (e.g., data splits, hyperparameters, how they were chosen, type of optimizer) necessary to understand the results?
    \item[] Answer: \answerYes{} %
    \item[] Justification: We describe our evaluation procedure in \cref{sec:experiments}, including training and test protocols. We provide additional details on dataset processing in Appendix~\ref{app:sec_real_world_dataset_stats}, and detail hyperparameter selection and model optimization in Appendix~\ref{app:sec_hyperparameters}.

\item {\bf Experiment statistical significance}
    \item[] Question: Does the paper report error bars suitably and correctly defined or other appropriate information about the statistical significance of the experiments?
    \item[] Answer: \answerYes{} %
    \item[] Justification: For synthetic data (\cref{subsec:synthetic_data}, Appendix \ref{app:sec_results_synthetic}), we report mean performance and standard deviation across 10 independently sampled rule sets, each used to generate an independent training and corresponding test dataset.
    For real data, we report mean results and standard deviation across 4 random seeds in Appendix \ref{app:subsec_variability}.

\item {\bf Experiments compute resources}
    \item[] Question: For each experiment, does the paper provide sufficient information on the computer resources (type of compute workers, memory, time of execution) needed to reproduce the experiments?
    \item[] Answer: \answerYes{} %
    \item[] Justification: We report hardware and execution times in Appendix~\ref{app:sec_hardware_and_compute}. We also discuss runtime scalability in Appendix~\ref{app:sec_discussion_complexity}.
    
\item {\bf Code of ethics}
    \item[] Question: Does the research conducted in the paper conform, in every respect, with the NeurIPS Code of Ethics \url{https://neurips.cc/public/EthicsGuidelines}?
    \item[] Answer: \answerYes{} %
    \item[] Justification: We have reviewed the NeurIPS Code of Ethics and our research conforms to it.

\item {\bf Broader impacts}
    \item[] Question: Does the paper discuss both potential positive societal impacts and negative societal impacts of the work performed?
    \item[] Answer: \answerNA{} %
    \item[] Justification: We introduce a new architecture for sequential rule learning and prediction without targeting a specific application or deployment setting. Hence, our work has no direct societal impacts beyond those common to interpretable machine learning models.
    
\item {\bf Safeguards}
    \item[] Question: Does the paper describe safeguards that have been put in place for responsible release of data or models that have a high risk for misuse (e.g., pre-trained language models, image generators, or scraped datasets)?
    \item[] Answer: \answerNA{} %
    \item[] Justification: The paper poses no high misuse risks.

\item {\bf Licenses for existing assets}
    \item[] Question: Are the creators or original owners of assets (e.g., code, data, models), used in the paper, properly credited and are the license and terms of use explicitly mentioned and properly respected?
    \item[] Answer: \answerYes{} %
    \item[] Justification: We use only publicly available datasets and software sources, which we cite accordingly and use in accordance with the licenses and terms of use specified by their original creators. 
    In particular, we report dataset sources and their licenses when specified in Appendix~\ref{app:sec_real_world_dataset_stats}.

\item {\bf New assets}
    \item[] Question: Are new assets introduced in the paper well documented and is the documentation provided alongside the assets?
    \item[] Answer: \answerNA{} %
    \item[] Justification: We do not introduce new assets beyond the code and instructions provided for reproducibility.

\item {\bf Crowdsourcing and research with human subjects}
    \item[] Question: For crowdsourcing experiments and research with human subjects, does the paper include the full text of instructions given to participants and screenshots, if applicable, as well as details about compensation (if any)? 
    \item[] Answer: \answerNA{} %
    \item[] Justification: The paper does not involve crowdsourcing nor research with human subjects.

\item {\bf Institutional review board (IRB) approvals or equivalent for research with human subjects}
    \item[] Question: Does the paper describe potential risks incurred by study participants, whether such risks were disclosed to the subjects, and whether Institutional Review Board (IRB) approvals (or an equivalent approval/review based on the requirements of your country or institution) were obtained?
    \item[] Answer: \answerNA{} %
    \item[] Justification: The paper does not involve crowdsourcing nor research with human subjects.

\item {\bf Declaration of LLM usage}
    \item[] Question: Does the paper describe the usage of LLMs if it is an important, original, or non-standard component of the core methods in this research? Note that if the LLM is used only for writing, editing, or formatting purposes and does \emph{not} impact the core methodology, scientific rigor, or originality of the research, declaration is not required.
    \item[] Answer: \answerNA{} %
    \item[] Justification: The core method development in this research does not involve LLMs as any important, original, or non-standard components.

\end{enumerate}

\end{document}